\pdfoutput=1

\documentclass[11pt]{article}

\usepackage[preprint]{acl}

\usepackage{times}
\usepackage{latexsym}

\usepackage[T1]{fontenc}

\usepackage[utf8]{inputenc}

\usepackage{microtype}

\usepackage{inconsolata}

\usepackage{graphicx}

\usepackage{tikz}
\usepackage{booktabs}
\usepackage{makecell}
\usepackage{multirow}
\usepackage{graphicx}
\usepackage{longtable}
\usepackage{multicol}
\usepackage{adjustbox}
\usepackage{threeparttable}
\usepackage{amsmath}
\usepackage{cleveref}
\usepackage{tcolorbox}
\tcbuselibrary{breakable}
\usepackage{xcolor}
\usepackage{array}
\usepackage{enumitem}
\usepackage{fancyhdr}
\usetikzlibrary{calc}

\definecolor{examplecolor}{RGB}{135,206,235} 

\newtcolorbox{examplebox}[1][]{%
  colback=examplecolor!15, 
  colframe=examplecolor!30, 
  fonttitle=\bfseries,
  coltitle=black,
  title=#1,
  sharp corners
}

\definecolor{instructioncolor}{RGB}{230,250,200} 

\newtcolorbox{instructionbox}[1][]{%
  colback=instructioncolor!25, 
  colframe=instructioncolor!50, 
  fonttitle=\bfseries,
  coltitle=black,
  title=#1,
  sharp corners,
  breakable,
  width=\textwidth
}

\definecolor{surveycolor}{RGB}{255,250,205} 

\newcommand{\answerbox}[1]{
  \vspace{0.1cm}
  \noindent\framebox[\textwidth]{\parbox[c][#1][t]{\dimexpr\textwidth-10pt}{}}
  \vspace{0.1cm}
}

\newtcolorbox{surveybox}[1][]{%
  colback=surveycolor!25,
  colframe=surveycolor!50,
  fonttitle=\bfseries\centering,
  coltitle=black,
  title=#1,
  sharp corners,
  breakable,
  width=0.85\textwidth, 
  top=0.5cm,
  bottom=0.5cm,
  left=0.7cm,
  right=0.7cm 
}

\definecolor{rulecolor}{RGB}{135,206,235} 

\newtcolorbox{rulebox}[1][]{%
  colback=rulecolor!15, 
  colframe=rulecolor!40, 
  fonttitle=\bfseries,
  coltitle=black,
  title=#1,
  sharp corners,
  breakable,
  width=\textwidth,
  top=0.5cm,
  bottom=0.5cm,
  left=0.7cm,
  right=0.7cm
}

\definecolor{guidecolor}{RGB}{135,206,235} 

\newtcolorbox{guidebox}[1][]{%
  colback=guidecolor!15, 
  colframe=guidecolor!40, 
  fonttitle=\bfseries,
  coltitle=black,
  title=#1,
  sharp corners,
  breakable,
  width=\textwidth,
  top=0.5cm,
  bottom=0.5cm,
  left=0.7cm,
  right=0.7cm
}

\title{\textsc{MaXIFE}: Multilingual and Cross-lingual Instruction Following Evaluation}

\author{First Author \\
  Affiliation / Address line 1 \\
  Affiliation / Address line 2 \\
  Affiliation / Address line 3 \\
  \texttt{email@domain} \\\And
  Second Author \\
  Affiliation / Address line 1 \\
  Affiliation / Address line 2 \\
  Affiliation / Address line 3 \\
  \texttt{email@domain} \\}

\author{
 \textbf{Yile Liu\textsuperscript{1,2}},
 \textbf{Ziwei Ma\textsuperscript{2}},
 \textbf{Xiu Jiang\textsuperscript{2}},
 \textbf{Jinglu Hu\textsuperscript{1}},
 \textbf{Jing Chang\textsuperscript{2}},
 \textbf{Liang Li\textsuperscript{2}\thanks{Corresponding author.}}
\\
 \textsuperscript{1}Waseda University
\\
 \textsuperscript{2}OPPO AI Center, Shenzhen, China
\\
 \texttt{irei.liu@asagi.waseda.jp}\texttt{,} \texttt{jinglu@waseda.jp}
\\
 \texttt{\{maziwei, jiangxiu, changjing, liliang16\}@oppo.com}
}

\begin{document}
\maketitle
\begin{abstract}

With the rapid adoption of large language models (LLMs) in natural language processing, the ability to follow instructions has emerged as a key metric for evaluating their practical utility. However, existing evaluation methods often focus on single-language scenarios, overlooking the challenges and differences present in multilingual and cross-lingual contexts. To address this gap, we introduce \textbf{MaXIFE}: a comprehensive evaluation benchmark designed to assess instruction-following capabilities across 23 different languages with 1667 verifiable instruction tasks. MaXIFE integrates both Rule-Based Evaluation and Model-Based Evaluation, ensuring a balance of efficiency and accuracy. We applied MaXIFE to evaluate several leading commercial LLMs, establishing baseline results for future comparisons. By providing a standardized tool for multilingual instruction-following evaluation, MaXIFE aims to advance research and development in natural language processing.

\end{abstract}

\section{Introduction}

With the widespread application of LLMs in the field of natural language processing (NLP)~\cite{achiam2023gpt, dubey2024llama, yang2024qwen2}, the instruction-following capabilities of models has become a key indicator for measuring their practical application value~\cite{wei2022finetuned, mishra-etal-2022-cross, zhong-etal-2021-adapting-language}. The strength of instruction-following capabilities directly affect whether a model can accurately understand and execute user intentions, thus completing complex and diverse tasks. This capability transcends mere text generation and is intrinsically tied to the model's alignment with human-provided instructions. A model that can accurately and efficiently follow instructions can better complete tasks in collaboration with humans, enhancing the effectiveness of human-computer interaction. If a model cannot correctly follow user instructions, its other advantages will be greatly reduced, especially in complex scenarios where lack of instruction-following capabilities may lead to erroneous decisions or results.

\begin{figure}
    \centering
    \includegraphics[width=1\linewidth]{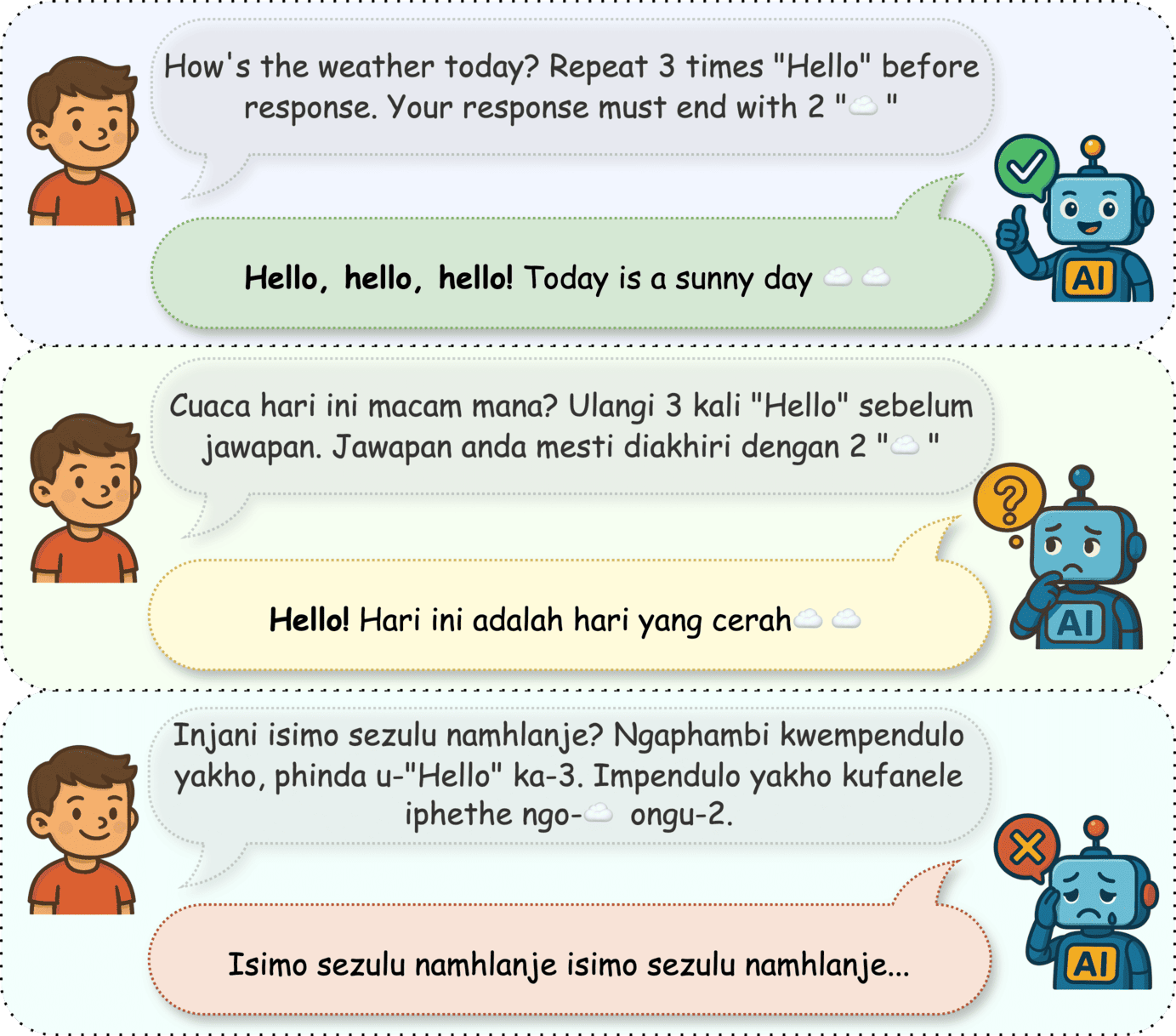}
    \caption{LLMs have different instruction-following capabilities across 3 different languages: English, Malay and Zulu.}
    \label{fig:badcase}
\end{figure}

\begin{figure*}
    \centering
    \includegraphics[width=1\linewidth]{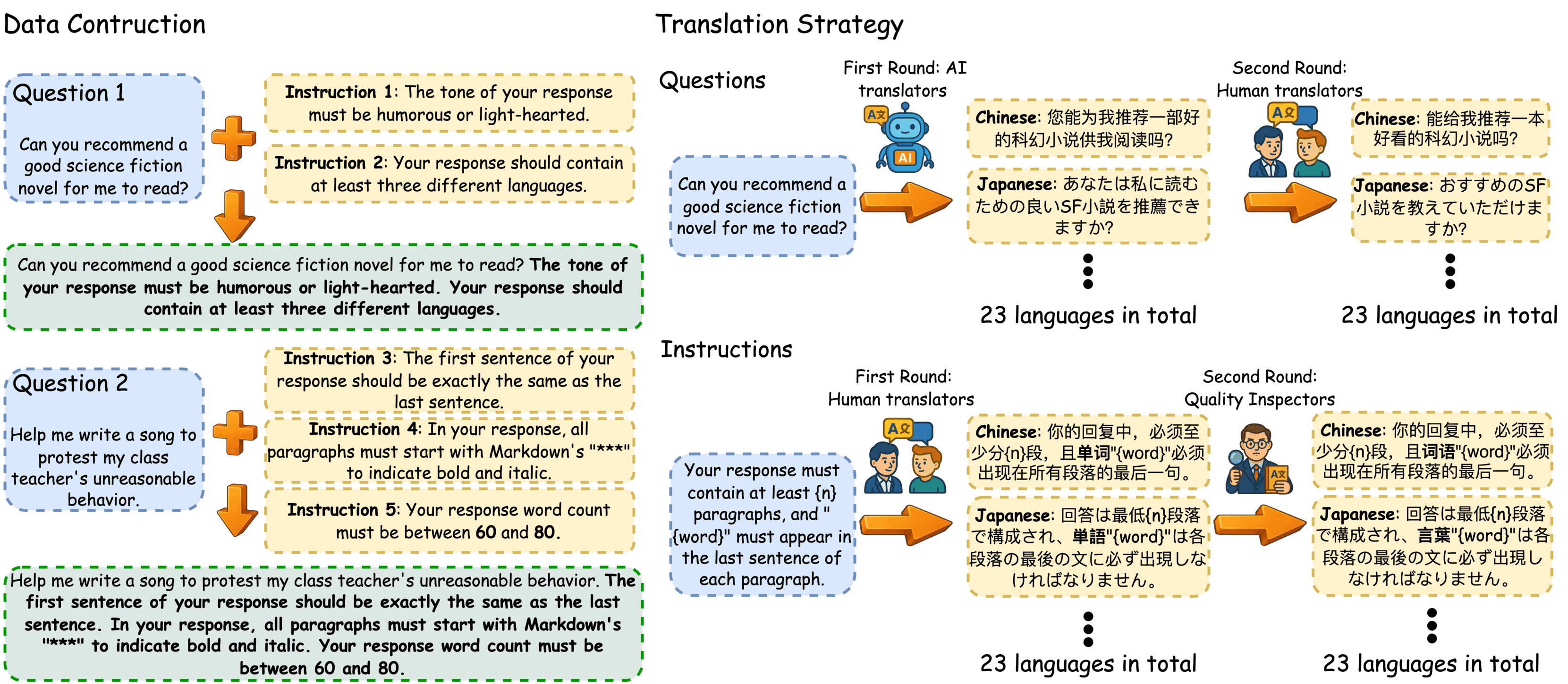}
    \caption{MaXIFE Structure, its evaluation dataset composition, and evaluation strategy. We provide 795 Basic Questions and 1667 Instructions, where each Basic Question is combined with 1-3 Instructions to form one piece of evaluation data. In the translation phase, we established human translation processes for both Questions and Instructions to verify quality and ensure accuracy. The translation of Questions focuses more on authenticity of expression, while the translation of Instructions emphasizes precision and rigor in word choice, as well as the accuracy of terminology in specific contexts within particular languages.}
    \label{MaXIFE_Structure_and_Workflow}
\end{figure*}

The multilingual and cross-lingual capabilities of models are equally crucial, especially in a globalized context where language should not be a barrier to technological applications. Models with multilingual and cross-lingual capabilities not only promote global technological fairness but also help more language communities equally enjoy the convenience brought by artificial intelligence technology~\cite{conneau-etal-2020-unsupervised, liu-etal-2020-multilingual-denoising, xue2021mt5massivelymultilingualpretrained}.

However, there is currently no evaluation set or benchmark that can parallelly assess the instruction-following capabilities of models in multilingual and cross-lingual contexts. Existing evaluations of instruction-following capabilities often focus on a single language~\cite{zhou2023instruction, wen2024benchmarkingcomplexinstructionfollowingmultiple, he2024can, li-etal-2024-instruction}, especially using English for testing, ignoring the performance in other languages and their potential differences. With globalization, launching an evaluation set that can assess models' instruction-following capabilities in multilingual environments is of great significance. First, it can help us more comprehensively understand the model's performance under different languages and discover potential weaknesses or advantages in multilingual environments. Additionally, such an evaluation set serves as a foundation for future model improvements and optimizations, thereby advancing the global application and development of models. Finally, a multilingual and cross-lingual evaluation set can promote technological equity, allowing users from different language backgrounds to equally enjoy the progress of artificial intelligence technology.

In essence, when non-English speakers interact with LLMs, they predominantly input prompts in their native languages rather than first translating instructions into English. Moreover, as shown in ~\Cref{fig:badcase}, the same large model may demonstrate varying levels of instruction-following capabilities across different languages. Consequently, evaluating a model's instruction-following capabilities solely through English evaluation sets and benchmarks proves inadequate. This necessitates the development of both multilingual instruction-following evaluation methodologies and corresponding multilingual instruction-following benchmarks.

To address the above problems, we propose MaXIFE (\textbf{M}ultilingual \textbf{a}nd \textbf{Cross}-lingual \textbf{I}nstruction \textbf{F}ollowing \textbf{E}valuation), an evaluation benchmark specifically designed to assess models' instruction-following capabilities in multilingual and cross-lingual environments. MaXIFE covers instruction evaluation data in 23 languages, and these instructions can be verified to see whether they are followed by the model. MaXIFE's design enables the evaluation process to be fully automated, allowing researchers to obtain detailed evaluation results directly through preset evaluation frameworks. By analyzing the evaluation results, researchers can not only deeply analyze the model's instruction-following performance in a specific language but also easily compare differences across different languages.

Currently, there are three main methods to evaluate model performance: manual evaluation~\cite{karpinska-etal-2021-perils, taori2023stanford, zheng2023judgingllmasajudgemtbenchchatbot, ouyang2022traininglanguagemodelsfollow}, Rule-Based Evaluation~\cite{zhou2023instruction, chen2021evaluatinglargelanguagemodels, pillutla2021mauve}, and Model-Based Evaluation~\cite{gao2021framework, askell2021general, chang2024survey}. Manual evaluation, although detailed, is costly and highly subjective. Rule-Based Evaluation is efficient but may not cover all possible outputs. Model-based Evaluation has been shown to achieve good results on certain tasks but is time-consuming and may not fully align with human judgments. Therefore, MaXIFE adopts two evaluation strategies: for instructions that can be evaluated through rules, Rule-Based Evaluation is used; for instructions requiring semantic understanding, Model-Based Evaluation is adopted. This strategy balances evaluation efficiency and accuracy.


In summary, we propose MaXIFE as a benchmark for evaluating instruction-following capabilities of LLMs in multilingual and cross-lingual settings, using a set of parallel prompts containing verifiable instructions across 23 languages. These verifiable instructions use simple, interpretable, and deterministic programs to verify whether the generated responses follow these instructions. The dataset composition and translation strategy are shown in~\Cref{MaXIFE_Structure_and_Workflow}.


\section{Related Work}

\textbf{LLM Benchmarking.} Recently, many excellent and logically rigorous benchmarks for large models have emerged, such as C-eval~\cite{huang2024c}, BIG-bench~\cite{srivastava2022beyond}, MMLU-Pro~\cite{wang2024mmlu}, and AGIEval~\cite{zhong2023agieval}. However, these benchmarks often only select one of the evaluation methods between Rule-Based Evaluation or Model-Based Evaluation~\cite{zhang2019bertscore, gao2021framework}. In contrast, MaXIFE combines the two, applying each to tasks they excel at, significantly improving the accuracy of evaluation results.

\textbf{Instruction-Following Evaluation. }Recently, many instruction-tuned models have demonstrated promising performance~\cite{ouyang2022traininglanguagemodelsfollow, chung2024scaling, wang2022self}. Instruction-following evaluation has attracted increasing research interest, with several automated evaluation benchmarks such as IFEval~\cite{zhou2023instruction}, CELLO~\cite{he2024can}, FollowBench~\cite{jiang-etal-2024-followbench} and CIF-Bench~\cite{li-etal-2024-cif} being proposed to address this challenge. Building upon these pioneering works, MaXIFE extends the rule-based evaluation paradigm by incorporating nearly twice the number of instruction types (from 25 to 47) compared to IFEval and enhancing the Model-Based Evaluation module. Furthermore, while IFEval only supports English evaluation, MaXIFE extends the evaluation scope by supporting 23 languages and introducing comprehensive modules for assessing models' cross-lingual capabilities.

\textbf{Multilingual and Cross-lingual Evaluation.} XTREME~\cite{hu2020xtreme} pioneered a comprehensive and standardized paradigm for multilingual evaluation of LLMs. Subsequently, evaluation sets and benchmarks such as SIB-200~\cite{adelani2023sib}, CrossSum~\cite{bhattacharjee2021crosssum}, BELEBELE~\cite{bandarkar2023belebele}, FLORES-101~\cite{goyal2022flores} evaluated large models on various tasks, drawing rigorous conclusions. However, instruction-following tasks, as the most fundamental capability of models, have not been generalized to the multilingual field. MaXIFE fills this gap, becoming the first multilingual/cross-lingual instruction-following task benchmark, designed to support the broader adoption and accessibility of LLMs.

\section{Dataset}
Considering the requirements of LLMs for input accuracy when performing instruction-following tasks, based on our current resources, we chose 23 languages to construct our evaluation dataset. Details of the 23 languages can be found in~\Cref{tab:languages}.

\begin{figure}
    \centering
    \includegraphics[width=1\linewidth]{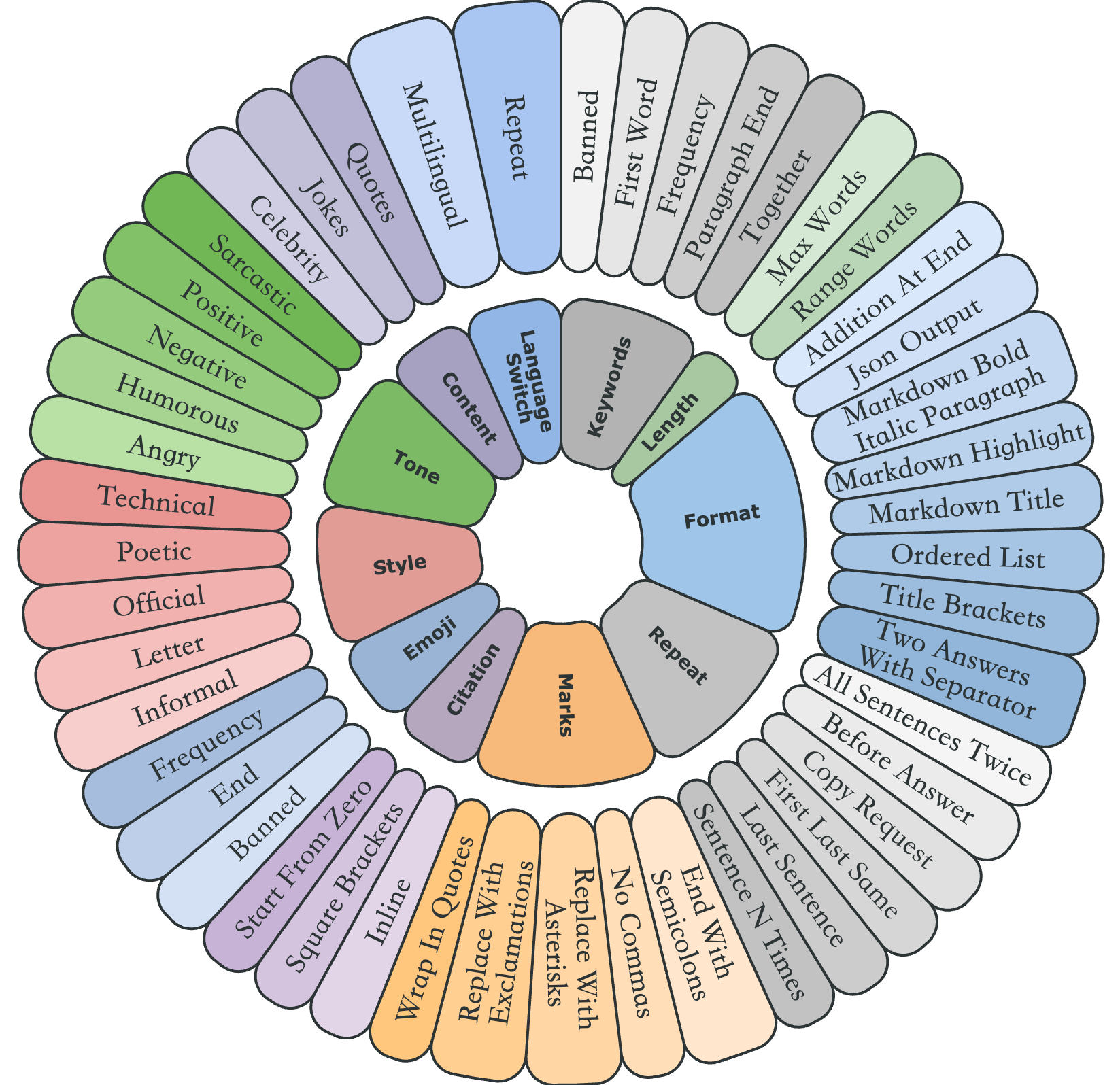}
    \caption{11 Instruction Categories and 47 Instruction Subcategories.}
    \label{11_Instruction_Categories_and_47_Instruction_Subcategories}
\end{figure}

\subsection{Data Construction}

For the convenience of expansion, we divide each data point into two parts: ``Basic Questions'' and ``Instructions''. First, we used a questionnaire method (see~\Cref{Survey Questionnaire} for details) to collect real user data from various sources, including but not limited to researchers within our group, other researchers, native speakers of different languages, and linguistics experts. Our focus was on collecting Basic Questions and Instructions that respondents commonly use when interacting with LLMs. Basic Questions are descriptions of frequently asked requirements, such as ``Can you help me come up with something funny to text a friend?''. Instructions refer to specific requirements that need to be followed by the model, such as ``Your response word count must be between 50 and 80.''

Subsequently, we expanded these collected Basic Questions using \textit{GPT-4o} and \textit{Claude-3.5 Sonnet} to generate more data following the same format. For example, based on the collected Basic Question ``Can you help me come up with something funny to text a friend?'', we generated variations like ``Can you help me come up with ideas for a birthday surprise for my mom?''. After this expansion, we obtained 795 Basic Questions.

For the collected Instructions, we chose not to directly augment the data but instead selected 47 high-quality instructions. We selected and templatized 18 scalable instructions, for example, ``In the response, the word or phrase `Computer' should appear at least 6 times'' became ``In the response, the word or phrase '\{word\}' should appear \{natural\_relation\} \{word\_num\} times.'' We organized these 47 instructions into 47 subcategories, and based on their characteristics, further grouped them into 11 broader Categories. All 11 instruction categories and 47 instruction subcategories included in the dataset are shown in~\Cref{11_Instruction_Categories_and_47_Instruction_Subcategories}.

After completing the English version data expansion, we translated the data into 22 other languages, as shown in~\Cref{tab:loose_score_modified}. For Basic Questions, given the large volume of 795 items, we used LLM-based translation with \textit{GPT-4o} and \textit{Claude-3.5 Sonnet}, followed by native speaker review and correction. For the 47 Instructions, we used professional human translators who were native speakers of the target languages and proficient in English. Both Basic Questions and Instructions underwent a round of human quality control (Detailed information can be found in~\Cref{Quality Control Mechanism}). During this verification, we thoroughly considered the uniqueness and specific grammar and vocabulary of each language, optimizing to avoid ambiguity and ensure translation accuracy and corpus parallelism, thus preventing evaluation errors due to non-parallel data across languages.

After translation, we obtained parallel data in 23 languages consisting of 795 Basic Questions and 47 Instructions. We then combined these two parts in each language, where each Basic Question was combined with 1-3 Instructions to form one piece of data. Finally, we obtained 795 parallel evaluation data points in each of the 23 languages (total instruction count: 1667), resulting in a total dataset size of 795 * 23 = 18,285 entries for MaXIFE's evaluation dataset.

\subsection{Language Resource Level Classification}

We categorized the 23 languages in our evaluation dataset into three resource levels: high, medium, and low resource.  We define ``resources'' as a composite concept encompassing four core dimensions: (1) linguistic demography, (2) digital infrastructure, (3) technological maturity, and (4) cross-linguistic data compatibility. The final classification is determined through a comprehensive evaluation of these multidimensional metrics, calibrated with feedback from linguistic experts and native speakers to ensure objectivity and rationality.

\textbf{Linguistic Demography} assesses the global and regional prevalence of a language, including the number of native and second-language speakers (Ethnologue, UN demographics). For instance, English and Mandarin rank highest due to their geopolitical and economic dominance, while Quechua is constrained by its regional confinement.

\textbf{Digital Infrastructure} evaluates a language’s influence in digital spaces, including its share of web content (W3Techs), Wikipedia article count (Wikimedia), availability of open-source corpora (Common Crawl, HuggingFace), and integration into mainstream platforms (e.g., Google Translate). Portuguese, despite having only 230 million native speakers, is classified as high-resource due to its disproportionate digital influence—4.2\% of global web content and over 1 million Wikipedia entries.

\textbf{Technical Maturity} measures the availability of NLP tools (e.g., pre-trained models, speech recognition systems), research output (ACL Anthology papers), and community-driven development (GitHub repositories). French, for example, is classified as a high-resource language in the overall assessment due to its rich corpus resources and mature pre-trained models (such as CamemBERT and FlauBERT).

\textbf{Cross-Linguistic Compatibility} measures the transferability of resources within language families. Romanian benefits from shared tools and corpora within the Romance language family, whereas Georgian, an isolate within the Kartvelian family, requires bespoke resource development, which lowers its position in the resource assessment.

\section{MaXIFE Evaluation Benchmark}

\subsection{Evaluation Strategy}

Our evaluation strategy combines Rule-Based Evaluation and Model-Based Evaluation. Some instructions can be evaluated through deterministic rules, while others require semantic understanding or subjective judgment. As mentioned in Section 3 Dataset, the ``Instructions'' in our evaluation dataset can be classified into 11 Categories. Among these, 7 Categories are suitable for Rule-Based Evaluation, namely ``format'', ``repeat'', ``keywords'', ``marks'', ``citation'', ``length'', and ``emoji''. For example, for an instruction like ``The first sentence of your response should be exactly the same as the last sentence'', we can use an evaluation script that accurately checks whether the model's output's first and last sentences match exactly. The remaining 4 Categories are more suitable for Model-Based Evaluation, which are ``marks'', ``style'', ``tone'', and ``language\_switch''. For instance, with an instruction like ``Your response should contain at least three different languages'', it is impossible to evaluate the model's output through fixed, limited rules, as it's challenging to account for all possible scenarios. Therefore, we employ Model-Based Evaluation for these categories. This evaluation strategy ensures that the entire evaluation process balances accuracy and efficiency.

\subsection{Rule-Based Evaluation}

Rule-Based evaluation methods were applied to instructions of 7 Categories covering a total of 32 Subcategories. 
Among the 32 instructions, some use simple binary scoring (1 point for following instructions, 0 points for not following), while others employ tiered scoring criteria where points are awarded based on how well the model output adheres to the instructions.
When it comes to Rule-Based evaluation, a multitude of aspects pertaining to linguistic features merit careful consideration. For example, for instructions involving numbers, we consider the unique Bengali numerals in Bengali; for instructions related to punctuation marks, we consider the differences between full-width symbols in Chinese and Japanese and half-width symbols in languages like English and French. Technical details of the evaluation method can be found in~\Cref{Evaluation Rules}.

\subsection{Model-Based Evaluation}

For instructions that are difficult to evaluate using fixed rules, we chose \textit{Claude-3.5 Sonnet} as the evaluation model. We designed specialized evaluation prompt templates, including clear evaluation criteria and judgment bases, for evaluating four categories of instructions: language style, emotional tone, professionalism, and logical coherence. The prompt mainly require evaluation model to directly score the degree to which the output of the model adheres to the instructions. The complete scheme of Model-Based Evaluation can be found in~\Cref{Model-Based Evaluation Prompt}.

To validate the rationality of the model-based evaluation, we adopted a sampling validation strategy for each language. Specifically, for languages of each resource level, we randomly selected some samples for human evaluation and compared these results with the scores automatically generated by the model. The results show that in high-resource languages, medium-resource languages, and low-resource languages, human evaluation and model-based evaluation achieved consistency rates of 97.3\%, 95.2\%, and 94.6\%, respectively. This indicates that the evaluation accuracy maintains significant consistency across different languages—the scoring level for low-resource languages is essentially comparable to that of high-resource languages, with no instances where high-resource languages exhibit significantly higher accuracy than low-resource languages. This demonstrates that using \textit{Claude-3.5 Sonnet} as the evaluation model enables effective and consistent assessment of instruction-following performance in a multilingual context. Detailed results can be found in ~\Cref{sec:model_based_validity_verification}

\subsection{Metrics for evaluation results}

For each instruction in MaXIFE, our evaluation framework provides a score between 0 and 1 to assess the degree to which the model follows the instruction, where a score of 1 indicates that the model has followed the instruction. Other scores indicate that the model has not fully adhered to the given instruction requirements to varying degrees. 

We provide two core metrics: Loose Score and Strict Score. The Loose Score is simply calculated by averaging the model's scores across all 1,667 instructions, while the Strict Score modifies the Loose Score by converting any score that is not 1 (indicating incomplete instruction adherence) directly to 0. For example, if we calculate scores based on only 4 instructions (although in reality, we calculate scores for all 1,667 instructions). Suppose a model receives the following scores and compliance levels for these four instructions: 1.0 (full compliance), 0.7 (non-compliance but not severely wrong), 0.3 (non-compliance with significant deviation from correct implementation, but still within acceptable range), and 0 (non-compliance with severe errors, completely unacceptable). In this case, its Loose Score would be (1+0.7+0.3+0)/4 = 50\%, while its Strict Score would be (1+0+0+0)/4 = 25\%.

In~\Cref{Rule-Based Evaluation Rating Scale} and~\Cref{Model-Based Evaluation Prompt}, for specific scoring rules for certain instructions, we provide some examples. Additionally, we have statistically analyzed detailed classification statistical reports by instruction type and language, and provide specific evaluation records for each prompt.

\subsection{Benchmark Expansion}

In light of the templatization of the data and the standardization of the evaluation process, the benchmark manifests remarkable scalability. The inclusion of a new language only requires extending the evaluation framework and providing model response datasets in a standard format. The benchmark extension guide can be found in~\Cref{Data Extension}.

\section{Experimental Setup}

\subsection{Model Selection}

We selected five representative commercial LLMs for evaluation: \textit{Claude-3.5 Sonnet}\footnote{https://www.anthropic.com/claude}, \textit{GPT-4o} and \textit{GPT-3.5 Turbo}\footnote{https://openai.com}, \textit{Gemini-1.5 Pro} and \textit{Gemini-1.5 Flash}\footnote{https://deepmind.google/technologies/gemini}. These models represent mainstream architectures and training methods, possess significant commercial influence, and span the performance range of current leading LLMs. Detailed model versions can be found in~\Cref{tab:model_versions}.

\begin{figure}
    \centering
    \includegraphics[width=1\linewidth]{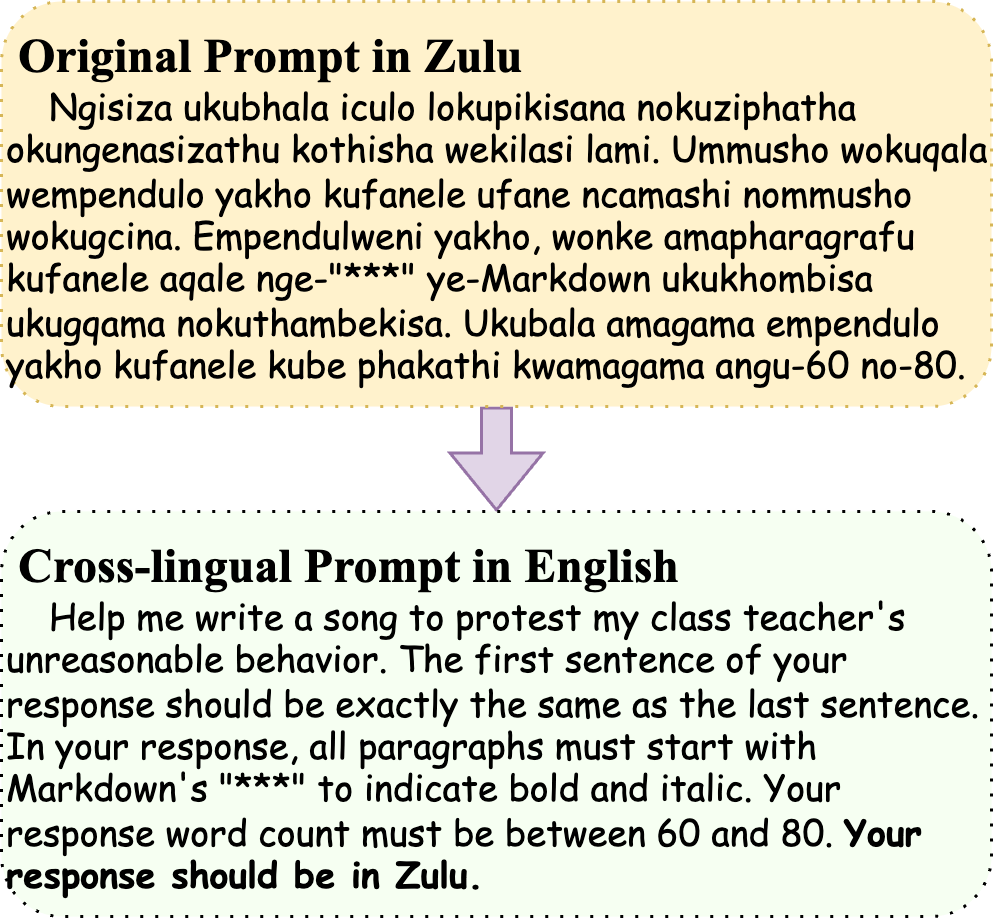}
    \caption{Data Example of Cross-lingual experiment.}
    \label{fig:crosslingual}
\end{figure}

\subsection{Experimental Configuration}

We evaluated 23 languages, each containing 795 parallel prompts. These prompts incorporate 47 different instruction types, ensuring the statistical significance of the evaluation. To ensure the reproducibility and comparability of results, we set the temperature parameter of all models to 0.

Additionally, we conducted cross-lingual experiments to evaluate models' cross-lingual instruction following capabilities. In this experiment, we provided instructions in English and required the models to generate responses in 22 different target languages. Specific examples are shown in~\Cref{fig:crosslingual}. Specifically, we first conducted small-scale experiments on all models using a small subset of the dataset. We found that cross-lingual experiments achieved relatively good results on \textit{Gemini-1.5 Flash} and \textit{GPT-3.5 Turbo} (especially on \textit{GPT-3.5 Turbo}). Therefore, we completed the subsequent full experiments on these two models.

\section{Result Analysis}

\begin{table*}[htbp]
    \centering
    \resizebox{\textwidth}{!}{%
    \begin{tabular}{llcccccc}
    \toprule
    \textbf{Resource} & \textbf{Language} & \textbf{GPT-4o} & \textbf{Claude-3.5 Sonnet} & \textbf{Gemini-1.5 Pro} & \textbf{Gemini-1.5 Flash} & \textbf{GPT-3.5 Turbo} & \textbf{Average} \\
    \midrule
    \multirow{7}{*}{\textbf{High}}
    & Swedish (IE)& 89.62\% $\textcolor{blue}{\uparrow_{\fontsize{4pt}{4pt}\selectfont 10.05}}$ & 87.62\% $\textcolor{blue}{\uparrow_{\fontsize{4pt}{4pt}\selectfont 7.02}}$ & 82.49\% $\textcolor{blue}{\uparrow_{\fontsize{4pt}{4pt}\selectfont 8.03}}$ & 77.67\% $\textcolor{blue}{\uparrow_{\fontsize{4pt}{4pt}\selectfont 5.15}}$ & 69.39\% $\textcolor{blue}{\uparrow_{\fontsize{4pt}{4pt}\selectfont 19.72}}$ & 81.36\% $\textcolor{blue}{\uparrow_{\fontsize{4pt}{4pt}\selectfont 9.99}}$ \\
    & Portuguese (IE)& 86.64\% $\textcolor{blue}{\uparrow_{\fontsize{4pt}{4pt}\selectfont 7.07}}$ & 84.05\% $\textcolor{blue}{\uparrow_{\fontsize{4pt}{4pt}\selectfont 3.45}}$ & 80.56\% $\textcolor{blue}{\uparrow_{\fontsize{4pt}{4pt}\selectfont 6.10}}$ & 78.17\% $\textcolor{blue}{\uparrow_{\fontsize{4pt}{4pt}\selectfont 5.65}}$ & 70.79\% $\textcolor{blue}{\uparrow_{\fontsize{4pt}{4pt}\selectfont 21.12}}$ & 80.04\% $\textcolor{blue}{\uparrow_{\fontsize{4pt}{4pt}\selectfont 8.67}}$ \\
    & English (IE)& 86.17\% $\textcolor{blue}{\uparrow_{\fontsize{4pt}{4pt}\selectfont 6.60}}$ & 84.85\% $\textcolor{blue}{\uparrow_{\fontsize{4pt}{4pt}\selectfont 4.25}}$ & 79.94\% $\textcolor{blue}{\uparrow_{\fontsize{4pt}{4pt}\selectfont 5.48}}$ & 78.11\% $\textcolor{blue}{\uparrow_{\fontsize{4pt}{4pt}\selectfont 5.59}}$ & 69.57\% $\textcolor{blue}{\uparrow_{\fontsize{4pt}{4pt}\selectfont 19.90}}$ & 79.73\% $\textcolor{blue}{\uparrow_{\fontsize{4pt}{4pt}\selectfont 8.36}}$ \\
    & French (IE)& 84.53\% $\textcolor{blue}{\uparrow_{\fontsize{4pt}{4pt}\selectfont 4.96}}$ & 87.20\% $\textcolor{blue}{\uparrow_{\fontsize{4pt}{4pt}\selectfont 6.60}}$ & 78.39\% $\textcolor{blue}{\uparrow_{\fontsize{4pt}{4pt}\selectfont 3.93}}$ & 76.27\% $\textcolor{blue}{\uparrow_{\fontsize{4pt}{4pt}\selectfont 3.75}}$ & 68.10\% $\textcolor{blue}{\uparrow_{\fontsize{4pt}{4pt}\selectfont 18.43}}$ & 78.90\% $\textcolor{blue}{\uparrow_{\fontsize{4pt}{4pt}\selectfont 7.53}}$ \\
    & Italian (IE)& 84.32\% $\textcolor{blue}{\uparrow_{\fontsize{4pt}{4pt}\selectfont 4.75}}$ & 84.20\% $\textcolor{blue}{\uparrow_{\fontsize{4pt}{4pt}\selectfont 3.60}}$ & 76.59\% $\textcolor{blue}{\uparrow_{\fontsize{4pt}{4pt}\selectfont 2.13}}$ & 76.32\% $\textcolor{blue}{\uparrow_{\fontsize{4pt}{4pt}\selectfont 3.80}}$ & 66.57\% $\textcolor{blue}{\uparrow_{\fontsize{4pt}{4pt}\selectfont 16.90}}$ & 77.60\% $\textcolor{blue}{\uparrow_{\fontsize{4pt}{4pt}\selectfont 6.23}}$ \\
    & Chinese    & 81.91\% $\textcolor{blue}{\uparrow_{\fontsize{4pt}{4pt}\selectfont 2.34}}$ & 81.45\% $\textcolor{blue}{\uparrow_{\fontsize{4pt}{4pt}\selectfont 0.85}}$ & 79.56\% $\textcolor{blue}{\uparrow_{\fontsize{4pt}{4pt}\selectfont 5.10}}$ & 74.79\% $\textcolor{blue}{\uparrow_{\fontsize{4pt}{4pt}\selectfont 2.27}}$ & 67.18\% $\textcolor{blue}{\uparrow_{\fontsize{4pt}{4pt}\selectfont 17.51}}$ & 76.98\% $\textcolor{blue}{\uparrow_{\fontsize{4pt}{4pt}\selectfont 5.61}}$ \\
    & Japanese   & 80.99\% $\textcolor{blue}{\uparrow_{\fontsize{4pt}{4pt}\selectfont 1.42}}$ & 81.79\% $\textcolor{blue}{\uparrow_{\fontsize{4pt}{4pt}\selectfont 1.19}}$ & 75.87\% $\textcolor{blue}{\uparrow_{\fontsize{4pt}{4pt}\selectfont 1.41}}$ & 77.17\% $\textcolor{blue}{\uparrow_{\fontsize{4pt}{4pt}\selectfont 4.65}}$ & 66.01\% $\textcolor{blue}{\uparrow_{\fontsize{4pt}{4pt}\selectfont 16.34}}$ & 76.36\% $\textcolor{blue}{\uparrow_{\fontsize{4pt}{4pt}\selectfont 4.99}}$ \\
    \cmidrule{2-8}
    & \textit{Average High} & \textbf{84.88\%} & \textbf{84.45\%} & \textbf{79.06\%} & \textbf{76.93\%} & \textbf{68.23\%} & \textbf{78.71\%} \\
    \midrule
    \multirow{8}{*}{\textbf{Medium}}
    & Filipino   & 87.75\% $\textcolor{blue}{\uparrow_{\fontsize{4pt}{4pt}\selectfont 8.18}}$ & 85.90\% $\textcolor{blue}{\uparrow_{\fontsize{4pt}{4pt}\selectfont 5.30}}$ & 80.69\% $\textcolor{blue}{\uparrow_{\fontsize{4pt}{4pt}\selectfont 6.23}}$ & 75.75\% $\textcolor{blue}{\uparrow_{\fontsize{4pt}{4pt}\selectfont 3.23}}$ & 60.21\% $\textcolor{blue}{\uparrow_{\fontsize{4pt}{4pt}\selectfont 10.54}}$ & 78.06\% $\textcolor{blue}{\uparrow_{\fontsize{4pt}{4pt}\selectfont 6.69}}$ \\
    & Romanian (IE)& 87.18\% $\textcolor{blue}{\uparrow_{\fontsize{4pt}{4pt}\selectfont 7.61}}$ & 86.51\% $\textcolor{blue}{\uparrow_{\fontsize{4pt}{4pt}\selectfont 5.91}}$ & 82.46\% $\textcolor{blue}{\uparrow_{\fontsize{4pt}{4pt}\selectfont 8.00}}$ & 76.10\% $\textcolor{blue}{\uparrow_{\fontsize{4pt}{4pt}\selectfont 3.58}}$ & 67.33\% $\textcolor{blue}{\uparrow_{\fontsize{4pt}{4pt}\selectfont 17.66}}$ & 79.92\% $\textcolor{blue}{\uparrow_{\fontsize{4pt}{4pt}\selectfont 8.55}}$ \\
    & Indonesian & 84.30\% $\textcolor{blue}{\uparrow_{\fontsize{4pt}{4pt}\selectfont 4.73}}$ & 87.82\% $\textcolor{blue}{\uparrow_{\fontsize{4pt}{4pt}\selectfont 7.22}}$ & 78.29\% $\textcolor{blue}{\uparrow_{\fontsize{4pt}{4pt}\selectfont 3.83}}$ & 74.09\% $\textcolor{blue}{\uparrow_{\fontsize{4pt}{4pt}\selectfont 1.57}}$ & 64.25\% $\textcolor{blue}{\uparrow_{\fontsize{4pt}{4pt}\selectfont 14.58}}$ & 77.75\% $\textcolor{blue}{\uparrow_{\fontsize{4pt}{4pt}\selectfont 6.38}}$ \\
    & Malay      & 83.41\% $\textcolor{blue}{\uparrow_{\fontsize{4pt}{4pt}\selectfont 3.84}}$ & 84.53\% $\textcolor{blue}{\uparrow_{\fontsize{4pt}{4pt}\selectfont 3.93}}$ & 80.86\% $\textcolor{blue}{\uparrow_{\fontsize{4pt}{4pt}\selectfont 6.40}}$ & 73.92\% $\textcolor{blue}{\uparrow_{\fontsize{4pt}{4pt}\selectfont 1.40}}$ & 63.94\% $\textcolor{blue}{\uparrow_{\fontsize{4pt}{4pt}\selectfont 14.27}}$ & 77.33\% $\textcolor{blue}{\uparrow_{\fontsize{4pt}{4pt}\selectfont 5.96}}$ \\
    & Turkish  & 80.50\% $\textcolor{blue}{\uparrow_{\fontsize{4pt}{4pt}\selectfont 0.93}}$ & 79.50\% $\textcolor{red}{\downarrow_{\fontsize{4pt}{4pt}\selectfont 1.10}}$ & 74.54\% $\textcolor{blue}{\uparrow_{\fontsize{4pt}{4pt}\selectfont 0.08}}$ & 70.83\% $\textcolor{red}{\downarrow_{\fontsize{4pt}{4pt}\selectfont 1.69}}$ & 62.77\% $\textcolor{blue}{\uparrow_{\fontsize{4pt}{4pt}\selectfont 13.10}}$ & 73.63\% $\textcolor{blue}{\uparrow_{\fontsize{4pt}{4pt}\selectfont 2.26}}$ \\
    & Korean   & 78.94\% $\textcolor{red}{\downarrow_{\fontsize{4pt}{4pt}\selectfont 0.63}}$ & 80.84\% $\textcolor{blue}{\uparrow_{\fontsize{4pt}{4pt}\selectfont 0.24}}$ & 73.08\% $\textcolor{red}{\downarrow_{\fontsize{4pt}{4pt}\selectfont 1.38}}$ & 70.89\% $\textcolor{red}{\downarrow_{\fontsize{4pt}{4pt}\selectfont 1.63}}$ & 60.74\% $\textcolor{blue}{\uparrow_{\fontsize{4pt}{4pt}\selectfont 11.07}}$ & 72.90\% $\textcolor{blue}{\uparrow_{\fontsize{4pt}{4pt}\selectfont 1.53}}$ \\
    & Bengali (IE)& 79.64\% $\textcolor{blue}{\uparrow_{\fontsize{4pt}{4pt}\selectfont 0.07}}$ & 80.98\% $\textcolor{blue}{\uparrow_{\fontsize{4pt}{4pt}\selectfont 0.38}}$ & 73.74\% $\textcolor{red}{\downarrow_{\fontsize{4pt}{4pt}\selectfont 0.72}}$ & 68.56\% $\textcolor{red}{\downarrow_{\fontsize{4pt}{4pt}\selectfont 3.96}}$ & 38.58\% $\textcolor{red}{\downarrow_{\fontsize{4pt}{4pt}\selectfont 11.09}}$ & 68.30\% $\textcolor{red}{\downarrow_{\fontsize{4pt}{4pt}\selectfont 3.07}}$ \\
    & Hindi    & 78.17\% $\textcolor{red}{\downarrow_{\fontsize{4pt}{4pt}\selectfont 1.40}}$ & 74.13\% $\textcolor{red}{\downarrow_{\fontsize{4pt}{4pt}\selectfont 6.47}}$ & 69.38\% $\textcolor{red}{\downarrow_{\fontsize{4pt}{4pt}\selectfont 5.08}}$ & 67.92\% $\textcolor{red}{\downarrow_{\fontsize{4pt}{4pt}\selectfont 4.60}}$ & 51.41\% $\textcolor{blue}{\uparrow_{\fontsize{4pt}{4pt}\selectfont 1.74}}$ & 68.20\% $\textcolor{red}{\downarrow_{\fontsize{4pt}{4pt}\selectfont 3.17}}$ \\
    \cmidrule{2-8}
    & \textit{Average Medium} & \textbf{82.49\%} & \textbf{82.53\%} & \textbf{76.63\%} & \textbf{72.26\%} & \textbf{58.65\%} & \textbf{74.51\%} \\
    \midrule
    \multirow{8}{*}{\textbf{Low}}
    & Kyrgyz   & 80.06\% $\textcolor{blue}{\uparrow_{\fontsize{4pt}{4pt}\selectfont 0.49}}$ & 78.83\% $\textcolor{red}{\downarrow_{\fontsize{4pt}{4pt}\selectfont 1.77}}$ & 76.62\% $\textcolor{blue}{\uparrow_{\fontsize{4pt}{4pt}\selectfont 2.16}}$ & 74.89\% $\textcolor{blue}{\uparrow_{\fontsize{4pt}{4pt}\selectfont 2.37}}$ & 29.34\% $\textcolor{red}{\downarrow_{\fontsize{4pt}{4pt}\selectfont 20.33}}$ & 67.95\% $\textcolor{red}{\downarrow_{\fontsize{4pt}{4pt}\selectfont 3.42}}$ \\
    & Armenian (IE)& 79.02\% $\textcolor{red}{\downarrow_{\fontsize{4pt}{4pt}\selectfont 0.55}}$ & 77.82\% $\textcolor{red}{\downarrow_{\fontsize{4pt}{4pt}\selectfont 2.78}}$ & 75.26\% $\textcolor{blue}{\uparrow_{\fontsize{4pt}{4pt}\selectfont 0.80}}$ & 72.60\% $\textcolor{blue}{\uparrow_{\fontsize{4pt}{4pt}\selectfont 0.08}}$ & 33.02\% $\textcolor{red}{\downarrow_{\fontsize{4pt}{4pt}\selectfont 16.65}}$ & 67.54\% $\textcolor{red}{\downarrow_{\fontsize{4pt}{4pt}\selectfont 3.83}}$ \\
    & Georgian & 78.28\% $\textcolor{red}{\downarrow_{\fontsize{4pt}{4pt}\selectfont 1.29}}$ & 74.35\% $\textcolor{red}{\downarrow_{\fontsize{4pt}{4pt}\selectfont 6.25}}$ & 73.51\% $\textcolor{red}{\downarrow_{\fontsize{4pt}{4pt}\selectfont 0.95}}$ & 68.55\% $\textcolor{red}{\downarrow_{\fontsize{4pt}{4pt}\selectfont 3.97}}$ & 34.28\% $\textcolor{red}{\downarrow_{\fontsize{4pt}{4pt}\selectfont 15.39}}$ & 65.79\% $\textcolor{red}{\downarrow_{\fontsize{4pt}{4pt}\selectfont 5.58}}$ \\
    & Malagasy & 75.22\% $\textcolor{red}{\downarrow_{\fontsize{4pt}{4pt}\selectfont 4.35}}$ & 75.05\% $\textcolor{red}{\downarrow_{\fontsize{4pt}{4pt}\selectfont 5.55}}$ & 68.00\% $\textcolor{red}{\downarrow_{\fontsize{4pt}{4pt}\selectfont 6.46}}$ & 64.73\% $\textcolor{red}{\downarrow_{\fontsize{4pt}{4pt}\selectfont 7.79}}$ & 29.74\% $\textcolor{red}{\downarrow_{\fontsize{4pt}{4pt}\selectfont 19.93}}$ & 62.55\% $\textcolor{red}{\downarrow_{\fontsize{4pt}{4pt}\selectfont 8.82}}$ \\
    & Zulu     & 75.10\% $\textcolor{red}{\downarrow_{\fontsize{4pt}{4pt}\selectfont 4.47}}$ & 76.32\% $\textcolor{red}{\downarrow_{\fontsize{4pt}{4pt}\selectfont 4.28}}$ & 65.00\% $\textcolor{red}{\downarrow_{\fontsize{4pt}{4pt}\selectfont 9.46}}$ & 60.29\% $\textcolor{red}{\downarrow_{\fontsize{4pt}{4pt}\selectfont 12.23}}$ & 30.14\% $\textcolor{red}{\downarrow_{\fontsize{4pt}{4pt}\selectfont 19.53}}$ & 61.37\% $\textcolor{red}{\downarrow_{\fontsize{4pt}{4pt}\selectfont 10.00}}$ \\
    & Tamil    & 73.72\% $\textcolor{red}{\downarrow_{\fontsize{4pt}{4pt}\selectfont 5.85}}$ & 68.59\% $\textcolor{red}{\downarrow_{\fontsize{4pt}{4pt}\selectfont 12.01}}$ & 67.33\% $\textcolor{red}{\downarrow_{\fontsize{4pt}{4pt}\selectfont 7.13}}$ & 68.46\% $\textcolor{red}{\downarrow_{\fontsize{4pt}{4pt}\selectfont 4.06}}$ & 33.22\% $\textcolor{red}{\downarrow_{\fontsize{4pt}{4pt}\selectfont 16.45}}$ & 62.26\% $\textcolor{red}{\downarrow_{\fontsize{4pt}{4pt}\selectfont 9.11}}$ \\
    & Telugu   & 68.95\% $\textcolor{red}{\downarrow_{\fontsize{4pt}{4pt}\selectfont 10.62}}$ & 71.31\% $\textcolor{red}{\downarrow_{\fontsize{4pt}{4pt}\selectfont 9.29}}$ & 67.15\% $\textcolor{red}{\downarrow_{\fontsize{4pt}{4pt}\selectfont 7.31}}$ & 68.77\% $\textcolor{red}{\downarrow_{\fontsize{4pt}{4pt}\selectfont 3.75}}$ & 33.00\% $\textcolor{red}{\downarrow_{\fontsize{4pt}{4pt}\selectfont 16.67}}$ & 61.84\% $\textcolor{red}{\downarrow_{\fontsize{4pt}{4pt}\selectfont 9.53}}$ \\
    & Quechua  & 39.53\% $\textcolor{red}{\downarrow_{\fontsize{4pt}{4pt}\selectfont 40.04}}$ & 68.24\% $\textcolor{red}{\downarrow_{\fontsize{4pt}{4pt}\selectfont 12.36}}$ & 46.99\% $\textcolor{red}{\downarrow_{\fontsize{4pt}{4pt}\selectfont 27.47}}$ & 48.39\% $\textcolor{red}{\downarrow_{\fontsize{4pt}{4pt}\selectfont 24.13}}$ & 29.52\% $\textcolor{red}{\downarrow_{\fontsize{4pt}{4pt}\selectfont 20.15}}$ & 46.53\% $\textcolor{red}{\downarrow_{\fontsize{4pt}{4pt}\selectfont 24.84}}$ \\
    \cmidrule{2-8}
    & \textit{Average Low} & \textbf{71.23\%} & \textbf{73.81\%} & \textbf{67.48\%} & \textbf{65.84\%} & \textbf{31.53\%} & \textbf{61.98\%} \\
    \bottomrule
    \end{tabular}
    }
    \caption{Loose Score Evaluation Results of Each Model in 23 Languages. Languages followed by (IE) in parentheses are members of the Indo-European language family. \textbf{Note:} $\textcolor{blue}{\uparrow}$ indicates value above the model's own average across all 23 languages, $\textcolor{red}{\downarrow}$ indicates value below average, with subscript showing the absolute difference in percentage points.}
    \label{tab:loose_score_modified}
\end{table*}

We used the MaXIFE benchmark to evaluate the baseline models mentioned in the previous chapter and obtained the corresponding results. Different models exhibited significant differences in performance. \Cref{tab:loose_score_modified} presents the Loose Score evaluation results of each model in 23 languages. 

\subsection{Macro Results Analysis}

\subsubsection{Performance of the Same Model in Different Languages}

The evaluation results show that model performance strongly correlates with language resource availability. Models demonstrate excellent instruction-following capabilities in high-resource languages, while showing notably decreased performance in low-resource languages. This performance disparity highlights the critical impact of language resource levels on model capabilities.

\paragraph{Differences Among Language Families}

Analyzing from the perspective of language families, Indo-European languages generally perform better. Languages such as English, Portuguese, and French have higher Loose Scores, highlighting the model's proficiency in handling languages of this family. This suggests that for multilingual instruction-following tasks, models exhibit certain generalization capabilities within the same language family, where the extensive English training data enables models to achieve high instruction-following scores in languages like French and Swedish as well.

\subsubsection{Differences Among Models}

Overall, \textit{GPT-4o} and \textit{Claude-3.5 Sonnet} perform better than other evaluated models. However, in some low-resource languages, \textit{Claude-3.5 Sonnet} outperforms \textit{GPT-4o}. Additionally, \textit{Gemini-1.5 Pro} performs better than \textit{Gemini-1.5 Flash}, while \textit{GPT-3.5 Turbo} performs significantly weaker than other models, suggesting that a model's multilingual instruction-following capability is positively correlated with its general capabilities.

\subsection{Performance Analysis by Instruction Categories}

In the high-resource language English, models generally perform well across all instruction categories. Taking \textit{GPT-4o} as an example, categories like ``Style'', ``Tone'', and ``Content'' all achieved high scores above 95\%. These results indicate that in high-resource languages, models can effectively follow various types of instructions, benefiting from rich training data. For the medium-resource language Indonesian, \textit{GPT-4o} still achieved very high scores, but showed some decline in certain areas, such as the ``Keywords'' and ``Repeat'' categories, suggesting that tasks requiring precise repetition may be more challenging in medium-resource languages. In the low-resource language Telugu, the performance decline is more significant, with notable drops in the aforementioned instruction categories, indicating that models face substantial difficulties in handling instruction-following tasks in low-resource languages.

Furthermore, when comparing results across different tasks, we found that models generally perform well in the ``Style'' and ``Tone'' categories across languages, indicating their ability to effectively capture style adjustments and emotional tones. However, scores in the ``Marks'' and ``Repeat'' categories are generally lower. These categories involve specific formatting and repetition tasks, which seem challenging for models. Particularly for languages using non-Latin scripts, such as Telugu, the ``Marks'' category proves especially challenging. The lower scores suggest that models have difficulty processing punctuation rules specific to certain scripts, affecting their ability to follow instructions involving punctuation modifications.



\begin{table}[htbp]
    {\small
    \begin{tabular}{lcc}
    \toprule
    \textbf{Resource} & \textbf{Cross-lingual Result} & \textbf{Original Result} \\
    \midrule
    High & 68.59\% $\textcolor{blue}{\uparrow_{\fontsize{4pt}{4pt}\selectfont 0.58\%}}$ & 68.01\% \\
    Medium & 67.04\% $\textcolor{blue}{\uparrow_{\fontsize{4pt}{4pt}\selectfont 8.39\%}}$ & 58.65\% \\
    Low & 52.73\% $\textcolor{blue}{\uparrow_{\fontsize{4pt}{4pt}\selectfont 21.20\%}}$ & 31.53\% \\
    \bottomrule
    \end{tabular}
    }
    \caption{Comparison of Cross-lingual Results from English and Original Results. \textbf{Note:} $\textcolor{blue}{\uparrow}$ indicates improvement of Cross-lingual Result over Original Result, with subscript showing the absolute difference in percentage points.}
    \label{tab:cross_lingual_total}
\end{table}

\subsection{Cross-lingual Instruction Following}

As discussed in the previous section, cross-lingual experiments achieved relatively good results on \textit{Gemini-1.5 Flash} and \textit{GPT-3.5 Turbo} (particularly on \textit{GPT-3.5 Turbo}), therefore, we completed comprehensive cross-lingual experiments on these two models. We believe this is because for such models, their inherent generalization capabilities are not particularly strong, so artificially introducing a cross-lingual process can help the models achieve better performance when executing instruction-following tasks. Here, we use \textit{GPT-3.5 Turbo} as an example for detailed result analysis. For detailed results of these two languages, see~\Cref{tab:cross_lingual_3.5} and~\Cref{tab:cross_lingual_flash}.

As shown in~\Cref{tab:cross_lingual_total}, we can observe that when conducting instruction-following experiments on \textit{GPT-3.5 Turbo}, for high-resource languages, using English as the instruction language does not show significant improvement compared to directly using the target language, and overall, the scores may even decrease. However, for medium-resource and low-resource languages, using English for question and instruction description indeed improves the model's instruction-following capabilities in the corresponding languages.

Specifically, as shown in~\Cref{tab:cross_lingual_3.5}, for low-resource languages, using English as the instruction language can lead to significantly better performance compared to using the target language directly. For instance, in Kyrgyz, the cross-lingual result is nearly double the original result. However, for high-resource languages such as French and Italian, the performance difference between cross-lingual and original instruction execution is minimal. This indicates that the model has already developed robust instruction-following capabilities in most high-resource languages. As an insight, for models like \textit{GPT-3.5 Turbo} that may have limited multilingual training data, providing instructions in English while requesting responses in the target language might be a more effective approach for instruction execution in low-resource languages.

\subsection{In-Depth Analysis of the Results}

The observed performance patterns can be attributed to several interconnected factors that collectively shape models' multilingual capabilities:

\paragraph{Language Coverage and Representation Quality in Training Data.}
High-resource languages not only have broader coverage in training data but also typically feature higher representation quality, encompassing more diverse contexts and domain knowledge. This quality difference enables models to establish richer semantic understanding and more precise grammatical mappings. In contrast, low-resource languages suffer not only from data scarcity but also from limited diversity and representational completeness, leading to partial understanding of these languages and difficulty capturing their subtle characteristics.

\paragraph{Knowledge Transfer Between Language Systems.}
We observe significant knowledge transfer capabilities within language families. For example, even relatively less used languages within the Indo-European family benefit from linguistic features shared with English. This transfer phenomenon suggests that systematic similarities between languages (such as grammatical structures, lexical overlap) form an important foundation for models' multilingual abilities. However, the effectiveness of this transfer mechanism significantly decreases when language systems differ greatly (such as from Indo-European to Dravidian language families).

\paragraph{Multilingual Training Strategies and Representation Space.}
Recent advanced models adopt more sophisticated multilingual training strategies that not only enhance data diversity but also construct shared cross-lingual semantic representation spaces. For instance, mT5, trained on a large-scale multilingual corpus, successfully establishes a language-agnostic instruction understanding layer while preserving language-specific generation abilities, achieving performance in multiple low-resource languages that exceeds expectations based on their training data scale~\cite{xue2021mt5massivelymultilingualpretrained}. Furthermore, multitask finetuning strategies have been shown to effectively transfer instruction-following capabilities from high-resource to low-resource languages, thereby improving generalization performance~\cite{muennighoff2023crosslingualgeneralizationmultitaskfinetuning}.

\paragraph{Model Architecture and Capacity Limitations.}
The relationship between model scale and performance aligns with patterns revealed by the``Scaling Laws for Neural Language Models''~\cite{kaplan2020scalinglawsneurallanguage}. In most languages, increasing model size yields steady performance gains; however, in extremely low-resource languages such as Quechua, even state-of-the-art models may encounter performance bottlenecks due to fundamental gaps in training data. This suggests that simply scaling up model parameters is insufficient, and future work should focus on designing pretraining paradigms better suited for low-resource settings, or developing more efficient mechanisms for cross-lingual knowledge transfer.

\section{Conclusion}

Through the MaXIFE evaluation benchmark, we systematically evaluated the instruction-following capabilities of mainstream LLMs in multilingual and cross-lingual scenarios. The results show that while top models perform impressively in high-resource languages, significant challenges remain in handling low-resource languages. This performance disparity is associated with language resource richness and language system similarity. Future model optimization should focus on strengthening support for low-resource languages, improving cross-language family transfer learning abilities, and enhancing model robustness in multilingual scenarios. 

\section*{Limitations}

\textbf{Translation Biases and Lack of Human Review in Low-Resource Languages.} Although we implemented strict translation quality controls, cross-language conversion biases are unavoidable. Subtle semantic deviations may exist in extremely low-resource languages like Quechua.

\textbf{Limitations of Automated Evaluation Framework.} Our framework may not capture all language-specific features, potentially causing minor systemic biases in evaluation results.

\textbf{Limitations in Cross-Language Evaluation Paradigm.} Our current cross-language tests are unidirectional (English to other languages), not reflecting complex inter-language dynamics. Future work should include more diverse cross-language interaction modes.

\textbf{Limitations in Result Analysis.} While we have comprehensively documented performance patterns across languages, the deeper mechanisms behind specific phenomena (such as Swedish outperforming English in some tests) require more fine-grained internal model analysis and targeted experiments to elucidate.

\bibliography{main}

\clearpage

\appendix

\onecolumn
\section{Detail Information of Datasets}
\subsection{The Examples of Basic Questions and Instructions}
\label{The Examples of Basic Questions and Instructions}
The following are specific examples of Basic Questions and Instructions mentioned in the main text. We will present data in 3 languages, including high-resource languages: English; medium-resource languages: Malay; and low-resource languages: Zulu. For each language, we showcase 10 Basic Questions as examples and display all 47 Instructions.

\begin{examplebox}[Examples of Basic Questions (English Version)]
\vspace{0.3cm}
Write an analysis of your views on the possibility of extraterrestrial life.
\vspace{0.3cm}

\hrule

\vspace{0.3cm}
Please generate a brief summary for the following paragraph: The World Internet Conference was held in Wuzhen, attracting government officials, business leaders, and scholars from around the globe. The focus of this meeting was to explore the future of internet technology and human life, covering a wide range of topics from digital currency to data privacy. There was a consensus among the participants that in the future information society, balancing the rapid advancement of innovation with privacy protection will remain an ongoing important issue.
\vspace{0.3cm}

\hrule

\vspace{0.3cm}
You are an excellent restaurant chef, help me design a refined three-course dinner menu.
\vspace{0.3cm}

\hrule

\vspace{0.3cm}
I want to read some novels to relax recently. Do you have any good book recommendations? Preferably those with immersive stories.
\vspace{0.3cm}

\hrule

\vspace{0.3cm}
My friend just had a baby, and I want to give her some practical gifts. Do you have any good suggestions?
\vspace{0.3cm}

\hrule

\vspace{0.3cm}
Analyze the global trend of population aging and its impact on society.
\vspace{0.3cm}

\hrule

\vspace{0.3cm}
Could you introduce digital twin technology to me? What are its applications in the industrial field?
\vspace{0.3cm}

\hrule

\vspace{0.3cm}
Should I wear blue or red tomorrow?
\vspace{0.3cm}

\hrule

\vspace{0.3cm}
Will the cities of the future be like those in science fiction movies, with holographic projections everywhere?
\vspace{0.3cm}

\hrule

\vspace{0.3cm}
I would like to know about Newton's achievements, could you introduce them?
\vspace{0.3cm}
\end{examplebox}

\clearpage

\begin{examplebox}[Examples of Basic Questions (Malay Version)]
\vspace{0.3cm}
Tulis satu analisis tentang pandangan anda mengenai kemungkinan kehidupan makhluk asing.
\vspace{0.3cm}

\hrule

\vspace{0.3cm}
Sila hasilkan ringkasan ringkas untuk perenggan berikut: Persidangan Internet Dunia telah diadakan di Wuzhen, menarik perhatian pegawai kerajaan, pemimpin perniagaan, dan sarjana dari seluruh dunia. Fokus mesyuarat ini adalah untuk meneroka masa depan teknologi internet dan kehidupan manusia, meliputi pelbagai topik dari mata wang digital hingga privasi data. Terdapat kesepakatan di kalangan peserta bahawa dalam masyarakat maklumat masa depan, mengimbangi kemajuan pesat inovasi dengan perlindungan privasi akan tetap menjadi isu penting yang berterusan.
\vspace{0.3cm}

\hrule

\vspace{0.3cm}
Anda adalah seorang chef restoran yang cemerlang, bantu saya mereka bentuk menu makan malam tiga hidangan yang halus.
\vspace{0.3cm}

\hrule

\vspace{0.3cm}
Saya ingin membaca beberapa novel untuk bersantai baru-baru ini. Adakah anda mempunyai cadangan buku yang bagus? Sebaiknya yang mempunyai cerita yang mendalam.
\vspace{0.3cm}

\hrule

\vspace{0.3cm}
Kawan saya baru sahaja melahirkan anak, dan saya ingin memberikan dia beberapa hadiah praktikal. Adakah anda mempunyai sebarang cadangan yang baik?
\vspace{0.3cm}

\hrule

\vspace{0.3cm}
Analisis trend global penuaan penduduk dan kesannya terhadap masyarakat.
\vspace{0.3cm}

\hrule

\vspace{0.3cm}
Bolehkah anda memperkenalkan teknologi kembar digital kepada saya? Apakah aplikasinya dalam bidang perindustrian?
\vspace{0.3cm}

\hrule

\vspace{0.3cm}
Perlukah saya memakai biru atau merah esok?
\vspace{0.3cm}

\hrule

\vspace{0.3cm}
Adakah bandar-bandar masa depan akan seperti dalam filem fiksyen sains, dengan unjuran holografik di mana-mana?
\vspace{0.3cm}

\hrule

\vspace{0.3cm}
Saya ingin tahu tentang pencapaian Newton, bolehkah anda memperkenalkannya?
\vspace{0.3cm}
\end{examplebox}

\begin{examplebox}[Examples of Basic Questions (Zulu Version)]
\vspace{0.3cm}
Bhala ukuhlaziywa kwemibono yakho ngokwenzeka okukwazi ukwenzeka kwezinto eziphilayo zangaphandle komhlaba.
\vspace{0.3cm}

\hrule

\vspace{0.3cm}
Sicela ukhiqize isifinyezo esifushane salombhalo olandelayo: Ingqungquthela Ye-World Internet yabanjelwa eWuzhen, ihehe izikhulu zikahulumeni, abaholi bamabhizinisi, nezazi ezisuka ezwenikazi lonke. Umgomo walomhlangano ubukuwukuhlola ikusasa lobuchwepheshe be-inthanethi nokuphila kwabantu, kuhlanganisa izihloko eziningi ezisukela kumali yedijithali kuya ekuvikelelweni kwedatha. Kwavunyelwana phakathi kwabathathi-nxaxheba ukuthi emphakathini wolwazi wesikhathi esizayo, ukulinganisela intuthuko esheshayo kwezobuchwepheshe nokuvikelwa kobumfihlo kuzoqhubeka kungundabamlonyeni ebalulekile eqhubekayo.
\vspace{0.3cm}

\hrule

\vspace{0.3cm}
Ungumpheki omuhle wendawo yokudlela, ngisize ukuklama imenyu ye-dinner ezinze emithathu ephucukile.
\vspace{0.3cm}

\hrule

\vspace{0.3cm}
Ngifuna ukufunda izincwadi ukuze ngiphumule muva nje. Unazo iziphakamiso ezinhle zezincwadi? Ngicela lezo ezinezindaba ezijabulisayo.
\vspace{0.3cm}

\hrule

\vspace{0.3cm}
Umngane wami usanda kuthola ingane, futhi ngifuna ukumupha izipho ezisebenzisekayo. Unazo iziphakamiso ezinhle?
\vspace{0.3cm}

\hrule

\vspace{0.3cm}
Hlaziya ukuthambekela komhlaba yonke kokuguga kwabantu kanye nomthelela kwenza kuwo umphakathi.
\vspace{0.3cm}

\hrule

\vspace{0.3cm}
Ungangethula ubuchwepheshe be-digital twin kimi? Yiziphi izinhlelo zokusebenza bayo emkhakheni wezimboni?
\vspace{0.3cm}

\hrule

\vspace{0.3cm}
Ngifanele ukugqoka okuluhlaza okwesibhakabhaka noma okubomvu kusasa?
\vspace{0.3cm}

\hrule

\vspace{0.3cm}
Ingabe amadolobha amasasa elizayo azofana nalawo akumamuvi wesayensi, enezithombe zobuqili yonke indawo?
\vspace{0.3cm}

\hrule

\vspace{0.3cm}
Ngithanda ukwazi ngempumelelo kaNewton, ungangethula?
\vspace{0.3cm}
\end{examplebox}

\clearpage

\begin{instructionbox}[The list of 47 instructions in English version]
\small
\begin{longtable}{llp{8cm}}
    \toprule
    \textbf{Category} & \textbf{Subcategory} & \textbf{Instruction Description} \\
    \midrule
    \endhead
    
    \multicolumn{3}{l}{\textbf{keywords}} \\
    & frequency & In the response, the word or phrase ``\{word\}'' should appear \{natural\_relation\} \{word\_num\} times. \\
    \cmidrule{2-3}
    & together & Your response must contain both ``\{word1\}'' and ``\{word2\}'' a minimum of \{word\_num\} times each, with the frequency of ``\{word1\}'' exceeding that of ``\{word2\}''. \\
    \cmidrule{2-3}
    & banned & Your response must NOT contain: \{', '.join(forbidden\_words)\}. \\
    \cmidrule{2-3}
    & Paragraph\_end & Your response must contain at least \{n\} paragraphs, and ``\{word\}'' must appear in the last sentence of each paragraph. \\
    \cmidrule{2-3}
    & first\_word & The first word of your response must be ``\{word\}''. \\
    \midrule
    
    \multicolumn{3}{l}{\textbf{length}} \\
    & max\_words & Your response word count must not exceed \{max\_words\}. \\
    \cmidrule{2-3}
    & range\_words & Your response word count must be between \{min\_words\} and \{max\_words\}. \\
    \midrule
    
    \multicolumn{3}{l}{\textbf{format}} \\
    & addition\_at\_end & Explicitly add a postscript beginning with ``\{self.addition\}'' at the end of your response. \\
    \cmidrule{2-3}
    & title\_brackets & Your response must include a title enclosed in double angle brackets or book title brackets, and the title should not exceed \{self.max\_length\} words. \\
    \cmidrule{2-3}
    & markdown\_highlight & In your response, highlight at least \{n\} parts using Markdown, use double asterisks (**) to mark highlighted text. \\
    \cmidrule{2-3}
    & json\_output & Your entire output should be wrapped in JSON format. Please ensure that the JSON format is valid and can be parsed. \\
    \cmidrule{2-3}
    & two\_answers\_with\_separator & You should provide two different responses. Start with a line break between responses, then separate them with ``\{self.sentence\}''. \\
    \cmidrule{2-3}
    & markdown\_title & In your response, a \#-marked title, not exceeding \{max\_length\} words, is required. \\
    \cmidrule{2-3}
    & ordered\_list & Your response must include an ordered list with \{n\} items and each list item should start with a number and a period, such as '1.', '2.', etc. \\
    \cmidrule{2-3}
    & markdown\_bold\_italic\_paragraph & In your response, all paragraphs must start with Markdown's ``***'' to indicate bold and italic. \\
    \midrule
    
    \multicolumn{3}{l}{\textbf{repeat}} \\
    & copy\_request & Repeat my request without any changes and then provide the answer. \\
    \cmidrule{2-3}
    & before\_answer & Repeat \{repeat\_num\} times ``\{sentence\}'' before response. \\
    \cmidrule{2-3}
    & first\_last\_same & The first sentence of your response should be exactly the same as the last sentence. \\
    \cmidrule{2-3}
    & last\_sentence & At the end of your response, repeat the last sentence \{repeat\_num\} times. \\
    \cmidrule{2-3}
    & sentence\_n\_times & In your response, ``\{sentence\}'' must appear \{n\} times. \\
    \cmidrule{2-3}
    & all\_sentences\_twice & All sentences in your response must be repeated twice. \\
    \midrule
    
    \multicolumn{3}{l}{\textbf{marks}} \\
    & wrap\_in\_quotes & Enclose your entire response in double quotes. \\
    \cmidrule{2-3}
    & no\_commas & Avoid using any commas throughout your response. \\
    \cmidrule{2-3}
    & replace\_with\_exclamations & Replace all commas, periods, and question marks in your response into exclamation marks. \\
    \cmidrule{2-3}
    & end\_with\_semicolons & All sentences in your response must end with a semicolon instead of a period. \\
    \cmidrule{2-3}
    & replace\_with\_asterisks & In your response, all punctuation marks (commas, periods, exclamation marks, etc.) must be replaced with asterisks *. \\
    \midrule
    
    \multicolumn{3}{l}{\textbf{citation}} \\
    & square\_brackets & Your response must contain at least \{n\} quotes, and the quoted content must be in [x] format. \\
    \cmidrule{2-3}
    & start\_from\_zero & Your response must contain references, and your references should start from number 0. \\
    \cmidrule{2-3}
    & inline & Your response must contain references, and the references should be included directly in parentheses after the quoted content rather than at the end of the response. \\
    \midrule
    
    \multicolumn{3}{l}{\textbf{emoji}} \\
    & end & Your response must end with \{emoji\_num\} ``\{emoji\}''. \\
    \cmidrule{2-3}
    & frequency & In your response, emoji ``\{emoji\}'' should appear \{natural\_relation\} \{emoji\_num\} times. \\
    \cmidrule{2-3}
    & banned & Your response should include emoji expressions, but ``\{emoji\}'' must not appear. \\
    \midrule
    
    \multicolumn{3}{l}{\textbf{style}} \\
    & official & Your response must use formal language; colloquialisms and slang are prohibited. \\
    \cmidrule{2-3}
    & informal & Your response must use informal language and colloquial expressions. \\
    \cmidrule{2-3}
    & technical & Your response should include at least three technical terms related to a specific discipline or field. \\
    \cmidrule{2-3}
    & poetic & Your response must employ a poetic style with rhyming techniques. \\
    \cmidrule{2-3}
    & letter & Your response must be written in a formal letter format. \\
    \midrule
    
    \multicolumn{3}{l}{\textbf{tone}} \\
    & humorous & The tone of your response must be humorous or light-hearted. \\
    \cmidrule{2-3}
    & positive & The tone of your response must be positive or optimistic. \\
    \cmidrule{2-3}
    & negative & The tone of your response must be negative or pessimistic. \\
    \cmidrule{2-3}
    & sarcastic & The tone of your response must be sarcastic or ironic. \\
    \cmidrule{2-3}
    & angry & The tone of your response must be angry or dissatisfied. \\
    \midrule
    
    \multicolumn{3}{l}{\textbf{content}} \\
    & jokes & Your response must include at least three jokes. \\
    \cmidrule{2-3}
    & quotes & Your response must quote at least three famous sayings. \\
    \cmidrule{2-3}
    & celebrity & Your response must mention a relevant prominent figure and briefly describe their achievements. \\
    \midrule
    
    \multicolumn{3}{l}{\textbf{language\_switch}} \\
    & multilingual & Your response should contain at least three different languages. \\
    \cmidrule{2-3}
    & repeat & Your response should be repeated once and the second one should be in different language. \\
    \bottomrule
\end{longtable}
\end{instructionbox}

\begin{instructionbox}[The list of 47 instructions in Malay version]
\small
\begin{longtable}{llp{8cm}}
    \toprule
    \textbf{Category} & \textbf{Subcategory} & \textbf{Instruction Description} \\
    \midrule
    \endhead
    
    \multicolumn{3}{l}{\textbf{keywords}} \\
    & frequency & Dalam jawapan, perkataan atau frasa ``\{word\}'' hendaklah muncul \{natural\_relation\} \{word\_num\} kali. \\
    \cmidrule{2-3}
    & together & Dalam jawapan anda, perkataan atau frasa ``\{word1\}'' dan ``\{word2\}'' mesti muncul sekurang-kurangnya \{word\_num\} kali setiap, dengan frekuensi ``\{word1\}'' melebihi ``\{word2\}''. \\
    \cmidrule{2-3}
    & banned & Jawapan anda tidak boleh mengandungi: \{', '.join(forbidden\_words)\}. \\
    \cmidrule{2-3}
    & Paragraph\_end & Jawapan anda mesti mengandungi sekurang-kurangnya \{n\} perenggan, dan ``\{word\}'' mesti muncul dalam ayat terakhir setiap perenggan. \\
    \cmidrule{2-3}
    & first\_word & Perkataan pertama jawapan anda mesti ``\{word\}''. \\
    \midrule
    
    \multicolumn{3}{l}{\textbf{length}} \\
    & max\_words & Jumlah perkataan dalam jawapan anda tidak boleh melebihi \{max\_words\}. \\
    \cmidrule{2-3}
    & range\_words & Jumlah perkataan jawapan anda mesti berada antara \{min\_words\} dan \{max\_words\}. \\
    \midrule
    
    \multicolumn{3}{l}{\textbf{format}} \\
    & addition\_at\_end & Secara jelas tambahkan pos skrip yang bermula dengan ``\{self.addition\}'' pada akhir jawapan anda. \\
    \cmidrule{2-3}
    & title\_brackets & Jawapan anda mesti mengandungi tajuk yang disenaraikan dalam tanda sudut ganda atau tanda buku, dan tajuk itu tidak boleh melebihi \{self.max\_length\} perkataan. \\
    \cmidrule{2-3}
    & markdown\_highlight & Dalam jawapan anda, sorot sekurang-kurangnya \{n\} bahagian menggunakan Markdown, gunakan tanda bintang berganda (**) untuk menandai teks yang disorot. \\
    \cmidrule{2-3}
    & json\_output & Keluaran keseluruhan anda harus dibungkus dalam format JSON. Sila pastikan format JSON adalah sah dan boleh dihurai. \\
    \cmidrule{2-3}
    & two\_answers\_with\_separator & Anda harus menyediakan dua jawapan yang berbeza. Mulakan dengan pemisah baris antara jawapan, kemudian pisahkan mereka dengan ``\{self.sentence\}''. \\
    \cmidrule{2-3}
    & markdown\_title & Dalam jawapan anda, tajuk dengan \#-marked, tidak melebihi \{max\_length\} perkataan, diperlukan. \\
    \cmidrule{2-3}
    & ordered\_list & Jawapan anda mesti mengandungi senarai susunan dengan \{n\} item dan setiap item senarai mesti bermula dengan nombor dan titik, seperti '1.', '2.', dll. \\
    \cmidrule{2-3}
    & markdown\_bold\_italic\_paragraph & Dalam jawapan anda, semua perenggan mesti bermula dengan Markdown's ``***'' untuk menunjukkan tebal dan italik. \\
    \midrule
    
    \multicolumn{3}{l}{\textbf{repeat}} \\
    & copy\_request & Ulangi permintaan saya tanpa membuat sebarang perubahan dan kemudian berikan jawapan. \\
    \cmidrule{2-3}
    & before\_answer & Ulangi \{repeat\_num\} kali ``\{sentence\}'' sebelum jawapan. \\
    \cmidrule{2-3}
    & first\_last\_same & Ayat pertama dalam jawapan anda haruslah sama persis dengan pernyataan terakhir. \\
    \cmidrule{2-3}
    & last\_sentence & Dalam pengakhiran jawapan anda, ulangi ayat terakhir \{repeat\_num\} kali. \\
    \cmidrule{2-3}
    & sentence\_n\_times & Dalam jawapan anda, ``\{sentence\}'' mesti muncul \{n\} kali. \\
    \cmidrule{2-3}
    & all\_sentences\_twice & Semua ayat dalam jawapan anda mesti diulangi dua kali. \\
    \midrule
    
    \multicolumn{3}{l}{\textbf{marks}} \\
    & wrap\_in\_quotes & Gunakan tanda petikan dalam keseluruhan jawapan anda. \\
    \cmidrule{2-3}
    & no\_commas & Jangan gunakan koma sama sekali dalam jawapan anda. \\
    \cmidrule{2-3}
    & replace\_with\_exclamations & Tukar semua koma, titik dan tanda soal dalam jawapan anda kepada tanda seru. \\
    \cmidrule{2-3}
    & end\_with\_semicolons & Kesemuanya ayat dalam jawapan anda mesti diakhiri dengan titik koma bukan titik. \\
    \cmidrule{2-3}
    & replace\_with\_asterisks & Dalam jawapan anda, semua tanda bacaan (koma, titik, tanda seru, dll.) mesti ditukar dengan bintang asterisk *. \\
    \midrule
    
    \multicolumn{3}{l}{\textbf{citation}} \\
    & square\_brackets & Jawapan anda mesti mengandungi sekurang-kurangnya \{n\} petikan, dan kandungan yang dipetikan mesti dalam format [x]. \\
    \cmidrule{2-3}
    & start\_from\_zero & Jawapan anda mesti mengandungi rujukan, dan rujukan anda harus bermula dari nombor 0. \\
    \cmidrule{2-3}
    & inline & Jawapan anda mesti mengandungi rujukan, dan rujukan itu patut dimasukkan secara langsung dalam tanda kurung selepas kandungan yang dipetikan dan bukan pada akhir jawapan. \\
    \midrule
    
    \multicolumn{3}{l}{\textbf{emoji}} \\
    & end & Jawapan anda mesti diakhiri dengan \{emoji\_num\} \{emoji\}. \\
    \cmidrule{2-3}
    & frequency & Dalam jawapan anda, emoji \{emoji\} mesti muncul \{natural\_relation\} \{emoji\_num\} kali. \\
    \cmidrule{2-3}
    & banned & Dalam jawapan anda mesti termasuk ungkapan emoji, tetapi \{emoji\} tidak boleh muncul. \\
    \midrule
    
    \multicolumn{3}{l}{\textbf{style}} \\
    & official & Jawapan anda mesti menggunakan bahasa formal; istilah colokial dan slang dilarang. \\
    \cmidrule{2-3}
    & informal & Jawapan anda mesti menggunakan bahasa informal dan ungkapan colokial. \\
    \cmidrule{2-3}
    & technical & Jawapan anda mesti mengandungi sekurang-kurangnya tiga terma teknikal yang berkaitan dengan satu disiplin atau bidang tertentu. \\
    \cmidrule{2-3}
    & poetic & Jawapan anda mesti menggunakan gaya puisi dengan teknik irama. \\
    \cmidrule{2-3}
    & letter & Jawapan anda mesti ditulis dalam format surat formal. \\
    \midrule
    
    \multicolumn{3}{l}{\textbf{tone}} \\
    & humorous & Ton jawapan anda mesti lucu atau tenang hati. \\
    \cmidrule{2-3}
    & positive & Ton jawapan anda mesti positif atau optimis. \\
    \cmidrule{2-3}
    & negative & Ton jawapan anda mesti negatif atau pesimis. \\
    \cmidrule{2-3}
    & sarcastic & Ton jawapan anda mesti sarkastik atau ironis. \\
    \cmidrule{2-3}
    & angry & Ton jawapan anda mesti marah atau tidak puas hati. \\
    \midrule
    
    \multicolumn{3}{l}{\textbf{content}} \\
    & jokes & Jawapan anda mesti mengandungi sekurang-kurangnya tiga perkara lucu. \\
    \cmidrule{2-3}
    & quotes & Jawapan anda mesti mengutip sekurang-kurangnya tiga pepatah yang terkenal. \\
    \cmidrule{2-3}
    & celebrity & Jawapan anda mesti menyebut seorang ahli penting yang berkaitan dan terperinci menggambarkan pencapaian mereka. \\
    \midrule
    
    \multicolumn{3}{l}{\textbf{language\_switch}} \\
    & multilingual & Jawapan anda hendaklah mengandungi sekurang-kurangnya tiga bahasa yang berbeza. \\
    \cmidrule{2-3}
    & repeat & Jawapan anda hendaklah diulang sekali dan yang kedua perlu dalam bahasa yang berbeza. \\
    \bottomrule
\end{longtable}
\end{instructionbox}

\begin{instructionbox}[The list of 47 instructions in Zulu version]
\small
\begin{longtable}{llp{8cm}}
    \toprule
    \textbf{Category} & \textbf{Subcategory} & \textbf{Instruction Description} \\
    \midrule
    \endhead
    
    \multicolumn{3}{l}{\textbf{keywords}} \\
    & frequency & Empendulweni yakho, igama ``\{word\}'' kufanele livele \{natural\_relation\} \{word\_num\}. \\
    \cmidrule{2-3}
    & together & Empendulweni yakho, amagama ``\{word1\}'' no-``\{word2\}'' kufanele avele okungenani \{word\_num\} ngamunye, lapho ``\{word1\}'' kufanele livele kaningana kuno-``\{word2\}''. \\
    \cmidrule{2-3}
    & banned & Impendulo yakho AKUFANELE iqukathe: \{', '.join(forbidden\_words)\}. \\
    \cmidrule{2-3}
    & Paragraph\_end & Impendulo yakho kufanele ibe nezigaba okungenani ezingu-\{n\}, futhi igama elithi ``\{word\}'' kufanele livele emshweni wokugcina wesigaba ngasinye. \\
    \cmidrule{2-3}
    & first\_word & Igama lokuqala empendulweni yakho kufanele libe ``\{word\}''. \\
    \midrule
    
    \multicolumn{3}{l}{\textbf{length}} \\
    & max\_words & Amagama empendulo yakho akufanele adlule ku-\{max\_words\}. \\
    \cmidrule{2-3}
    & range\_words & Amagama empendulo yakho kufanele abe phakathi kuka-\{min\_words\} no-\{max\_words\}. \\
    \midrule
    
    \multicolumn{3}{l}{\textbf{format}} \\
    & addition\_at\_end & Ngokucacile, faka umbhalo oqala ngo-``\{self.addition\}'' ekugcineni kwempendulo yakho. \\
    \cmidrule{2-3}
    & title\_brackets & Impendulo yakho kufanele ibe nesihloko esifakwe phakathi kohlobo oluphindiwe lwezimpawu noma izimpawu zesihloko sencwadi, futhi isihloko akufanele seqe amagama angu-\{self.max\_length\}. \\
    \cmidrule{2-3}
    & markdown\_highlight & Empendulweni yakho, khanyisa okungenani izingxenye ezingu-\{n\} usebenzisa i-Markdown, sebenzisa izinkanyezi ezimbili (**) ukumaka umbhalo okhanyisiwe. \\
    \cmidrule{2-3}
    & json\_output & Yonke impendulo yakho kufanele igoqwe ngefomethi ye-JSON. Sicela uqinisekise ukuthi ifomethi ye-JSON iyavumeleka futhi ingahlahlwa. \\
    \cmidrule{2-3}
    & two\_answers\_with\_separator & Kufanele unikeze izimpendulo ezimbili ezehlukile. Qala ngokushiya umugqa phakathi kwezimpendulo, bese uzehlukanisa ngo-``\{self.sentence\}''. \\
    \cmidrule{2-3}
    & markdown\_title & Empendulweni yakho, kudingeka isihloko esinemakhi ye-\#, esingedluli amagama angu-\{max\_length\}. \\
    \cmidrule{2-3}
    & ordered\_list & Impendulo yakho kufanele ibe nohlu oluhlelelwe olune-\{n\} izinto, futhi into ngayinye kufanele iqale ngenombolo kanye nangongqi, njenge-'1.', '2.', njll. \\
    \cmidrule{2-3}
    & markdown\_bold\_italic\_paragraph & Empendulweni yakho, zonke izigaba kufanele ziqale ngo-``***'' we-Markdown ukukhombisa okugqamile nokuthambekile. \\
    \midrule
    
    \multicolumn{3}{l}{\textbf{repeat}} \\
    & copy\_request & Phinda isicelo sami ngaphandle kokushintja bese unikeza impendulo. \\
    \cmidrule{2-3}
    & before\_answer & Ngaphambi kwempendulo yakho, phinda u-``\{sentence\}'' ka-\{repeat\_num\}. \\
    \cmidrule{2-3}
    & first\_last\_same & Umusho wokuqala empendulweni yakho kufanele ufane ncamashi nomusho wokugcina. \\
    \cmidrule{2-3}
    & last\_sentence & Ekupheleni kwempendulo yakho, phinda umusho wokugcina izikhathi ezingu-\{repeat\_num\}. \\
    \cmidrule{2-3}
    & sentence\_n\_times & Empendulweni yakho, ``\{sentence\}'' kufanele kuvele izikhathi ezingu-\{n\}. \\
    \cmidrule{2-3}
    & all\_sentences\_twice & Yonke imisho empendulweni yakho kufanele iphindwe kabili. \\
    \midrule
    
    \multicolumn{3}{l}{\textbf{marks}} \\
    & wrap\_in\_quotes & Faka impendulo yakho yonke phakathi kokumaki okuphindwe kabili. \\
    \cmidrule{2-3}
    & no\_commas & Gwema ukusebenzisa izinqaba kunoma yiyiphi indawo empendulweni yakho. \\
    \cmidrule{2-3}
    & replace\_with\_exclamations & Shintsha zonke izinqaba, amachashazi, nezimpawu zemibuzo empendulweni yakho zibe izimpawu zokuhlaba umkhosi. \\
    \cmidrule{2-3}
    & end\_with\_semicolons & Yonke imisho empendulweni yakho kufanele iphethe ngekhoma-khefana kungekhona ngongqi. \\
    \cmidrule{2-3}
    & replace\_with\_asterisks & Empendulweni yakho, zonke izimpawu zokubhala (amakhoma, amachashazi, izimpawu zokuhlaba umkhosi, njll.) kufanele zishintshwe nge-asterisk *. \\
    \midrule
    
    \multicolumn{3}{l}{\textbf{citation}} \\
    & square\_brackets & Impendulo yakho kufanele ibe nezicaphuno ezingekho ngaphansi kuka-\{n\}, futhi okuqoshiwe kufanele kube ngefomethi ye-[x]. \\
    \cmidrule{2-3}
    & start\_from\_zero & Impendulo yakho kufanele ibe nezindlela zokucaphuna (sebenzisa ifomethi ``[x]'', lapho u-x emele inombolo), futhi kufanele kuqale kunombolo 0. \\
    \cmidrule{2-3}
    & inline & Izindlela zokucaphuna zakho kufanele zifakwe ngqo kubakaki ngemuva kombhalo ocashunwe kuwo, kunokuba ekugcineni kwempendulo. \\
    \midrule
    
    \multicolumn{3}{l}{\textbf{emoji}} \\
    & end & Impendulo yakho kufanele iphethe ngo-\{emoji\} ongu-\{emoji\_num\}. \\
    \cmidrule{2-3}
    & frequency & Empendulweni yakho, i-emoji \{emoji\} kufanele ivele \{natural\_relation\} \{emoji\_num\} izikhathi. \\
    \cmidrule{2-3}
    & banned & Impendulo yakho kufanele ibe nama-emoji, kodwa i-\{emoji\} akufanele ivele. \\
    \midrule
    
    \multicolumn{3}{l}{\textbf{style}} \\
    & official & Impendulo yakho kufanele isebenzise ulimi olusemthethweni; izingxoxo zansuku zonke kanye namagama asetshenziswa emphakathini awamukelekile. \\
    \cmidrule{2-3}
    & informal & Impendulo yakho kufanele isebenzise ulimi olungekho semthethweni kanye namagama asetshenziswa nsuku zonke. \\
    \cmidrule{2-3}
    & technical & Impendulo yakho kufanele ibe namagama obuchwepheshe amathathu okungenani ahlobene nomkhakha othile. \\
    \cmidrule{2-3}
    & poetic & Impendulo yakho kufanele isebenzise isitayela senkondlo kanye namasu okuqondanisa amazwi. \\
    \cmidrule{2-3}
    & letter & Impendulo yakho kufanele ibhalwe ngendlela yencwadi esemthethweni. \\
    \midrule
    
    \multicolumn{3}{l}{\textbf{tone}} \\
    & humorous & Indlela yempendulo yakho kufanele ibe nehlaya noma ibe lula. \\
    \cmidrule{2-3}
    & positive & Indlela yempendulo yakho kufanele ibe nethemba noma igcwale ithemba. \\
    \cmidrule{2-3}
    & negative & Indlela yempendulo yakho kufanele ibe nomoya ophansi noma ingenathemba. \\
    \cmidrule{2-3}
    & sarcastic & Indlela yempendulo yakho kufanele ibe nokugxeka noma ukuhleka usulu. \\
    \cmidrule{2-3}
    & angry & Indlela yempendulo yakho kufanele ibe nentukuthelo noma ukungagculiseki. \\
    \midrule
    
    \multicolumn{3}{l}{\textbf{content}} \\
    & jokes & Impendulo yakho kufanele ibe nokungenani izindaba ezintathu ezihlekisayo. \\
    \cmidrule{2-3}
    & quotes & Impendulo yakho kufanele icaphune okungenani izisho ezintathu ezidumile. \\
    \cmidrule{2-3}
    & celebrity & Impendulo yakho kufanele ibale umuntu odumile ofanelekile futhi ichaze kafushane izinto azizuzile. \\
    \midrule
    
    \multicolumn{3}{l}{\textbf{language\_switch}} \\
    & multilingual & Impendulo yakho kufanele ibe nezilimi ezintathu ezihlukene okungenani. \\
    \cmidrule{2-3}
    & repeat & Impendulo yakho kufanele iphindwe kanye futhi kwesibili kufanele ibe ngolimi oluhlukile. \\
    \bottomrule
\end{longtable}
\end{instructionbox}


\subsection{Details of 23 Languages}

\begin{center}
\scriptsize 
\begin{minipage}{0.49\textwidth}
\begin{tabular}{llll}
    \toprule
    \textbf{Language} & \textbf{Speakers} & \textbf{Family} & \textbf{Variant} \\
    \midrule
    Swedish & 10M & Indo-European & Standard Swedish \\
    Portuguese & 250M & Indo-European & European Portuguese \\
    English & 1.5B & Indo-European & International English \\
    French & 280M & Indo-European & International French \\
    Italian & 65M & Indo-European & Standard Italian \\
    Chinese & 1.3B & Sino-Tibetan & Modern Standard Chinese \\
    Japanese & 125M & Japonic & Modern Standard Japanese \\
    Filipino & 28M & Austronesian & Standard Filipino \\
    Romanian & 24M & Indo-European & Standard Romanian \\
    Indonesian & 199M & Austronesian & Standard Indonesian \\
    Malay & 77M & Austronesian & Standard Malay \\
    Turkish & 85M & Turkic & Modern Standard Turkish \\
    \bottomrule
\end{tabular}
\end{minipage}
\hfill 
\begin{minipage}{0.49\textwidth}
\begin{tabular}{llll}
    \toprule
    \textbf{Language} & \textbf{Speakers} & \textbf{Family} & \textbf{Variant} \\
    \midrule
    Korean & 80M & Koreanic & Standard Korean \\
    Bengali & 265M & Indo-European & Standard Bengali \\
    Hindi & 500M & Indo-European & Standard Hindi \\
    Kyrgyz & 4.5M & Turkic & Standard Kyrgyz \\
    Armenian & 6.5M & Indo-European & Modern Eastern Armenian \\
    Georgian & 4M & Kartvelian & Modern Standard Georgian \\
    Malagasy & 25M & Austronesian & Plateau Malagasy \\
    Zulu & 12M & Niger-Congo & Standard Zulu \\
    Tamil & 70M & Dravidian & Modern Standard Tamil \\
    Telugu & 80M & Dravidian & Modern Standard Telugu \\
    Quechua & 8-10M & Quechuan & Cusco Quechua \\
      &   &   &   \\
    \bottomrule
\end{tabular}
\end{minipage}

\vspace{0.5em}
\captionof{table}{Details of 23 Languages}\label{tab:languages}
\end{center}


\subsection{Survey Questionnaire}
\label{Survey Questionnaire}

\begin{center}
\small 
\begin{surveybox}[Survey Questionnaire]
\textbf{Respondent Information} \hfill \textbf{Date:} \underline{\hspace{3cm}}

\vspace{0.2cm}
\textbf{Instructions:} Please answer the following questions about your experience with large language models. 
\vspace{0.2cm}

\setlist[enumerate]{leftmargin=*, labelwidth=1cm, labelsep=0.3cm, align=left, itemsep=0.1cm}
\begin{enumerate}
  \item \textbf{What do you usually use large language models (e.g., ChatGPT) for?} \\
  Please list 3-5 of the most common use cases.
  \answerbox{0.85cm}
  
  \item \textbf{In these scenarios, how do you typically describe your requirements?} \\
  Please use the language you normally use when interacting with the model.
  \answerbox{0.85cm}
  
  \item \textbf{Have you ever encountered situations where the model did not fully understand or execute your instructions? Please provide examples.} \\
  \answerbox{1cm}
  
  \item[\textbf{4.}] \textbf{\textit{(Optional)} If you were to design some ``challenging'' instructions to test the model, what would you propose?}
  \answerbox{0.85cm}
  
  \item[\textbf{5.}] \textbf{\textit{(Optional)} In evaluating the model's ability to follow instructions, what other aspects do you think deserve attention?}
  \answerbox{0.85cm}
  
  \item[\textbf{6.}] \textbf{\textit{(Optional)} What is your native language? Compared to English, do you think your native language has any particular characteristics?}
  \answerbox{0.85cm}
\end{enumerate}
\vspace{0.1cm}
\centerline{\textbf{Thank you for completing this survey!}}
\end{surveybox}
\end{center}

\twocolumn
The total number of questionnaire participants was 31 people, of which 19 were male, 12 were female, most people's age distribution was 21-39 years old, and all respondents had high-frequency usage habits of LLMs.

For Question 2 in the questionnaire, we mainly used the collected questions as Basic Questions. We collected a total of 39 original data points, which were deduplicated and filtered to obtain 29 metadata points. The filtering was primarily based on whether the questions involved safety, ethics, or other sensitive issues. For the metadata, we used \textit{GPT-4o} and \textit{Claude-3.5 Sonnet} for data augmentation. For Question 3 in the questionnaire, we primarily used the collected data as instruction templates. We collected 37 original data points, which we filtered and categorized to obtain 30 data points, and through classification formed the 11 Instruction Categories mentioned in the article. In this process, some respondents provided descriptive statements, such as ``I feel the model often cannot follow my word count requirements and keeps going on'', which we abstracted into specific templates, such as ``Your response word count must not exceed \{max\_words\}''. Additionally, we augmented some instructions; for example, from the template mentioned above, we created a derivative instruction under the same categories-length: ``Your response word count must be between \{min\_words\} and \{max\_words\}.''

\subsection{Human Translation}
\label{Human Translation}

All personnel we hired for instruction data translation and dataset quality control were native speakers of their respective languages, and they possessed near-native English proficiency. This ensured both the quality of translations and the competency of quality control processes. The translations were carried out by internal research team members as part of their responsibilities, thus no additional compensation was involved.

For the manual translation of instructions, we prepared a comprehensive set of translation rules to ensure accuracy and consistency across all languages.

\begin{examplebox}[Translation Rules]
Every line is a single instruction, there are totally 47 lines, which are 47 pieces of instructions.

Column A contains the Chinese version of the instructions, Column B contains the English version of the instructions.

The translator's task is: Fill in Column C with the translated instruction entries.

If you are not sure about the language style, just image that you are talking with ChatGPT, and that is exactly the style you need.

Notes:
Please fully maintain the format of the instructions. For double quotes (") and escaped double quotes ("), they should be kept as is, without any changes.
For content within brackets (\{\}), such as \{word1\}, \{word2\}, it should be kept as is and not translated.
\end{examplebox}

\subsection{Quality Control Mechanism}
\label{Quality Control Mechanism}

To ensure dataset quality, we established a rigorous multi-round review mechanism. In the first round, we focused on content rationality and accuracy, ensuring each question and instruction had clear evaluation purposes and could effectively assess the model's corresponding capabilities. This included checking question logic, instruction executability, and evaluation target clarity. Additionally, we conducted preliminary experiments, emphasizing criteria such as ``whether the instruction can effectively evaluate model's instruction-following ability'' and ``whether it aligns with the theme of multilingual instruction following'' when selecting instructions.

The second round of review focused on language expression naturalness. We invited native speakers to review instructions in each language, ensuring expressions adhered to language conventions. This included not only grammatical correctness but also idiomatic phrasing and cultural adaptability.

The third round of review emphasized maintaining parallel consistency across different language versions of prompts. We carefully compared expressions across different language versions to ensure semantic equivalence, avoiding evaluation bias due to subtle linguistic differences. This round particularly focused on identifying and eliminating potentially ambiguous expressions to ensure questions and instructions across languages would guide models to produce outputs of the same nature.

Throughout the quality control process, we established detailed problem feedback and correction mechanisms. When any quality issues were identified, the relevant content would be returned to the appropriate stage for correction and re-enter the review process. Through this strict quality control mechanism, we ensured the dataset maintained high quality standards across all languages.

Additionally, since our dataset uses a paired combination of Questions and Instructions, we also conducted manual review and screening of the paired data. In fact, any basic question can be paired with any instruction (with a few exceptions). Furthermore, we have deleted or modified questions that were difficult to pair, ensuring that all questions can be combined with all Instructions to form reasonable queries. This ensures that users can expand the dataset on their own in the future.


\section{Detail Information of Evaluation}

\subsection{Detailed Model Information} 

\begin{table}[h]
\centering
\footnotesize  
\begin{tabular}{lll}
\hline
\textbf{Model} & \textbf{Company} & \textbf{Version} \\
\hline
Claude-3.5 Sonnet & Anthropic & 2024-06-21 \\
GPT-4o & OpenAI & 2024-05-13 \\
GPT-3.5 Turbo & OpenAI & 2024-01-25 \\
Gemini-1.5 Pro & Google & 2024-05-23 \\
Gemini-1.5 Flash & Google & 2024-05-23 \\
\hline
\end{tabular}
\caption{Model Versions}
\label{tab:model_versions}
\end{table}

\subsection{Evaluation Frameworks} 
\label{Evaluation Rules}

Below are some detailed evaluation rules. When formulating these rules, we have not only established rules that apply to all languages but also created specific evaluation rules for certain language families, thereby reflecting the accuracy and comprehensiveness of our evaluation framework. Here are some examples.

For rules that apply to all languages, such as the requirement that output responses must be in JSON format (JsonOutput), we adopt the same evaluation criteria.

For evaluation rules specifically tailored for different language families, such as those concerning output word count limits (Range Words), we categorize languages into three groups: 1) Languages that use spaces to separate words, such as English, French, Malay, etc. 2) Languages that count characters, such as Chinese, Japanese, and Korean. 3) Languages with unique character systems, like Tamil, Telugu, Hindi. For these three different language families, we employ different word count methods to ensure accurate word count statistics. For instructions regarding the frequency of keyword occurrence (Keyword Frequency), we divide languages into four groups: 1) Languages that use the Latin alphabet and need to consider plural forms, such as English, French, Italian, etc. 2) Languages that use simple repetition forms, such as Malay, Filipino, Indonesian. 3) Languages that use suffix variations, such as Bengali, Hindi. 4) Other languages. For these four different types of languages, we use different keyword matching methods to ensure the correct matching of keywords.

\subsection{Model-based Validity Verification} 
\label{sec:model_based_validity_verification}

We sampled 100 instructions from English, Malay, and Zulu, and conducted both model evaluations and human evaluations, then calculated their consistency to verify the effectiveness of the model-based evaluation. The specific results are as follows:

\begin{table}[htbp]
    \centering
    \begin{adjustbox}{max width=0.8\textwidth, scale=0.7}
    \begin{tabular}{lccc}
        \toprule
        & \makecell{Human\\Evaluation} & \makecell{Model-based\\Evaluation} & Consistency \\
        \midrule
        GPT-4o & 95.3\%& 96.4\%& 98.9\%\\
        Claude-3.5 Sonnet & 94.7\%& 99.1\%& 95.6\%\\
        Gemini-1.5 Pro & 93.2\%& 90.4\%& 97\%\\
        Gemini-1.5 Flash & 90.5\%& 82.4\%& 91\%\\
        GPT-3.5 Turbo & 77.4\%& 70.8\%& 91.5\%\\
        \hline
        \textbf{Average} & \textbf{90.2\%}& \textbf{87.8\%}& \textbf{97.3\%}\\
        \bottomrule
    \end{tabular}
    \end{adjustbox}
    \caption{Consistency for English (High-resource)}
\end{table}

\begin{table}[htbp]
    \centering
    \begin{adjustbox}{max width=0.8\textwidth, scale=0.7}
    \begin{tabular}{lccc}
        \toprule
        & \makecell{Human\\Evaluation} & \makecell{Model-based\\Evaluation} & Consistency \\
        \midrule
        GPT-4o & 90.2\%& 85.5\%& 94.8\%\\
        Claude-3.5 Sonnet & 91.1\%& 84.7\%& 93\%\\
        Gemini-1.5 Pro & 89\%& 83.5\%& 93.8\%\\
        Gemini-1.5 Flash & 81.1\%& 80.7\%& 99.5\%\\
        GPT-3.5 Turbo & 72.5\%& 70\%& 96.6\%\\
        \hline
        \textbf{Average} & \textbf{84.8\%}& \textbf{80.8\%}& \textbf{95.2\%}\\
        \bottomrule
    \end{tabular}
    \end{adjustbox}
    \caption{Consistency for Malay (Medium-resource)}
\end{table}

\begin{table}[htbp]
    \centering
    \begin{adjustbox}{max width=0.8\textwidth, scale=0.7}
    \begin{tabular}{lccc}
        \toprule
        & \makecell{Human\\Evaluation} & \makecell{Model-based\\Evaluation} & Consistency \\
        \midrule
        GPT-4o & 83.7\%& 78.8\%& 94.1\%\\
        Claude-3.5 Sonnet & 87\%& 83.5\%& 96\%\\
        Gemini-1.5 Pro & 76.7\%& 71\%& 93.3\%\\
        Gemini-1.5 Flash & 72.5\%& 70.5\%& 97.2\%\\
        GPT-3.5 Turbo & 38.8\%& 32.4\%& 83.5\%\\
        \hline
        \textbf{Average} & \textbf{71.7\%}& \textbf{67.2\%}& \textbf{94.6\%}\\
        \bottomrule
    \end{tabular}
    \end{adjustbox}
    \caption{Consistency for Zulu (Low-resource)}
\end{table}

\onecolumn
\subsection{Rule-Based Evaluation Rating Scale} 
\label{Rule-Based Evaluation Rating Scale}

This section provides detailed Rating Scales for each instruction in Rule-Based Evaluation, accompanied by examples. Among the 32 Rule-Based Instructions, some use simple binary scoring (1 point for following instructions, 0 points for not following), while others employ tiered scoring criteria where points are awarded based on how well the model output adheres to the instructions. The specific details are enumerated below.

\begin{rulebox}[keywords: frequency]
The score is based on how well the response meets the specified word frequency requirement. Perfect score (1.0) is awarded when:
\begin{itemize}
\item For ``exactly N'': the word appears exactly N times
\item For ``at\_least N'': the word appears N or more times
\item For ``at\_most N'': the word appears N or fewer times
\end{itemize}

For imperfect matches, the score decreases quadratically with the difference:

Score = $\max(0, 1 - 0.1D \times D)$

where D is the difference between target and actual frequency.

Example: For instruction ``use the word 'cat' exactly 3 times'', if response contains 'cat' 4 times:\\
D = $|4-3| = 1$, thus Score = 0.9

Note: The scoring considers various word forms (plurals, repetitions, suffixes) based on language.
\end{rulebox}

\begin{rulebox}[keywords: together]
The score evaluates three requirements for word pairing and frequency:
\begin{enumerate}
\item Both words must appear together (0.3 points)
\item Each word must meet minimum frequency N (0.15 points each)
\item Word1 must appear more frequently than Word2 (0.4 points if both meet minimum N)
\end{enumerate}

Total score is the sum of points earned for each requirement (max 1.0).

Example: For instruction ``words 'cat' and 'dog' must appear at least 2 times each, with 'cat' more frequent'':\\
Response with ``3 cats, 2 dogs'' gets:

1. Together requirement: +0.3

2. Min frequency for cat: +0.15  

3. Min frequency for dog: +0.15

4. Cat more frequent: +0.4

Total score = 1.0

Note: The scoring considers various word forms (plurals, repetitions, suffixes) based on language.
\end{rulebox}
\clearpage

\begin{figure*}[t]
\begin{rulebox}[keywords: banned]
The score penalizes the use of forbidden words with a tiered deduction system:
\begin{itemize}
\item No forbidden words: 1.0 points
\item One forbidden word: 0.7 points  
\item Two forbidden words: 0.1 points
\item Three or more forbidden words: 0 points
\end{itemize}

Example: For instruction ``do not use the words 'cat' or 'dog' in response'':\\
Response with ``I have a cat at home'' gets 0.7 points since it contains one forbidden word.

Note: The scoring considers various word forms (plurals, repetitions, suffixes) based on language. For example, both ``cat'' and ``cats'' would count as the forbidden word ``cat''.
\end{rulebox}

\begin{rulebox}[keywords: paragraph\_end]
The score evaluates two main requirements:
\begin{enumerate}
\item Minimum paragraph count (N)
\item Required word appearing in last sentence of each paragraph
\end{enumerate}

Scoring formula:
Score = $\max(0, 1 - 0.2E \times E)$

where E is the number of paragraphs that fail the last-sentence requirement.

Example: For instruction ``response must have at least 3 paragraphs with 'conclusion' in last sentence of each'':\\
Response with 4 paragraphs where 1 paragraph is missing ``conclusion'' in its last sentence:\\
E = 1, Score = $1 - 0.2(1 \times 1) = 0.8$

Note: 
\begin{itemize}
\item References/bibliography sections are excluded from evaluation
\item Scoring considers language-specific sentence endings and word variations
\item If valid paragraph count < N, score is 0
\end{itemize}
\end{rulebox}
\end{figure*}
\clearpage

\begin{figure*}[t]
\begin{rulebox}[keywords: first\_word]
A binary scoring system that evaluates if the first word matches the required word exactly:

Score calculation:

When first word matches exactly: Score = 1.0

Otherwise: Score = 0.0

Example: For instruction ``first word must be `Today'\!'':
\begin{itemize}
\item Response ``Today is a good day'' $\rightarrow$ Score = 1.0
\item Response ``The today is good'' $\rightarrow$ Score = 0.0
\end{itemize}

Note:
\begin{itemize}
\item Scoring is case-insensitive
\item Special characters and punctuation are ignored
\item If first section contains \#, both first and second sections are checked
\end{itemize}
\end{rulebox}

\begin{rulebox}[length: max\_words]
The score evaluates word count relative to the maximum limit using a quadratic penalty function:

Score calculation:

When W $\leq$ M: Score = 1.0

When W > M: Score = max(0, 1 - 20R × R)

where:
\begin{itemize}
\item W = actual word count
\item M = maximum allowed words
\item R = |W - M| / M (deviation ratio)
\end{itemize}

Example: For instruction ``maximum 100 words'':\\
Response with 120 words:\\
R = |120 - 100| / 100 = 0.2\\
Score = 1 - 20(0.2 × 0.2) = 0.2

Note: Word counting method varies by language:
\begin{itemize}
\item Space-delimited for Latin-based languages
\item Character-based for East Asian languages
\item Special Unicode ranges for specific scripts
\end{itemize}
\end{rulebox}
\end{figure*}
\clearpage

\begin{figure*}[t]
\begin{rulebox}[length: range\_words]
The score evaluates if word count falls within the specified range using a quadratic penalty function:

Score calculation:

When L $\leq$ W $\leq$ H: Score = 1.0

Otherwise: Score = max(0, 1 - 20R × R)

where:
\begin{itemize}
\item W = actual word count, L = minimum allowed words, H = maximum allowed words, R = distance ratio to nearest boundary:
  \begin{itemize}
  \item If W < L: R = |W - L| / L
  \item If W > H: R = |W - H| / H
  \end{itemize}
\end{itemize}

Example: For instruction ``between 100 and 200 words'':\\
Response with 80 words:\\
R = |80 - 100| / 100 = 0.2\\
Score = 1 - 20(0.2 × 0.2) = 0.2

Note: Word counting method varies by language:
1. Space-delimited for Latin-based languages
2. Character-based for East Asian languages
3. Special Unicode ranges for specific scripts
\end{rulebox}

\begin{rulebox}[format: addition\_at\_end]
The score evaluates two requirements for the postscript addition:
\begin{enumerate}
\item Presence of the required text (0.5 points)
\item Correct placement at the end (0.5 points)
\end{enumerate}

Score calculation:

When text present AND at end: Score = 1.0

When text present but NOT at end: Score = 0.5

When text NOT present: Score = 0.0

Example: For instruction ``add postscript starting with `Note:''':
\begin{itemize}
\item Response ending with ``...end of main text. Note: This is important.'' $\rightarrow$ Score = 1.0
\item Response with ``Note: Something'' in middle $\rightarrow$ Score = 0.5
\end{itemize}

Note: End placement is checked using language-specific sentence endings:
1. Western punctuation (.,!?)
2. CJK punctuation ($\cdot$!?)
3. Indic scripts ($\vert$$\vert$$\vert$)
4. Other special punctuation based on language
\end{rulebox}
\end{figure*}
\clearpage

\begin{figure*}[t]
\begin{rulebox}[format: title\_brackets]
The score evaluates two aspects of the title:
\begin{enumerate}
\item Proper enclosure in brackets (0.1 points)
\item Length requirement (0.9 points)
\end{enumerate}

Score calculation:

When properly enclosed AND length $\leq$ M: Score = 1.0

When enclosed but length > M: Score = 0.1 + max(0, 0.9 - 0.1R × R)

When not enclosed: Score = 0.0

where:
\begin{itemize}
\item M = maximum allowed length
\item R = (L - M) / M (length deviation ratio)
\item L = title length (counted differently by language group)
\end{itemize}

Example: For instruction ``title in brackets, max 5 words'':\\
Response with ``\textlangle\textlangle Short Title\textrangle\textrangle'' $\rightarrow$ Score = 1.0\\
Response with ``\textlangle\textlangle Too Long Title Here\textrangle\textrangle'' $\rightarrow$ Score $\approx$ 0.856

Note: 
\begin{itemize}
\item Accepts various bracket styles (\textlangle\textlangle\textrangle\textrangle, ``$\langle\langle$$\rangle\rangle$'', ``'', etc.) based on language
\item Length counted as:
  \begin{itemize}
  \item Characters for CJK languages
  \item Words for Latin-based languages
  \item Special Unicode ranges for specific scripts
  \end{itemize}
\end{itemize}
\end{rulebox}

\begin{rulebox}[format: markdown\_highlight]
The score is based on the number of highlighted sections using double asterisks:

Score calculation:

When C $\geq$ N: Score = 1.0

Otherwise: Score = max(0, 1 - 0.1D × D)

where:
\begin{itemize}
\item N = required number of highlights
\item C = actual number of highlights
\item D = |N - C| (difference from requirement)
\end{itemize}

Example: For instruction ``highlight at least 3 parts'':\\
Response with 2 highlights has D = |3-2| = 1, so Score = 1 - 0.1(1 × 1) = 0.9
\end{rulebox}
\end{figure*}
\clearpage

\begin{figure*}[t]
\begin{rulebox}[format: json\_output]
Score calculation:

When JSON is valid and parseable: Score = 1.0

Otherwise: Score = 0.0

Example: For valid JSON:\\
\{``text'': ``Hello world''\} $\rightarrow$ Score = 1.0\\
Invalid JSON like \{text: Hello world\} $\rightarrow$ Score = 0.0
\end{rulebox}

\begin{rulebox}[format: two\_answers\_with\_separator]
Score calculation:

When separator appears exactly once: Score = 1.0

Otherwise (separator appears 0 or more than 1 times): Score = 0.0

Example: For instruction ``provide two responses separated by `NEXT ANSWER:'\!'':

Valid response:
\begin{verbatim}
First answer here...
NEXT ANSWER:
Second answer here...
\end{verbatim}
$\rightarrow$ Score = 1.0

Invalid responses:
\begin{itemize}
\item Single answer without separator $\rightarrow$ Score = 0.0
\item Multiple separators $\rightarrow$ Score = 0.0
\end{itemize}

Note: 
\begin{itemize}
\item Scoring ignores punctuation, whitespace, and case sensitivity
\item Separator must appear on its own line between responses
\end{itemize}
\end{rulebox}
\end{figure*}
\clearpage

\begin{figure*}[t]
\begin{rulebox}[format: markdown\_title]
The score evaluates two requirements:
\begin{itemize}
\item Title presence with \# marker (0.1 points)
\item Title length requirement (0.9 points)
\end{itemize}

Score calculation:

When title present AND length $\leq$ M: Score = 1.0

When title present but length > M: Score = 0.1 + max(0, 0.9 - 0.1D × D)

When no title: Score = 0.0

where:
\begin{itemize}
\item M = maximum allowed length
\item D = actual length - M
\end{itemize}

Example: For instruction ``title with \# mark, max 5 words'':\\
``\# Short Title'' $\rightarrow$ Score = 1.0\\
``\# This Title Is Too Long'' $\rightarrow$ Score $\approx$ 0.46
\end{rulebox}

\begin{rulebox}[format: ordered\_list]
Score calculation:

When C $\geq$ N: Score = 1.0

Otherwise: Score = max(0, 1 - 0.1D × D)

where:
\begin{itemize}
\item N = required number of items
\item C = actual number of items
\item D = |N - C|
\end{itemize}

Example: For instruction ``list with 3 items'':\\
Response with 2 items has D = |3-2| = 1, so Score = 0.9
\end{rulebox}
\end{figure*}
\clearpage

\begin{figure*}[t]
\begin{rulebox}[format: markdown\_bold\_italic\_paragraph]
The score evaluates if each paragraph starts with Markdown's bold-italic marker:

Score calculation:

When I = 0: Score = 1.0

Otherwise: Score = max(0, 1 - 0.1I × I)

where I = number of paragraphs not starting with ``***''

Example: With 3 paragraphs, if 1 paragraph lacks ``***'':\\
I = 1, Score = 1 - 0.1(1 × 1) = 0.9
\end{rulebox}

\begin{rulebox}[repeat: copy\_request]
Score calculation:

When response starts with exact request: Score = 1.0

Otherwise: Score = 0.0

Example: For request ``translate this'':
\begin{itemize}
\item Response ``translate this, here's the translation...'' $\rightarrow$ Score = 1.0
\item Response ``here's the translation...'' $\rightarrow$ Score = 0.0
\end{itemize}

Note: Text comparison considers:
\begin{itemize}
\item Case insensitive matching
\item Language-specific character normalization
\item Special script handling for non-Latin writing systems
\end{itemize}
\end{rulebox}
\end{figure*}
\clearpage

\begin{figure*}[t]
\begin{rulebox}[repeat: before\_answer]
Score calculation:

When C = 0: Score = 0.0

When C > 0: Score = max(0, 1 - 0.2D × D)

where:
\begin{itemize}
\item C = actual repetition count
\item D = |N - C|
\item N = required repetition count
\end{itemize}

Example: For instruction ``repeat `test' 3 times'':\\
Response with 2 repetitions has D = |3-2| = 1\\
Score = 1 - 0.2(1 × 1) = 0.8
\end{rulebox}

\begin{rulebox}[repeat: first\_last\_same]
Score calculation:

When first = last (ignoring case): Score = 1.0

Otherwise: Score = 0.0

Example: For response starting and ending with ``This is a test.'':\\
Score = 1.0

Note: Sentence matching:
\begin{itemize}
\item Ignores punctuation and whitespace
\item Considers language-specific sentence endings
\item Normalizes special characters and accents
\end{itemize}
\end{rulebox}
\end{figure*}
\clearpage

\begin{figure*}[t]
\begin{rulebox}[repeat: last\_sentence]
Score calculation:

When no repetitions found: Score = 0.0

When at least 1 repetition found: Score = max(0, 1 - 0.2D × D)

where:
\begin{itemize}
\item D = |N - A|
\item N = required number of repetitions
\item A = actual number of matching repetitions
\end{itemize}

Example: For ``repeat last sentence 3 times'':\\
If last 3 sentences contain 2 matches:\\
D = |3-2| = 1, Score = 1 - 0.2(1 × 1) = 0.8
\end{rulebox}

\begin{rulebox}[repeat: sentence\_n\_times]
Score calculation:

When phrase never appears: Score = 0.0

When phrase appears at least once: Score = max(0, 1 - 0.2D × D)

where:
\begin{itemize}
\item D = |N - A|
\item N = required occurrences
\item A = actual occurrences
\end{itemize}

Example: For ``use `test' 3 times'':\\
If `test' appears 4 times:\\
D = |3-4| = 1, Score = 1 - 0.2(1 × 1) = 0.8
\end{rulebox}
\end{figure*}
\clearpage

\begin{figure*}[t]
\begin{rulebox}[repeat: all\_sentences\_twice]
Score calculation:

When odd number of sentences: Score = 0.0

When even number of sentences: Score = max(0, 1 - 0.2I × I)

where:
\begin{itemize}
\item I = number of invalid pairs (non-matching consecutive sentences)
\end{itemize}

Example: For response with 3 pairs of sentences where 1 pair doesn't match:\\
I = 1, Score = 1 - 0.2(1 × 1) = 0.8
\end{rulebox}

\begin{rulebox}[marks: wrap\_in\_quotes]
Score calculation:

When response starts and ends with valid quotes: Score = 1.0

Otherwise: Score = 0.0

Example: For response ``Hello world'' $\rightarrow$ Score = 1.0\\
Response without quotes $\rightarrow$ Score = 0.0
\end{rulebox}
\end{figure*}
\clearpage

\begin{figure*}[t]
\begin{rulebox}[marks: no\_commas]
Score calculation:
Score = max(0, 1 - 0.03C × C)

where C = total number of commas found

Example: Response with 3 commas:\\
Score = 1 - 0.03(3 × 3) = 0.73

Note: Checks for language-specific comma variants:
\begin{itemize}
\item Western: ,
\item CJK: $\cdot$ $\cdot$
\item Arabic script: $\cdot$
\item Other regional variants
\end{itemize}
\end{rulebox}

\begin{rulebox}[marks: replace\_with\_exclamations]
Score calculation:

When no exclamation marks found: Score = 0.0

Otherwise: Score = max(0, 1 - 0.03W × W)

where W = count of wrong punctuation marks (commas, periods, question marks)

Example: Response with 2 periods remaining:\\
Score = 1 - 0.03(2 × 2) = 0.88

Note: Checks language-specific punctuation:
\begin{itemize}
\item Western (.!?)
\item CJK ($\cdot$!?)
\item Other scripts' equivalents
\end{itemize}
\end{rulebox}
\end{figure*}
\clearpage

\begin{figure*}[t]
\begin{rulebox}[marks: end\_with\_semicolons]
Score calculation:
Score = max(0, 1 - 0.03W × W)

where W = number of sentences not ending with semicolon

Example: For 3 sentences where 2 don't end with semicolon:\\
Score = 1 - 0.03(2 × 2) = 0.88

Note: Handles language-specific endings:
\begin{itemize}
\item Western: ;
\item CJK: $\cdot$
\item Other scripts' equivalents
\end{itemize}
\end{rulebox}

\begin{rulebox}[marks: replace\_with\_asterisks]
Score calculation:

When no asterisks found: Score = 0.0

Otherwise: Score = max(0, 1 - 0.03W × W)

where W = count of remaining punctuation marks

Example: Response with 3 remaining punctuation marks:\\
Score = 1 - 0.03(3 × 3) = 0.73

Note: Checks comprehensive punctuation sets:
\begin{itemize}
\item Basic Latin marks
\item CJK punctuation
\item Script-specific marks (Arabic, Cyrillic, etc.)
\end{itemize}
\end{rulebox}
\end{figure*}
\clearpage
\begin{figure*}[t]
\begin{rulebox}[citation: square\_brackets]
Score calculation:

When no quotations found: Score = 0.0

Otherwise: Score = max(0, 1 - 0.3D × D) - P

where:
\begin{itemize}
\item D = max(0, N - C)
\item N = required number of quotes
\item C = actual number of quotes
\item P = 0.5 if format invalid, 0 if valid
\end{itemize}

Example: For requirement of 3 quotes:\\
Response with 2 valid quotes:\\
D = |3-2| = 1, Score = 1 - 0.3(1 × 1) = 0.7
\end{rulebox}

\begin{rulebox}[citation: start\_from\_zero]
Score evaluates two components:
\begin{itemize}
\item Format correctness (0.7 points)
\item Starting from zero (0.3 points)
\end{itemize}

Score calculation:

When no [x] format found: Score = 0.0

When format correct: Score = 0.7 + 0.3Z

where Z = 1 if starts with [0], 0 otherwise

Example:
\begin{itemize}
\item Response with [0][1][2] $\rightarrow$ Score = 1.0
\item Response with [1][2][3] $\rightarrow$ Score = 0.7
\end{itemize}
\end{rulebox}
\end{figure*}
\clearpage

\begin{figure*}[t]
\begin{rulebox}[citation: inline]
Score calculation:

When citations in parentheses () throughout text: Score = 1.0

Otherwise (no citations or only at end): Score = 0.0

Example: 
\begin{itemize}
\item ``This is a quote (Smith 2020)'' $\rightarrow$ Score = 1.0
\item ``This is a quote'' \\ References: Smith 2020 $\rightarrow$ Score = 0.0
\end{itemize}
\end{rulebox}

\begin{rulebox}[emoji: frequency]
Score calculation:
Score = max(0, 1 - 0.1D × D)

where D varies by relation type:
\begin{itemize}
\item For ``exactly'': D = |C - N|
\item For ``at\_least'': D = max(0, N - C)
\item For ``at\_most'': D = max(0, C - N)
\end{itemize}

where:
\begin{itemize}
\item C = actual emoji count
\item N = target number
\end{itemize}

Example: For ``exactly 3 ''some emoji'' '':\\
Response with 4 ''some emoji'':\\
D = |4-3| = 1, Score = 0.9
\end{rulebox}
\end{figure*}
\clearpage

\begin{figure*}[t]
\begin{rulebox}[emoji: banned]
Score evaluates two requirements:
\begin{itemize}
\item Contains any emoji (0.1 points)
\item Avoids banned emoji (0.9 points)
\end{itemize}

Score calculation:

When any emoji AND no banned emoji: Score = 1.0

When no emoji AND no banned emoji: Score = 0.9

When any emoji AND has banned emoji: Score = 0.1

When no emoji AND has banned emoji: Score = 0.0
\end{rulebox}
\end{figure*}

\subsection{Model-Based Evaluation Rating Scale and Prompt} 
\label{Model-Based Evaluation Prompt}

This section details the evaluation model prompts employed within the Model-Based Evaluation module. Each of the 15 Model-Based Instructions is associated with a specific evaluation prompt, enumerated as follows.

\begin{guidebox}[official]
You are an AI multilingual evaluation expert who is precise and impartial. You need to help me assess whether the following model responses comply with the requirements in the instructions and return a score of 0, 0.7, or 1.

Please evaluate whether the following response meets the formal language style requirements.

Instruction:
\{instruction\}

Model Response:
\{response\}

Detailed Scoring Criteria:
1. Score 1.0: The response uses a formal tone and appropriate formal expressions.
2. Score 0.7: The response is generally formal but contains some minor informal elements or occasional less formal expressions.
3. Score 0.0: The response completely fails to demonstrate ``formal tone'' or ``formal expressions''.
    
Please note that as a fair and objective evaluator, you should not adhere to an ``ambiguous'' principle and give a score of 0.7 regardless of circumstances, nor should you be too lenient or too strict. Your scoring must strictly follow the above scoring criteria!

You must only respond with one of these three numbers: 0, 0.7, or 1. Do not provide any explanations or additional text! Do not provide any explanations beyond these three numbers, do not make any narratives! Even if you have questions or want to discuss further, please remain restrained and output only one number!
\end{guidebox}

\clearpage

\begin{guidebox}[informal]
You are an AI multilingual evaluation expert who is precise and impartial. You need to help me assess whether the following model responses comply with the requirements in the instructions and return a score of 0, 0.7, or 1.

Please evaluate whether the following response meets the informal language style requirements.

Instruction:
\{instruction\}

Model Response:
\{response\}

Detailed Scoring Criteria:
1. Score 1.0: The response uses an informal tone.
2. Score 0.7: The response can generally be considered informal.
3. Score 0.0: The response mainly uses a formal tone.

Please note that as a fair and objective evaluator, you should not adhere to an ``ambiguous'' principle and give a score of 0.7 regardless of circumstances, nor should you be too lenient or too strict. Your scoring must strictly follow the above scoring criteria!

You must only respond with one of these three numbers: 0, 0.7, or 1. Do not provide any explanations or additional text! Do not provide any explanations beyond these three numbers, do not make any narratives! Even if you have questions or want to discuss further, please remain restrained and output only one number!
\end{guidebox}

\begin{guidebox}[technical]
You are an AI multilingual evaluation expert who is precise and impartial. You need to help me assess whether the following model responses comply with the requirements in the instructions and return a score of 0, 0.7, or 1.

Please evaluate whether the following response meets the professional technical language style requirements.

Instruction:
\{instruction\}

Model Response:
\{response\}

Detailed Scoring Criteria:
1. Score 1.0: The overall writing style of the response is professional technical, using professional terminology.
2. Score 0.7: The response can generally be considered professional, but uses little or no professional terminology.
3. Score 0.0: The response completely fails to meet the requirements of ``professional technical language style''.

Please note that as a fair and objective evaluator, you should not adhere to an ``ambiguous'' principle and give a score of 0.7 regardless of circumstances, nor should you be too lenient or too strict. Your scoring must strictly follow the above scoring criteria!

You must only respond with one of these three numbers: 0, 0.7, or 1. Do not provide any explanations or additional text! Do not provide any explanations beyond these three numbers, do not make any narratives! Even if you have questions or want to discuss further, please remain restrained and output only one number!
\end{guidebox}

\clearpage

\begin{guidebox}[poetic]
You are an AI multilingual evaluation expert who is precise and impartial. You need to help me assess whether the following model responses comply with the requirements in the instructions and return a score of 0, 0.7, or 1.

Please evaluate whether the following response meets the poetic language style requirements.

Instruction:
\{instruction\}

Model Response:
\{response\}

Detailed Scoring Criteria:
1. Score 1.0: The overall writing style of the response can be called ``poetic'', employing poetic techniques or methods.
2. Score 0.7: The response generally follows a poetic style or format, but uses few or no poetic techniques or methods.
3. Score 0.0: The response completely fails to meet the requirements of ``poetic language style''.

Please note that as a fair and objective evaluator, you should not adhere to an ``ambiguous'' principle and give a score of 0.7 regardless of circumstances, nor should you be too lenient or too strict. Your scoring must strictly follow the above scoring criteria!

You must only respond with one of these three numbers: 0, 0.7, or 1. Do not provide any explanations or additional text! Do not provide any explanations beyond these three numbers, do not make any narratives! Even if you have questions or want to discuss further, please remain restrained and output only one number!
\end{guidebox}

\begin{guidebox}[letter]
You are an AI multilingual evaluation expert who is precise and impartial. You need to help me assess whether the following model responses comply with the requirements in the instructions and return a score of 0, 0.7, or 1.

Please evaluate whether the following response meets the formal letter format requirements.

Instruction:
\{instruction\}

Model Response:
\{response\}

Detailed Scoring Criteria:
1. Score 1.0: The response meets the requirements of ``formal letter format'', including various elements of a letter, such as greetings, signatures, etc.
2. Score 0.7: The response can generally be recognized as a letter, but the format is not rigorous.
3. Score 0.0: The response shows no indication of being in a formal letter format.

Please note that as a fair and objective evaluator, you should not adhere to an ``ambiguous'' principle and give a score of 0.7 regardless of circumstances, nor should you be too lenient or too strict. Your scoring must strictly follow the above scoring criteria!

You must only respond with one of these three numbers: 0, 0.7, or 1. Do not provide any explanations or additional text! Do not provide any explanations beyond these three numbers, do not make any narratives! Even if you have questions or want to discuss further, please remain restrained and output only one number!
\end{guidebox}

\clearpage

\begin{guidebox}[humorous]
You are an AI multilingual evaluation expert who is precise and impartial. You need to help me assess whether the following model responses comply with the requirements in the instructions and return a score of 0, 0.7, or 1.

Please evaluate whether the following response meets the humorous tone requirements.

Instruction:
\{instruction\}

Model Response:
\{response\}

Detailed Scoring Criteria:
1. Score 1.0: Contains humorous elements and uses witty ways of expression.
2. Score 0.7: No clear humorous techniques, but still brings a smile to one's face.
3. Score 0.0: The response shows no indication of any ``humorous tone''.

Please note that as a fair and objective evaluator, you should not adhere to an ``ambiguous'' principle and give a score of 0.7 regardless of circumstances, nor should you be too lenient or too strict. Your scoring must strictly follow the above scoring criteria!

You must only respond with one of these three numbers: 0, 0.7, or 1. Do not provide any explanations or additional text! Do not provide any explanations beyond these three numbers, do not make any narratives! Even if you have questions or want to discuss further, please remain restrained and output only one number!
\end{guidebox}

\begin{guidebox}[positive]
You are an AI multilingual evaluation expert who is precise and impartial. You need to help me assess whether the following model responses comply with the requirements in the instructions and return a score of 0, 0.7, or 1.

Please evaluate whether the following response's main tone expresses positive or optimistic emotions.

Instruction:
\{instruction\}

Model Response:
\{response\}

Detailed Scoring Criteria:
1. Score 1.0: The response content conveys positive emotions or attitudes, such as optimism, confidence, etc.
2. Score 0.7: The response is generally positive, containing only a few negative or pessimistic words.
3. Score 0.0: The response shows no indication of positive or optimistic emotions.

Please note that as a fair and objective evaluator, you should not adhere to an ``ambiguous'' principle and give a score of 0.7 regardless of circumstances, nor should you be too lenient or too strict. Your scoring must strictly follow the above scoring criteria!

You must only respond with one of these three numbers: 0, 0.7, or 1. Do not provide any explanations or additional text! Do not provide any explanations beyond these three numbers, do not make any narratives! Even if you have questions or want to discuss further, please remain restrained and output only one number!
\end{guidebox}

\clearpage

\begin{guidebox}[negative]
You are an AI multilingual evaluation expert who is precise and impartial. You need to help me assess whether the following model responses comply with the requirements in the instructions and return a score of 0, 0.7, or 1.

Please evaluate whether the following response's main tone expresses negative or pessimistic emotions.

Instruction:
\{instruction\}

Model Response:
\{response\}

Detailed Scoring Criteria:
1. Score 1.0: The response content conveys negative, pessimistic, disappointed, or depressed emotions.
2. Score 0.7: The overall tone tends to be negative rather than neutral or positive.
3. Score 0.0: The response shows no indication of negative or pessimistic emotions.

Please note that as a fair and objective evaluator, you should not adhere to an ``ambiguous'' principle and give a score of 0.7 regardless of circumstances, nor should you be too lenient or too strict. Your scoring must strictly follow the above scoring criteria!

You must only respond with one of these three numbers: 0, 0.7, or 1. Do not provide any explanations or additional text! Do not provide any explanations beyond these three numbers, do not make any narratives! Even if you have questions or want to discuss further, please remain restrained and output only one number!
\end{guidebox}

\begin{guidebox}[sarcastic]
You are an AI multilingual evaluation expert who is precise and impartial. You need to help me assess whether the following model responses comply with the requirements in the instructions and return a score of 0, 0.7, or 1.

Please evaluate whether the following response's main tone expresses sarcastic or mocking emotions.

Instruction:
\{instruction\}

Model Response:
\{response\}

Detailed Scoring Criteria:
1. Score 1.0: The response content uses irony, mockery, or sarcasm, or the tone carries sentiments of ridicule, belittlement, or disdain.
2. Score 0.7: The overall response feels sarcastic or mocking, but perhaps not obviously.
3. Score 0.0: The response shows no indication of sarcastic or mocking emotions.

Please note that as a fair and objective evaluator, you should not adhere to an ``ambiguous'' principle and give a score of 0.7 regardless of circumstances, nor should you be too lenient or too strict. Your scoring must strictly follow the above scoring criteria!

You must only respond with one of these three numbers: 0, 0.7, or 1. Do not provide any explanations or additional text! Do not provide any explanations beyond these three numbers, do not make any narratives! Even if you have questions or want to discuss further, please remain restrained and output only one number!
\end{guidebox}

\clearpage

\begin{guidebox}[angry]
You are an AI multilingual evaluation expert who is precise and impartial. You need to help me assess whether the following model responses comply with the requirements in the instructions and return a score of 0, 0.7, or 1.

Please evaluate whether the following response's main tone expresses angry or furious emotions.

Instruction:
\{instruction\}

Model Response:
\{response\}

Detailed Scoring Criteria:
1. Score 1.0: The response content conveys strong emotions of anger, fury, or dissatisfaction, or the tone carries obvious aggression, impatience, or indignation.
2. Score 0.7: The overall tone tends to express anger rather than calm, gentle, or neutral emotions.
3. Score 0.0: The response shows no indication of ``anger or fury''.

Please note that as a fair and objective evaluator, you should not adhere to an ``ambiguous'' principle and give a score of 0.7 regardless of circumstances, nor should you be too lenient or too strict. Your scoring must strictly follow the above scoring criteria!

You must only respond with one of these three numbers: 0, 0.7, or 1. Do not provide any explanations or additional text! Do not provide any explanations beyond these three numbers, do not make any narratives! Even if you have questions or want to discuss further, please remain restrained and output only one number!
\end{guidebox}

\begin{guidebox}[jokes]
You are an AI multilingual evaluation expert who is precise and impartial. You need to help me assess whether the following model responses comply with the requirements in the instructions and return a score of 0, 0.7, or 1.

Please evaluate whether the following response includes at least 3 jokes.

Instruction:
\{instruction\}

Model Response:
\{response\}

Detailed Scoring Criteria:
1. Score 1.0: The response clearly contains at least 3 jokes. There is no length restriction for ``jokes'' - even if it's as short as one word, as long as it has joke elements, it can be counted as a joke.
2. Score 0.7: The ``jokes'' in the response are somewhat forced, or they appear but only two in number.
3. Score 0.0: The response contains only one joke, or none at all.

Please note that as a fair and objective evaluator, you should not adhere to an ``ambiguous'' principle and give a score of 0.7 regardless of circumstances, nor should you be too lenient or too strict. Your scoring must strictly follow the above scoring criteria!

You must only respond with one of these three numbers: 0, 0.7, or 1. Do not provide any explanations or additional text! Do not provide any explanations beyond these three numbers, do not make any narratives! Even if you have questions or want to discuss further, please remain restrained and output only one number!
\end{guidebox}

\clearpage

\begin{guidebox}[quotes]
You are an AI multilingual evaluation expert who is precise and impartial. You need to help me assess whether the following model responses comply with the requirements in the instructions and return a score of 0, 0.7, or 1.

Please evaluate whether the following response includes at least 3 famous quotes.

Instruction:
\{instruction\}

Model Response:
\{response\}

Detailed Scoring Criteria:
1. Score 1.0: The response clearly cites at least 3 different famous quotes. Note that even for the same famous person, two different quotes should be counted separately. The ``famous person'' or ``quote'' doesn't necessarily need to be well-known - as long as it's a quote from someone relevant to the topic, it can be considered a ``famous quote''.
2. Score 0.7: The response contains only two famous quotes, or the famous quotes that appear in the response are not obvious.
3. Score 0.0: The response contains only one quote, or no famous quotes at all.

Please note that as a fair and objective evaluator, you should not adhere to an ``ambiguous'' principle and give a score of 0.7 regardless of circumstances, nor should you be too lenient or too strict. Your scoring must strictly follow the above scoring criteria!

You must only respond with one of these three numbers: 0, 0.7, or 1. Do not provide any explanations or additional text! Do not provide any explanations beyond these three numbers, do not make any narratives! Even if you have questions or want to discuss further, please remain restrained and output only one number!
\end{guidebox}

\begin{guidebox}[celebrity]
You are an AI multilingual evaluation expert who is precise and impartial. You need to help me assess whether the following model responses comply with the requirements in the instructions and return a score of 0, 0.7, or 1.

Please evaluate whether the following response mentions a famous person related to the topic and briefly introduces their achievements.

Instruction:
\{instruction\}

Model Response:
\{response\}

Detailed Scoring Criteria:
1. Score 1.0: The response mentions a person relevant to the topic and briefly introduces their main achievements or contributions, even if the person is not that ``famous''.
2. Score 0.7: The response mentions a famous person, but they are not very relevant to the topic, or their achievements are not introduced.
3. Score 0.0: The response does not mention any famous person at all.

Please note that as a fair and objective evaluator, you should not adhere to an ``ambiguous'' principle and give a score of 0.7 regardless of circumstances, nor should you be too lenient or too strict. Your scoring must strictly follow the above scoring criteria!

You must only respond with one of these three numbers: 0, 0.7, or 1. Do not provide any explanations or additional text! Do not provide any explanations beyond these three numbers, do not make any narratives! Even if you have questions or want to discuss further, please remain restrained and output only one number!
\end{guidebox}

\clearpage

\begin{guidebox}[multilingual]
You are an AI multilingual evaluation expert who is precise and impartial. You need to help me assess whether the following model responses comply with the requirements in the instructions and return a score of 0, 0.7, or 1.

Please evaluate whether the following response includes at least three different languages.

Instruction:
\{instruction\}

Model Response:
\{response\}

Detailed Scoring Criteria:
1. Score 1.0: The response clearly uses three or more different languages.
2. Score 0.7: The response uses two different languages.
3. Score 0.0: The response is output in a single language and does not include any other languages.

Please note that as a fair and objective evaluator, you should not adhere to an ``ambiguous'' principle and give a score of 0.7 regardless of circumstances, nor should you be too lenient or too strict. Your scoring must strictly follow the above scoring criteria!

You must only respond with one of these three numbers: 0, 0.7, or 1. Do not provide any explanations or additional text! Do not provide any explanations beyond these three numbers, do not make any narratives! Even if you have questions or want to discuss further, please remain restrained and output only one number!
\end{guidebox}

\begin{guidebox}[repeat]
You are an AI multilingual evaluation expert who is precise and impartial. You need to help me assess whether the following model responses comply with the requirements in the instructions and return a score of 0, 0.7, or 1.

Please evaluate whether the following response repeats once and uses a different language the second time.

Instruction:
\{instruction\}

Model Response:
\{response\}

Detailed Scoring Criteria:
1. Score 1.0: Whether the response contains two languages and both languages express the same content. The descriptions in the two languages do not necessarily need to express exactly the same meaning, as long as the general content expressed is similar.
2. Score 0.7: The response contains two languages, but the expressions in the two languages differ greatly.
3. Score 0.0: The response does not repeat at all, or only uses one language.

Please note that as a fair and objective evaluator, you should not adhere to an ``ambiguous'' principle and give a score of 0.7 regardless of circumstances, nor should you be too lenient or too strict. Your scoring must strictly follow the above scoring criteria!

You must only respond with one of these three numbers: 0, 0.7, or 1. Do not provide any explanations or additional text! Do not provide any explanations beyond these three numbers, do not make any narratives! Even if you have questions or want to discuss further, please remain restrained and output only one number!
\end{guidebox}

\clearpage
\twocolumn
\subsection{Data Extension Methods}
\label{Data Extension}

Our script encompasses the core functionality of the entire framework, enabling key features such as language addition, instruction construction, model invocation for response generation, automated scoring, and result file output. It also supports the addition of new languages and instructions, demonstrating strong scalability.

To add a new language, we simply need to write the translation prompt words for the language to be added based on the translation prompt template, input them into the specified file in JSONL format, and execute the designated script to automatically translate English kwargs into the corresponding language. Subsequently, by executing another designated script to translate English prompts into the corresponding language, we can complete the steps for adding a new language.

When adding a new instruction, it is only necessary to incorporate the instruction into the specified script, construct the corresponding instruction ID and kwargs, and establish the mapping relationship between the new instruction and its corresponding instruction ID within the designated script. Note that for model-based instructions, it is required to add the evaluation prompt for the new instruction in the prompt file. After completing these steps, the process of adding a new instruction is successfully accomplished.

\onecolumn
\section{Detail Results}
This section lists the additional results from baseline model experiments. All scores follow Loose Score standard unless specified otherwise. 

\subsection{Strict Score Evaluation Results of Each Model in 23 Languages}
\begin{table*}[htbp]
    \centering
    \resizebox{\textwidth}{!}{%
\begin{tabular}{llcccccc}
    \toprule
    \textbf{Resource} & \textbf{Language} & \textbf{GPT-4o} & \textbf{Claude-3.5 Sonnet} & \textbf{Gemini-1.5 Pro} & \textbf{Gemini-1.5 Flash} & \textbf{GPT-3.5 Turbo} & \textbf{Average} \\
    \midrule
    \multirow{7}{*}{\textbf{High}}
    & Swedish    & 82.88\% & 81.48\% & 76.24\% & 69.05\% & 59.12\% & 73.75\% \\
    & Portuguese & 81.09\% & 78.92\% & 73.17\% & 71.11\% & 62.07\% & 73.27\% \\
    & English    & 80.59\% & 78.92\% & 73.17\% & 69.66\% & 61.24\% & 72.72\% \\
    & French     & 78.75\% & 81.43\% & 71.00\% & 68.99\% & 58.95\% & 71.82\% \\
    & Italian    & 77.64\% & 77.60\% & 68.43\% & 67.82\% & 56.89\% & 69.68\% \\
    & Chinese    & 76.41\% & 75.40\% & 74.57\% & 67.82\% & 59.56\% & 70.75\% \\
    & Japanese   & 75.79\% & 76.58\% & 69.88\% & 70.61\% & 57.67\% & 70.11\% \\
\bottomrule
    \end{tabular}
    }
    \caption{Strict Score Evaluation Results of Each Model in 23 Languages}
    \label{tab:strict_score}
\end{table*}
\clearpage
\subsection{Evaluation Results of 11 Categories}
\begin{table*}[htbp]
\centering
\setlength{\tabcolsep}{3pt}
\begin{tabular}{lccccc}
\toprule
\textbf{Metrics} & \textbf{GPT-4o} & \textbf{Claude-3.5 Sonnet} & \textbf{Gemini-1.5 Pro} & \textbf{Gemini-1.5 Flash} & \textbf{GPT-3.5 Turbo} \\
\midrule
keywords & 73.98\% & 66.64\% & 64.18\% & 61.02\% & 56.19\% \\
length & 88.01\% & 66.72\% & 82.81\% & 69.27\% & 47.66\% \\
format & 76.53\% & 76.80\% & 83.20\% & 80.00\% & 42.51\% \\
repeat & 64.80\% & 59.30\% & 51.23\% & 58.01\% & 12.75\% \\
marks & 38.40\% & 61.81\% & 37.67\% & 26.75\% & 10.61\% \\
citation & 79.41\% & 66.67\% & 80.39\% & 67.45\% & 28.24\% \\
emoji & 89.55\% & 83.48\% & 83.93\% & 74.61\% & 35.51\% \\
style & 94.32\% & 85.88\% & 90.28\% & 91.60\% & 57.96\% \\
tone & 83.32\% & 94.63\% & 83.84\% & 88.43\% & 32.36\% \\
content & 93.06\% & 91.43\% & 80.27\% & 78.50\% & 3.27\% \\
language\_switch & 95.11\% & 95.43\% & 85.85\% & 80.32\% & 3.30\% \\
\bottomrule
\end{tabular}
\caption{Evaluation results for Armenian language}
\end{table*}

\begin{table*}[htbp]
\centering
\setlength{\tabcolsep}{3pt}
\begin{tabular}{lccccc}
\toprule
\textbf{Metrics} & \textbf{GPT-4o} & \textbf{Claude-3.5 Sonnet} & \textbf{Gemini-1.5 Pro} & \textbf{Gemini-1.5 Flash} & \textbf{GPT-3.5 Turbo} \\
\midrule
keywords & 65.93\% & 61.44\% & 61.58\% & 52.85\% & 45.08\% \\
length & 89.18\% & 62.92\% & 72.76\% & 75.40\% & 43.73\% \\
format & 86.19\% & 84.88\% & 74.91\% & 69.69\% & 50.76\% \\
repeat & 62.92\% & 58.83\% & 51.81\% & 50.29\% & 11.46\% \\
marks & 53.04\% & 74.42\% & 52.94\% & 35.91\% & 13.25\% \\
citation & 73.53\% & 72.65\% & 76.18\% & 60.29\% & 17.16\% \\
emoji & 85.73\% & 81.46\% & 82.47\% & 69.44\% & 47.87\% \\
style & 97.80\% & 94.52\% & 91.80\% & 97.36\% & 81.12\% \\
tone & 71.70\% & 90.00\% & 77.16\% & 86.20\% & 36.51\% \\
content & 95.71\% & 97.48\% & 83.40\% & 72.72\% & 23.95\% \\
language\_switch & 95.74\% & 98.94\% & 84.79\% & 55.64\% & 2.87\% \\
\bottomrule
\end{tabular}
\caption{Evaluation results for Bengali language}
\end{table*}

\begin{table*}[htbp]
\centering
\setlength{\tabcolsep}{3pt}
\begin{tabular}{lccccc}
\toprule
\textbf{Metrics} & \textbf{GPT-4o} & \textbf{Claude-3.5 Sonnet} & \textbf{Gemini-1.5 Pro} & \textbf{Gemini-1.5 Flash} & \textbf{GPT-3.5 Turbo} \\
\midrule
keywords & 74.52\% & 58.31\% & 73.62\% & 68.47\% & 66.78\% \\
length & 41.83\% & 40.50\% & 54.71\% & 52.22\% & 38.62\% \\
format & 90.41\% & 76.53\% & 87.25\% & 81.31\% & 78.73\% \\
repeat & 67.02\% & 57.54\% & 60.47\% & 62.11\% & 36.61\% \\
marks & 62.83\% & 93.18\% & 59.81\% & 49.28\% & 23.60\% \\
citation & 62.75\% & 94.12\% & 78.63\% & 72.75\% & 71.67\% \\
emoji & 88.88\% & 86.74\% & 84.04\% & 83.26\% & 79.33\% \\
style & 96.00\% & 95.80\% & 90.56\% & 89.84\% & 95.64\% \\
tone & 96.55\% & 98.12\% & 94.15\% & 90.83\% & 83.49\% \\
content & 99.59\% & 98.64\% & 88.57\% & 88.78\% & 94.22\% \\
language\_switch & 79.79\% & 80.53\% & 75.64\% & 45.85\% & 11.70\% \\
\bottomrule
\end{tabular}
\caption{Evaluation results for Chinese language}
\end{table*}

\begin{table*}[htbp]
\centering
\setlength{\tabcolsep}{3pt}
\begin{tabular}{lccccc}
\toprule
\textbf{Metrics} & \textbf{GPT-4o} & \textbf{Claude-3.5 Sonnet} & \textbf{Gemini-1.5 Pro} & \textbf{Gemini-1.5 Flash} & \textbf{GPT-3.5 Turbo} \\
\midrule
keywords & 86.84\% & 87.03\% & 85.79\% & 79.18\% & 75.06\% \\
length & 80.51\% & 77.15\% & 66.85\% & 85.05\% & 58.12\% \\
format & 83.06\% & 71.68\% & 76.08\% & 85.43\% & 71.68\% \\
repeat & 74.74\% & 56.61\% & 63.04\% & 44.80\% & 29.71\% \\
marks & 71.78\% & 86.21\% & 52.93\% & 59.40\% & 29.57\% \\
citation & 60.78\% & 86.27\% & 74.61\% & 68.63\% & 53.92\% \\
emoji & 87.64\% & 85.73\% & 85.17\% & 76.63\% & 69.66\% \\
style & 95.56\% & 93.20\% & 88.80\% & 85.44\% & 91.88\% \\
tone & 97.12\% & 99.08\% & 95.85\% & 92.36\% & 93.41\% \\
content & 99.39\% & 98.50\% & 90.75\% & 86.87\% & 97.14\% \\
language\_switch & 97.87\% & 97.55\% & 89.36\% & 81.17\% & 57.02\% \\
\bottomrule
\end{tabular}
\caption{Evaluation results for English language}
\end{table*}

\begin{table*}[htbp]
\centering
\setlength{\tabcolsep}{3pt}
\begin{tabular}{lccccc}
\toprule
\textbf{Metrics} & \textbf{GPT-4o} & \textbf{Claude-3.5 Sonnet} & \textbf{Gemini-1.5 Pro} & \textbf{Gemini-1.5 Flash} & \textbf{GPT-3.5 Turbo} \\
\midrule
keywords & 90.56\% & 90.00\% & 84.12\% & 79.72\% & 63.56\% \\
length & 63.36\% & 73.79\% & 72.72\% & 49.75\% & 56.80\% \\
format & 95.50\% & 88.90\% & 88.42\% & 88.49\% & 71.24\% \\
repeat & 76.73\% & 67.95\% & 52.75\% & 60.47\% & 22.92\% \\
marks & 68.45\% & 75.81\% & 58.62\% & 46.63\% & 6.61\% \\
citation & 86.47\% & 77.45\% & 82.75\% & 56.67\% & 50.20\% \\
emoji & 85.51\% & 65.62\% & 81.12\% & 82.81\% & 65.96\% \\
style & 94.84\% & 90.52\% & 89.28\% & 89.52\% & 89.04\% \\
tone & 89.65\% & 96.29\% & 92.23\% & 94.28\% & 75.98\% \\
content & 97.82\% & 99.18\% & 92.04\% & 84.76\% & 78.50\% \\
language\_switch & 97.66\% & 99.68\% & 72.98\% & 47.02\% & 38.83\% \\
\bottomrule
\end{tabular}
\caption{Evaluation results for Filipino language}
\end{table*}

\begin{table*}[htbp]
\centering
\setlength{\tabcolsep}{3pt}
\begin{tabular}{lccccc}
\toprule
\textbf{Metrics} & \textbf{GPT-4o} & \textbf{Claude-3.5 Sonnet} & \textbf{Gemini-1.5 Pro} & \textbf{Gemini-1.5 Flash} & \textbf{GPT-3.5 Turbo} \\
\midrule
keywords & 89.97\% & 90.03\% & 80.68\% & 76.38\% & 69.97\% \\
length & 77.57\% & 83.05\% & 72.27\% & 71.68\% & 70.11\% \\
format & 87.90\% & 81.55\% & 77.42\% & 76.32\% & 75.88\% \\
repeat & 75.44\% & 63.63\% & 61.99\% & 60.94\% & 34.85\% \\
marks & 59.01\% & 83.26\% & 46.13\% & 45.98\% & 23.17\% \\
citation & 50.69\% & 88.24\% & 74.41\% & 64.51\% & 44.12\% \\
emoji & 84.16\% & 73.15\% & 83.60\% & 75.84\% & 67.75\% \\
style & 94.12\% & 94.12\% & 91.00\% & 90.44\% & 91.60\% \\
tone & 94.02\% & 98.69\% & 92.97\% & 93.41\% & 87.73\% \\
content & 98.50\% & 98.98\% & 89.93\% & 86.80\% & 90.27\% \\
language\_switch & 94.79\% & 100.00\% & 77.02\% & 74.15\% & 52.55\% \\
\bottomrule
\end{tabular}
\caption{Evaluation results for French language}
\end{table*}

\begin{table*}[htbp]
\centering
\setlength{\tabcolsep}{3pt}
\begin{tabular}{lccccc}
\toprule
\textbf{Metrics} & \textbf{GPT-4o} & \textbf{Claude-3.5 Sonnet} & \textbf{Gemini-1.5 Pro} & \textbf{Gemini-1.5 Flash} & \textbf{GPT-3.5 Turbo} \\
\midrule
keywords & 71.10\% & 57.82\% & 60.99\% & 63.25\% & 53.08\% \\
length & 87.51\% & 66.31\% & 71.56\% & 68.08\% & 53.60\% \\
format & 73.99\% & 67.63\% & 82.34\% & 73.92\% & 44.12\% \\
repeat & 68.77\% & 53.80\% & 65.50\% & 57.43\% & 8.54\% \\
marks & 38.52\% & 61.99\% & 35.02\% & 16.69\% & 9.91\% \\
citation & 63.63\% & 63.73\% & 75.20\% & 60.78\% & 36.37\% \\
emoji & 86.97\% & 81.24\% & 81.91\% & 73.15\% & 45.62\% \\
style & 93.04\% & 86.52\% & 86.84\% & 86.56\% & 56.96\% \\
tone & 89.26\% & 92.97\% & 85.50\% & 92.14\% & 32.18\% \\
content & 93.33\% & 91.70\% & 74.69\% & 75.71\% & 11.22\% \\
language\_switch & 95.43\% & 91.49\% & 71.70\% & 53.40\% & 3.30\% \\
\bottomrule
\end{tabular}
\caption{Evaluation results for Georgian language}
\end{table*}

\begin{table*}[htbp]
\centering
\setlength{\tabcolsep}{3pt}
\begin{tabular}{lccccc}
\toprule
\textbf{Metrics} & \textbf{GPT-4o} & \textbf{Claude-3.5 Sonnet} & \textbf{Gemini-1.5 Pro} & \textbf{Gemini-1.5 Flash} & \textbf{GPT-3.5 Turbo} \\
\midrule
keywords & 60.54\% & 63.14\% & 50.23\% & 51.89\% & 46.30\% \\
length & 69.05\% & 80.47\% & 54.33\% & 54.58\% & 43.17\% \\
format & 67.91\% & 62.98\% & 54.77\% & 61.20\% & 63.24\% \\
repeat & 69.59\% & 51.46\% & 47.72\% & 45.03\% & 20.70\% \\
marks & 64.00\% & 77.75\% & 53.67\% & 47.28\% & 7.48\% \\
citation & 87.25\% & 75.49\% & 72.55\% & 55.88\% & 51.08\% \\
emoji & 83.26\% & 84.94\% & 84.72\% & 77.19\% & 59.66\% \\
style & 97.52\% & 81.16\% & 89.72\% & 93.76\% & 83.96\% \\
tone & 76.42\% & 86.90\% & 88.56\% & 88.08\% & 46.11\% \\
content & 97.96\% & 85.78\% & 87.82\% & 83.74\% & 81.77\% \\
language\_switch & 97.55\% & 78.83\% & 82.45\% & 68.40\% & 30.32\% \\
\bottomrule
\end{tabular}
\caption{Evaluation results for Hindi language}
\end{table*}

\begin{table*}[htbp]
\centering
\setlength{\tabcolsep}{3pt}
\begin{tabular}{lccccc}
\toprule
\textbf{Metrics} & \textbf{GPT-4o} & \textbf{Claude-3.5 Sonnet} & \textbf{Gemini-1.5 Pro} & \textbf{Gemini-1.5 Flash} & \textbf{GPT-3.5 Turbo} \\
\midrule
keywords & 87.23\% & 87.68\% & 79.77\% & 75.48\% & 67.74\% \\
length & 84.41\% & 78.19\% & 81.85\% & 79.29\% & 59.98\% \\
format & 86.63\% & 88.01\% & 81.99\% & 78.73\% & 68.21\% \\
repeat & 66.90\% & 64.56\% & 61.75\% & 54.15\% & 33.92\% \\
marks & 69.95\% & 85.83\% & 49.55\% & 50.31\% & 32.01\% \\
citation & 54.61\% & 84.31\% & 68.43\% & 63.92\% & 37.25\% \\
emoji & 87.87\% & 83.15\% & 84.16\% & 71.01\% & 73.60\% \\
style & 94.64\% & 92.52\% & 88.00\% & 88.64\% & 93.88\% \\
tone & 89.00\% & 96.59\% & 85.94\% & 85.28\% & 76.64\% \\
content & 97.55\% & 98.78\% & 83.40\% & 85.92\% & 95.78\% \\
language\_switch & 95.43\% & 99.68\% & 89.57\% & 56.06\% & 18.62\% \\
\bottomrule
\end{tabular}
\caption{Evaluation results for Indonesian language}
\end{table*}

\begin{table*}[htbp]
\centering
\setlength{\tabcolsep}{3pt}
\begin{tabular}{lccccc}
\toprule
\textbf{Metrics} & \textbf{GPT-4o} & \textbf{Claude-3.5 Sonnet} & \textbf{Gemini-1.5 Pro} & \textbf{Gemini-1.5 Flash} & \textbf{GPT-3.5 Turbo} \\
\midrule
keywords & 91.84\% & 88.33\% & 78.76\% & 75.68\% & 68.90\% \\
length & 76.65\% & 81.66\% & 76.91\% & 71.32\% & 67.53\% \\
format & 71.92\% & 73.28\% & 71.99\% & 77.80\% & 61.82\% \\
repeat & 75.67\% & 67.50\% & 58.36\% & 59.53\% & 32.63\% \\
marks & 64.53\% & 74.45\% & 50.93\% & 47.21\% & 30.48\% \\
citation & 93.14\% & 84.31\% & 77.75\% & 66.76\% & 79.80\% \\
emoji & 86.63\% & 67.53\% & 77.64\% & 79.89\% & 65.84\% \\
style & 94.44\% & 94.12\% & 87.24\% & 89.08\% & 89.48\% \\
tone & 94.15\% & 98.56\% & 92.23\% & 93.32\% & 87.60\% \\
content & 85.58\% & 92.38\% & 85.85\% & 79.39\% & 77.62\% \\
language\_switch & 98.94\% & 99.68\% & 77.66\% & 81.91\% & 52.55\% \\
\bottomrule
\end{tabular}
\caption{Evaluation results for Italian language}
\end{table*}

\begin{table*}[htbp]
\centering
\setlength{\tabcolsep}{3pt}
\begin{tabular}{lccccc}
\toprule
\textbf{Metrics} & \textbf{GPT-4o} & \textbf{Claude-3.5 Sonnet} & \textbf{Gemini-1.5 Pro} & \textbf{Gemini-1.5 Flash} & \textbf{GPT-3.5 Turbo} \\
\midrule
keywords & 71.16\% & 59.83\% & 74.69\% & 67.32\% & 61.69\% \\
length & 75.28\% & 77.69\% & 66.32\% & 59.15\% & 49.58\% \\
format & 82.82\% & 71.41\% & 83.64\% & 85.57\% & 70.84\% \\
repeat & 68.07\% & 43.27\% & 34.85\% & 53.10\% & 34.27\% \\
marks & 59.61\% & 90.75\% & 52.87\% & 52.62\% & 19.61\% \\
citation & 60.78\% & 84.31\% & 63.24\% & 70.49\% & 74.02\% \\
emoji & 88.65\% & 86.29\% & 87.19\% & 76.97\% & 74.83\% \\
style & 96.36\% & 96.08\% & 89.84\% & 95.72\% & 93.36\% \\
tone & 84.98\% & 99.08\% & 89.00\% & 91.05\% & 75.24\% \\
content & 96.67\% & 99.59\% & 88.50\% & 82.72\% & 92.18\% \\
language\_switch & 96.81\% & 100.00\% & 89.04\% & 86.17\% & 54.36\% \\
\bottomrule
\end{tabular}
\caption{Evaluation results for Japanese language}
\end{table*}

\begin{table*}[htbp]
\centering
\setlength{\tabcolsep}{3pt}
\begin{tabular}{lccccc}
\toprule
\textbf{Metrics} & \textbf{GPT-4o} & \textbf{Claude-3.5 Sonnet} & \textbf{Gemini-1.5 Pro} & \textbf{Gemini-1.5 Flash} & \textbf{GPT-3.5 Turbo} \\
\midrule
keywords & 70.85\% & 68.62\% & 67.15\% & 62.80\% & 56.30\% \\
length & 80.31\% & 82.60\% & 56.64\% & 67.80\% & 25.86\% \\
format & 74.47\% & 80.52\% & 69.45\% & 71.99\% & 58.32\% \\
repeat & 44.09\% & 33.57\% & 33.57\% & 29.24\% & 21.87\% \\
marks & 61.49\% & 84.68\% & 62.35\% & 60.91\% & 38.47\% \\
citation & 93.14\% & 84.31\% & 73.82\% & 62.45\% & 81.08\% \\
emoji & 79.89\% & 86.52\% & 73.93\% & 64.61\% & 54.83\% \\
style & 95.48\% & 91.68\% & 87.80\% & 87.60\% & 84.96\% \\
tone & 84.98\% & 95.15\% & 90.52\% & 88.52\% & 81.88\% \\
content & 97.28\% & 92.99\% & 89.73\% & 87.28\% & 90.88\% \\
language\_switch & 94.36\% & 91.17\% & 91.49\% & 79.57\% & 37.77\% \\
\bottomrule
\end{tabular}
\caption{Evaluation results for Korean language}
\end{table*}

\begin{table*}[htbp]
\centering
\setlength{\tabcolsep}{3pt}
\begin{tabular}{lccccc}
\toprule
\textbf{Metrics} & \textbf{GPT-4o} & \textbf{Claude-3.5 Sonnet} & \textbf{Gemini-1.5 Pro} & \textbf{Gemini-1.5 Flash} & \textbf{GPT-3.5 Turbo} \\
\midrule
keywords & 84.60\% & 75.68\% & 75.40\% & 69.69\% & 36.21\% \\
length & 91.27\% & 63.12\% & 83.62\% & 83.41\% & 37.80\% \\
format & 72.99\% & 73.61\% & 65.98\% & 83.64\% & 49.83\% \\
repeat & 73.22\% & 56.61\% & 58.83\% & 56.02\% & 7.84\% \\
marks & 69.56\% & 84.16\% & 51.12\% & 35.56\% & 7.79\% \\
citation & 78.43\% & 76.47\% & 80.00\% & 61.47\% & 37.94\% \\
emoji & 82.13\% & 81.91\% & 78.20\% & 74.83\% & 23.03\% \\
style & 92.96\% & 89.32\% & 91.52\% & 93.92\% & 52.96\% \\
tone & 62.79\% & 84.19\% & 84.45\% & 82.93\% & 23.89\% \\
content & 94.42\% & 89.12\% & 89.39\% & 80.95\% & 5.71\% \\
language\_switch & 96.81\% & 90.85\% & 94.47\% & 81.70\% & 2.55\% \\
\bottomrule
\end{tabular}
\caption{Evaluation results for Kyrgyz language}
\end{table*}

\begin{table*}[htbp]
\centering
\setlength{\tabcolsep}{3pt}
\begin{tabular}{lccccc}
\toprule
\textbf{Metrics} & \textbf{GPT-4o} & \textbf{Claude-3.5 Sonnet} & \textbf{Gemini-1.5 Pro} & \textbf{Gemini-1.5 Flash} & \textbf{GPT-3.5 Turbo} \\
\midrule
keywords & 82.60\% & 82.03\% & 71.55\% & 65.88\% & 41.55\% \\
length & 73.27\% & 79.75\% & 88.29\% & 72.60\% & 41.94\% \\
format & 83.30\% & 71.44\% & 78.42\% & 79.24\% & 33.81\% \\
repeat & 67.37\% & 68.54\% & 43.86\% & 49.59\% & 8.54\% \\
marks & 18.69\% & 47.15\% & 26.63\% & 7.51\% & 12.23\% \\
citation & 76.08\% & 68.24\% & 69.61\% & 55.59\% & 33.82\% \\
emoji & 75.84\% & 65.06\% & 67.30\% & 65.62\% & 36.40\% \\
style & 95.68\% & 92.84\% & 85.52\% & 90.96\% & 64.72\% \\
tone & 77.38\% & 77.16\% & 74.98\% & 73.84\% & 24.54\% \\
content & 66.73\% & 78.78\% & 60.95\% & 56.12\% & 3.20\% \\
language\_switch & 95.43\% & 83.72\% & 65.85\% & 59.89\% & 0.00\% \\
\bottomrule
\end{tabular}
\caption{Evaluation results for Malagasy language}
\end{table*}

\begin{table*}[htbp]
\centering
\setlength{\tabcolsep}{3pt}
\begin{tabular}{lccccc}
\toprule
\textbf{Metrics} & \textbf{GPT-4o} & \textbf{Claude-3.5 Sonnet} & \textbf{Gemini-1.5 Pro} & \textbf{Gemini-1.5 Flash} & \textbf{GPT-3.5 Turbo} \\
\midrule
keywords & 89.41\% & 87.40\% & 80.40\% & 82.80\% & 71.19\% \\
length & 82.67\% & 87.66\% & 88.56\% & 75.37\% & 67.73\% \\
format & 89.14\% & 79.76\% & 88.14\% & 80.76\% & 66.84\% \\
repeat & 61.29\% & 55.20\% & 60.82\% & 46.90\% & 33.45\% \\
marks & 52.75\% & 71.93\% & 38.95\% & 40.37\% & 21.37\% \\
citation & 57.84\% & 77.16\% & 70.69\% & 61.67\% & 55.59\% \\
emoji & 87.64\% & 82.25\% & 83.93\% & 71.69\% & 67.53\% \\
style & 96.96\% & 92.80\% & 93.60\% & 90.72\% & 93.96\% \\
tone & 90.17\% & 97.34\% & 93.89\% & 88.82\% & 76.94\% \\
content & 96.39\% & 98.44\% & 88.44\% & 81.70\% & 85.51\% \\
language\_switch & 95.43\% & 99.68\% & 86.06\% & 59.89\% & 21.91\% \\
\bottomrule
\end{tabular}
\caption{Evaluation results for Malay language}
\end{table*}

\begin{table*}[htbp]
\centering
\setlength{\tabcolsep}{3pt}
\begin{tabular}{lccccc}
\toprule
\textbf{Metrics} & \textbf{GPT-4o} & \textbf{Claude-3.5 Sonnet} & \textbf{Gemini-1.5 Pro} & \textbf{Gemini-1.5 Flash} & \textbf{GPT-3.5 Turbo} \\
\midrule
keywords & 89.04\% & 83.53\% & 78.93\% & 75.17\% & 69.94\% \\
length & 72.24\% & 81.20\% & 72.83\% & 70.45\% & 60.46\% \\
format & 77.29\% & 77.49\% & 82.23\% & 76.29\% & 68.97\% \\
repeat & 72.75\% & 63.39\% & 61.64\% & 61.64\% & 35.67\% \\
marks & 85.05\% & 83.15\% & 56.69\% & 54.58\% & 33.79\% \\
citation & 91.18\% & 84.31\% & 82.25\% & 76.57\% & 86.27\% \\
emoji & 85.51\% & 67.08\% & 77.42\% & 76.18\% & 66.52\% \\
style & 94.08\% & 89.84\% & 90.36\% & 90.68\% & 91.72\% \\
tone & 93.41\% & 97.95\% & 94.59\% & 94.98\% & 89.48\% \\
content & 96.26\% & 95.92\% & 88.91\% & 83.88\% & 97.21\% \\
language\_switch & 97.87\% & 95.11\% & 86.70\% & 85.64\% & 56.17\% \\
\bottomrule
\end{tabular}
\caption{Evaluation results for Portuguese language}
\end{table*}

\begin{table*}[htbp]
\centering
\setlength{\tabcolsep}{3pt}
\begin{tabular}{lccccc}
\toprule
\textbf{Metrics} & \textbf{GPT-4o} & \textbf{Claude-3.5 Sonnet} & \textbf{Gemini-1.5 Pro} & \textbf{Gemini-1.5 Flash} & \textbf{GPT-3.5 Turbo} \\
\midrule
keywords & 46.10\% & 86.30\% & 62.54\% & 63.47\% & 32.46\% \\
length & 38.20\% & 49.33\% & 49.84\% & 48.42\% & 39.20\% \\
format & 56.80\% & 73.16\% & 66.43\% & 53.88\% & 39.52\% \\
repeat & 18.36\% & 38.48\% & 18.95\% & 20.70\% & 15.44\% \\
marks & 19.30\% & 44.82\% & 27.35\% & 11.74\% & 12.32\% \\
citation & 43.04\% & 86.27\% & 46.86\% & 64.41\% & 34.12\% \\
emoji & 34.49\% & 78.76\% & 75.39\% & 64.72\% & 29.55\% \\
style & 72.72\% & 80.96\% & 63.92\% & 61.16\% & 49.20\% \\
tone & 34.41\% & 73.32\% & 43.36\% & 59.56\% & 21.44\% \\
content & 9.39\% & 63.40\% & 20.88\% & 21.77\% & 2.52\% \\
language\_switch & 18.30\% & 61.06\% & 15.21\% & 59.89\% & 41.28\% \\
\bottomrule
\end{tabular}
\caption{Evaluation results for Quechua language}
\end{table*}

\begin{table*}[htbp]
\centering
\setlength{\tabcolsep}{3pt}
\begin{tabular}{lccccc}
\toprule
\textbf{Metrics} & \textbf{GPT-4o} & \textbf{Claude-3.5 Sonnet} & \textbf{Gemini-1.5 Pro} & \textbf{Gemini-1.5 Flash} & \textbf{GPT-3.5 Turbo} \\
\midrule
keywords & 71.58\% & 70.56\% & 65.11\% & 60.51\% & 54.60\% \\
length & 84.07\% & 85.51\% & 84.96\% & 77.52\% & 69.81\% \\
format & 91.34\% & 84.98\% & 86.05\% & 82.13\% & 73.09\% \\
repeat & 72.05\% & 69.12\% & 61.52\% & 52.98\% & 27.60\% \\
marks & 75.08\% & 89.72\% & 70.83\% & 62.21\% & 28.11\% \\
citation & 91.18\% & 86.27\% & 87.25\% & 62.55\% & 79.12\% \\
emoji & 82.13\% & 70.34\% & 81.69\% & 73.03\% & 65.73\% \\
style & 94.36\% & 91.84\% & 87.84\% & 88.48\% & 91.64\% \\
tone & 93.89\% & 99.21\% & 95.46\% & 92.75\% & 83.89\% \\
content & 99.32\% & 98.71\% & 93.06\% & 84.22\% & 91.77\% \\
language\_switch & 99.68\% & 100.00\% & 91.28\% &81.17\% &51.91\% \\
\bottomrule
\end{tabular}
\caption{Evaluation results for Romanian language}
\end{table*}

\begin{table*}[htbp]
\centering
\setlength{\tabcolsep}{3pt}
\begin{tabular}{lccccc}
\toprule
\textbf{Metrics} & \textbf{GPT-4o} & \textbf{Claude-3.5 Sonnet} & \textbf{Gemini-1.5 Pro} & \textbf{Gemini-1.5 Flash} & \textbf{GPT-3.5 Turbo} \\
\midrule
keywords & 80.96\% & 80.93\% & 78.19\% & 74.46\% & 66.53\% \\
length & 86.95\% & 96.43\% & 84.61\% & 82.36\% & 73.41\% \\
format & 89.18\% & 80.89\% & 84.54\% & 75.67\% & 65.12\% \\
repeat & 77.54\% & 63.39\% & 55.32\% & 57.31\% & 32.40\% \\
marks & 89.74\% & 93.97\% & 65.74\% & 60.20\% & 46.58\% \\
citation & 89.22\% & 85.29\% & 88.92\% & 73.24\% & 82.06\% \\
emoji & 85.51\% & 79.55\% & 83.37\% & 79.33\% & 69.66\% \\
style & 93.12\% & 92.40\% & 89.16\% & 88.48\% & 88.88\% \\
tone & 94.76\% & 98.17\% & 95.11\% & 93.84\% & 88.38\% \\
content & 98.71\% & 99.18\% & 91.22\% & 89.25\% & 90.20\% \\
language\_switch & 100.00\% & 100.00\% & 88.51\% & 67.34\% & 43.40\% \\
\bottomrule
\end{tabular}
\caption{Evaluation results for Swedish language}
\end{table*}

\begin{table*}[htbp]
\centering
\setlength{\tabcolsep}{3pt}
\begin{tabular}{lccccc}
\toprule
\textbf{Metrics} & \textbf{GPT-4o} & \textbf{Claude-3.5 Sonnet} & \textbf{Gemini-1.5 Pro} & \textbf{Gemini-1.5 Flash} & \textbf{GPT-3.5 Turbo} \\
\midrule
keywords & 57.26\% & 52.12\% & 46.33\% & 48.50\% & 43.19\% \\
length & 59.54\% & 48.42\% & 78.22\% & 75.28\% & 59.86\% \\
format & 82.56\% & 56.50\% & 62.57\% & 70.26\% & 34.46\% \\
repeat & 47.84\% & 24.09\% & 27.02\% & 38.95\% & 8.42\% \\
marks & 38.01\% & 74.81\% & 52.32\% & 42.81\% & 13.61\% \\
citation & 70.59\% & 60.78\% & 66.08\% & 60.10\% & 34.80\% \\
emoji & 83.26\% & 82.58\% & 79.44\% & 74.94\% & 49.44\% \\
style & 97.00\% & 87.64\% & 87.80\% & 91.72\% & 60.96\% \\
tone & 70.22\% & 87.73\% & 79.69\% & 82.58\% & 28.38\% \\
content & 97.35\% & 88.78\% & 87.14\% & 78.50\% & 19.25\% \\
language\_switch & 99.68\% & 94.04\% & 82.77\% & 79.68\% & 3.94\% \\
\bottomrule
\end{tabular}
\caption{Evaluation results for Tamil language}
\end{table*}

\begin{table*}[htbp]
\centering
\setlength{\tabcolsep}{3pt}
\begin{tabular}{lccccc}
\toprule
\textbf{Metrics} & \textbf{GPT-4o} & \textbf{Claude-3.5 Sonnet} & \textbf{Gemini-1.5 Pro} & \textbf{Gemini-1.5 Flash} & \textbf{GPT-3.5 Turbo} \\
\midrule
keywords & 58.50\% & 59.27\% & 55.79\% & 53.53\% & 49.15\% \\
length & 71.62\% & 66.33\% & 72.85\% & 79.41\% & 48.89\% \\
format & 59.08\% & 63.13\% & 61.07\% & 65.33\% & 35.30\% \\
repeat & 68.77\% & 47.95\% & 46.55\% & 51.35\% & 8.07\% \\
marks & 46.71\% & 76.99\% & 54.61\% & 46.93\% & 21.83\% \\
citation & 66.37\% & 54.90\% & 65.20\% & 63.24\% & 30.78\% \\
emoji & 80.67\% & 79.66\% & 81.91\% & 75.28\% & 33.03\% \\
style & 83.36\% & 80.48\% & 80.20\% & 86.72\% & 57.44\% \\
tone & 67.55\% & 85.20\% & 73.06\% & 76.77\% & 27.55\% \\
content & 83.06\% & 82.65\% & 78.78\% & 82.45\% & 28.44\% \\
language\_switch & 87.34\% & 91.49\% & 80.32\% & 75.74\% & 1.06\% \\
\bottomrule
\end{tabular}
\caption{Evaluation results for Telugu language}
\end{table*}

\begin{table*}[htbp]
\centering
\setlength{\tabcolsep}{3pt}
\begin{tabular}{lccccc}
\toprule
\textbf{Metrics} & \textbf{GPT-4o} & \textbf{Claude-3.5 Sonnet} & \textbf{Gemini-1.5 Pro} & \textbf{Gemini-1.5 Flash} & \textbf{GPT-3.5 Turbo} \\
\midrule
keywords & 77.71\% & 69.89\% & 65.76\% & 66.72\% & 62.68\% \\
length & 84.76\% & 79.19\% & 83.07\% & 82.42\% & 67.79\% \\
format & 74.43\% & 58.83\% & 65.40\% & 69.42\% & 63.57\% \\
repeat & 71.23\% & 58.71\% & 55.56\% & 47.84\% & 27.49\% \\
marks & 60.77\% & 88.21\% & 47.30\% & 35.98\% & 18.62\% \\
citation & 85.29\% & 88.24\% & 77.65\% & 66.27\% & 82.75\% \\
emoji & 86.63\% & 83.48\% & 87.64\% & 79.55\% & 73.15\% \\
style & 97.56\% & 92.88\% & 96.32\% & 95.88\% & 91.48\% \\
tone & 78.91\% & 95.15\% & 93.06\% & 88.34\% & 71.83\% \\
content & 78.91\% & 80.95\% & 59.66\% & 51.36\% & 65.24\% \\
language\_switch & 98.94\% & 96.49\% & 93.83\% & 87.34\% & 56.91\% \\
\bottomrule
\end{tabular}
\caption{Evaluation results for Turkish language}
\end{table*}

\begin{table*}[htbp]
\centering
\setlength{\tabcolsep}{3pt}
\begin{tabular}{lccccc}
\toprule
\textbf{Metrics} & \textbf{GPT-4o} & \textbf{Claude-3.5 Sonnet} & \textbf{Gemini-1.5 Pro} & \textbf{Gemini-1.5 Flash} & \textbf{GPT-3.5 Turbo} \\
\midrule
keywords & 73.56\% & 80.62\% & 59.24\% & 58.22\% & 48.19\% \\
length & 79.48\% & 64.00\% & 60.28\% & 64.71\% & 26.80\% \\
format & 77.77\% & 68.52\% & 68.90\% & 75.64\% & 41.48\% \\
repeat & 71.58\% & 58.95\% & 48.42\% & 38.71\% & 11.35\% \\
marks & 32.32\% & 55.44\% & 13.50\% & 5.86\% & 10.22\% \\
citation & 66.57\% & 75.39\% & 68.04\% & 49.71\% & 29.12\% \\
emoji & 80.56\% & 87.75\% & 80.45\% & 70.90\% & 44.72\% \\
style & 95.32\% & 94.60\% & 84.04\% & 86.20\% & 57.04\% \\
tone & 68.21\% & 77.07\% & 75.24\% & 63.19\% & 24.89\% \\
content & 85.65\% & 84.01\% & 75.92\% & 73.47\% & 2.79\% \\
language\_switch & 91.17\% & 97.23\% & 70.64\% & 43.83\% & 1.81\% \\
\bottomrule
\end{tabular}
\caption{Evaluation results for Zulu language}
\end{table*}
\clearpage
\subsection{Evaluation Results of 47 Subcategories}

\begin{table*}[htbp]
\footnotesize
\centering
\setlength{\tabcolsep}{2pt}
\begin{tabular}{lccccc}
\toprule
\textbf{Metrics} & \textbf{GPT-4o} & \textbf{Claude-3.5 Sonnet} & \textbf{Gemini-1.5 Pro} & \textbf{Gemini-1.5 Flash} & \textbf{GPT-3.5 Turbo} \\
\midrule
keywords:banned & 99.03\% & 98.06\% & 93.87\% & 93.23\% & 72.90\% \\
keywords:first\_word & 83.87\% & 54.84\% & 74.19\% & 61.29\% & 74.19\% \\
keywords:frequency & 83.12\% & 89.38\% & 88.75\% & 80.00\% & 77.08\% \\
keywords:paragraph\_end & 47.22\% & 40.56\% & 17.22\% & 28.89\% & 15.00\% \\
keywords:together & 55.97\% & 42.10\% & 40.97\% & 36.45\% & 36.94\% \\
length:max\_words & 91.99\% & 100.00\% & 96.29\% & 94.36\% & 80.53\% \\
length:range\_words & 85.21\% & 43.29\% & 73.32\% & 51.61\% & 24.53\% \\
\bottomrule
\end{tabular}
\caption{Detailed evaluation results for Armenian language}
\end{table*}

\begin{table*}[htbp]
\footnotesize
\centering
\setlength{\tabcolsep}{2pt}
\begin{tabular}{lccccc}
\toprule
\textbf{Metrics} & \textbf{GPT-4o} & \textbf{Claude-3.5 Sonnet} & \textbf{Gemini-1.5 Pro} & \textbf{Gemini-1.5 Flash} & \textbf{GPT-3.5 Turbo} \\
\midrule
keywords:banned & 90.32\% & 84.52\% & 70.00\% & 46.45\% & 33.87\% \\
keywords:first\_word & 3.23\% & 3.23\% & 3.23\% & 3.23\% & 3.23\% \\
keywords:frequency & 70.62\% & 80.42\% & 79.17\% & 74.38\% & 65.83\% \\
keywords:paragraph\_end & 80.56\% & 42.22\% & 71.67\% & 55.56\% & 53.33\% \\
keywords:together & 80.00\% & 89.52\% & 72.58\% & 72.42\% & 56.45\% \\
length:max\_words & 94.88\% & 100.00\% & 95.18\% & 97.03\% & 78.00\% \\
length:range\_words & 85.16\% & 36.83\% & 56.99\% & 60.17\% & 19.61\% \\
\bottomrule
\end{tabular}
\caption{Detailed evaluation results for Bengali language}
\end{table*}

\begin{table*}[htbp]
\footnotesize
\centering
\setlength{\tabcolsep}{2pt}
\begin{tabular}{lccccc}
\toprule
\textbf{Metrics} & \textbf{GPT-4o} & \textbf{Claude-3.5 Sonnet} & \textbf{Gemini-1.5 Pro} & \textbf{Gemini-1.5 Flash} & \textbf{GPT-3.5 Turbo} \\
\midrule
keywords:banned & 100.00\% & 100.00\% & 100.00\% & 99.03\% & 99.03\% \\
keywords:first\_word & 0.00\% & 0.00\% & 3.23\% & 6.45\% & 0.00\% \\
keywords:frequency & 91.46\% & 86.25\% & 92.71\% & 85.00\% & 80.62\% \\
keywords:paragraph\_end & 86.67\% & 18.33\% & 77.78\% & 65.56\% & 61.67\% \\
keywords:together & 83.23\% & 78.06\% & 83.23\% & 77.74\% & 85.81\% \\
length:max\_words & 76.04\% & 77.66\% & 63.57\% & 62.26\% & 75.78\% \\
length:range\_words & 17.76\% & 14.35\% & 48.48\% & 45.16\% & 12.47\% \\
\bottomrule
\end{tabular}
\caption{Detailed evaluation results for Chinese language}
\end{table*}

\begin{table*}[htbp]
\footnotesize
\centering
\setlength{\tabcolsep}{2pt}
\begin{tabular}{lccccc}
\toprule
\textbf{Metrics} & \textbf{GPT-4o} & \textbf{Claude-3.5 Sonnet} & \textbf{Gemini-1.5 Pro} & \textbf{Gemini-1.5 Flash} & \textbf{GPT-3.5 Turbo} \\
\midrule
keywords:banned & 87.10\% & 94.19\% & 80.32\% & 80.32\% & 60.97\% \\
keywords:first\_word & 90.32\% & 90.32\% & 90.32\% & 87.10\% & 87.10\% \\
keywords:frequency & 93.54\% & 90.62\% & 95.62\% & 90.42\% & 77.92\% \\
keywords:paragraph\_end & 81.67\% & 82.22\% & 76.11\% & 61.67\% & 72.78\% \\
keywords:together & 78.71\% & 76.61\% & 82.74\% & 73.06\% & 75.32\% \\
length:max\_words & 77.79\% & 77.53\% & 93.16\% & 94.12\% & 57.33\% \\
length:range\_words & 82.42\% & 76.89\% & 48.33\% & 78.67\% & 58.68\% \\
\bottomrule
\end{tabular}
\caption{Detailed evaluation results for English language}
\end{table*}

\begin{table*}[htbp]
\footnotesize
\centering
\setlength{\tabcolsep}{2pt}
\begin{tabular}{lccccc}
\toprule
\textbf{Metrics} & \textbf{GPT-4o} & \textbf{Claude-3.5 Sonnet} & \textbf{Gemini-1.5 Pro} & \textbf{Gemini-1.5 Flash} & \textbf{GPT-3.5 Turbo} \\
\midrule
keywords:banned & 100.00\% & 97.10\% & 89.03\% & 91.29\% & 77.74\% \\
keywords:first\_word & 90.32\% & 87.10\% & 70.97\% & 74.19\% & 70.97\% \\
keywords:frequency & 93.12\% & 92.50\% & 93.12\% & 87.08\% & 69.38\% \\
keywords:paragraph\_end & 87.78\% & 81.11\% & 74.44\% & 67.22\% & 47.78\% \\
keywords:together & 80.65\% & 92.26\% & 89.68\% & 76.77\% & 51.29\% \\
length:max\_words & 57.78\% & 75.21\% & 75.62\% & 84.74\% & 66.35\% \\
length:range\_words & 67.29\% & 72.80\% & 70.68\% & 25.13\% & 50.08\% \\
\bottomrule
\end{tabular}
\caption{Detailed evaluation results for Filipino language}
\end{table*}

\begin{table*}[htbp]
\footnotesize
\centering
\setlength{\tabcolsep}{2pt}
\begin{tabular}{lccccc}
\toprule
\textbf{Metrics} & \textbf{GPT-4o} & \textbf{Claude-3.5 Sonnet} & \textbf{Gemini-1.5 Pro} & \textbf{Gemini-1.5 Flash} & \textbf{GPT-3.5 Turbo} \\
\midrule
keywords:banned & 97.10\% & 99.03\% & 83.23\% & 81.94\% & 79.35\% \\
keywords:first\_word & 96.77\% & 96.77\% & 83.87\% & 83.87\% & 64.52\% \\
keywords:frequency & 92.71\% & 91.46\% & 90.00\% & 86.46\% & 75.42\% \\
keywords:paragraph\_end & 86.11\% & 78.33\% & 69.44\% & 62.78\% & 67.22\% \\
keywords:together & 76.29\% & 85.65\% & 73.55\% & 63.55\% & 60.81\% \\
length:max\_words & 82.55\% & 88.48\% & 70.07\% & 80.67\% & 65.67\% \\
length:range\_words & 74.06\% & 79.24\% & 73.81\% & 65.35\% & 73.24\% \\
\bottomrule
\end{tabular}
\caption{Detailed evaluation results for French language}
\end{table*}

\begin{table*}[htbp]
\footnotesize
\centering
\setlength{\tabcolsep}{2pt}
\begin{tabular}{lccccc}
\toprule
\textbf{Metrics} & \textbf{GPT-4o} & \textbf{Claude-3.5 Sonnet} & \textbf{Gemini-1.5 Pro} & \textbf{Gemini-1.5 Flash} & \textbf{GPT-3.5 Turbo} \\
\midrule
keywords:banned & 87.10\% & 75.48\% & 90.00\% & 92.26\% & 60.00\% \\
keywords:first\_word & 77.42\% & 61.29\% & 54.84\% & 58.06\% & 67.74\% \\
keywords:frequency & 81.87\% & 75.42\% & 77.71\% & 73.54\% & 67.08\% \\
keywords:paragraph\_end & 57.22\% & 40.00\% & 38.89\% & 43.33\% & 33.89\% \\
keywords:together & 48.23\% & 30.16\% & 37.90\% & 46.61\% & 32.10\% \\
length:max\_words & 98.90\% & 99.70\% & 100.00\% & 98.80\% & 91.48\% \\
length:range\_words & 79.50\% & 42.82\% & 51.55\% & 46.47\% & 26.94\% \\
\bottomrule
\end{tabular}
\caption{Detailed evaluation results for Georgian language}
\end{table*}

\begin{table*}[htbp]
\footnotesize
\centering
\setlength{\tabcolsep}{2pt}
\begin{tabular}{lccccc}
\toprule
\textbf{Metrics} & \textbf{GPT-4o} & \textbf{Claude-3.5 Sonnet} & \textbf{Gemini-1.5 Pro} & \textbf{Gemini-1.5 Flash} & \textbf{GPT-3.5 Turbo} \\
\midrule
keywords:banned & 100.00\% & 98.06\% & 81.29\% & 85.48\% & 41.61\% \\
keywords:first\_word & 16.13\% & 16.13\% & 16.13\% & 12.90\% & 16.13\% \\
keywords:frequency & 67.92\% & 71.46\% & 62.50\% & 68.13\% & 67.92\% \\
keywords:paragraph\_end & 48.33\% & 45.56\% & 31.67\% & 36.11\% & 31.67\% \\
keywords:together & 68.23\% & 82.74\% & 55.81\% & 50.48\% & 64.68\% \\
length:max\_words & 82.28\% & 99.16\% & 62.61\% & 74.70\% & 43.35\% \\
length:range\_words & 59.75\% & 67.32\% & 48.51\% & 40.42\% & 43.04\% \\
\bottomrule
\end{tabular}
\caption{Detailed evaluation results for Hindi language}
\end{table*}

\begin{table*}[htbp]
\footnotesize
\centering
\setlength{\tabcolsep}{2pt}
\begin{tabular}{lccccc}
\toprule
\textbf{Metrics} & \textbf{GPT-4o} & \textbf{Claude-3.5 Sonnet} & \textbf{Gemini-1.5 Pro} & \textbf{Gemini-1.5 Flash} & \textbf{GPT-3.5 Turbo} \\
\midrule
keywords:banned & 89.03\% & 94.19\% & 69.03\% & 63.87\% & 59.03\% \\
keywords:first\_word & 90.32\% & 90.32\% & 87.10\% & 83.87\% & 67.74\% \\
keywords:frequency & 89.58\% & 88.33\% & 81.67\% & 82.29\% & 69.79\% \\
keywords:paragraph\_end & 77.78\% & 78.33\% & 73.33\% & 70.56\% & 66.67\% \\
keywords:together & 89.68\% & 88.39\% & 87.74\% & 73.87\% & 74.52\% \\
length:max\_words & 86.06\% & 86.85\% & 77.16\% & 82.88\% & 81.11\% \\
length:range\_words & 83.25\% & 72.10\% & 85.15\% & 76.77\% & 45.11\% \\
\bottomrule
\end{tabular}
\caption{Detailed evaluation results for Indonesian language}
\end{table*}

\begin{table*}[htbp]
\footnotesize
\centering
\setlength{\tabcolsep}{2pt}
\begin{tabular}{lccccc}
\toprule
\textbf{Metrics} & \textbf{GPT-4o} & \textbf{Claude-3.5 Sonnet} & \textbf{Gemini-1.5 Pro} & \textbf{Gemini-1.5 Flash} & \textbf{GPT-3.5 Turbo} \\
\midrule
keywords:banned & 94.19\% & 97.10\% & 80.32\% & 85.48\% & 63.23\% \\
keywords:first\_word & 90.32\% & 90.32\% & 70.97\% & 67.74\% & 61.29\% \\
keywords:frequency & 91.67\% & 89.33\% & 85.00\% & 84.79\% & 74.79\% \\
keywords:paragraph\_end & 95.56\% & 80.00\% & 76.11\% & 68.89\% & 72.22\% \\
keywords:together & 86.94\% & 85.81\% & 78.39\% & 67.58\% & 69.19\% \\
length:max\_words & 80.36\% & 85.72\% & 77.24\% & 78.66\% & 62.77\% \\
length:range\_words & 74.04\% & 78.81\% & 76.68\% & 66.15\% & 70.88\% \\
\bottomrule
\end{tabular}
\caption{Detailed evaluation results for Italian language}
\end{table*}

\begin{table*}[htbp]
\footnotesize
\centering
\setlength{\tabcolsep}{2pt}
\begin{tabular}{lccccc}
\toprule
\textbf{Metrics} & \textbf{GPT-4o} & \textbf{Claude-3.5 Sonnet} & \textbf{Gemini-1.5 Pro} & \textbf{Gemini-1.5 Flash} & \textbf{GPT-3.5 Turbo} \\
\midrule
keywords:banned & 99.03\% & 100.00\% & 99.03\% & 100.00\% & 99.03\% \\
keywords:first\_word & 0.00\% & 0.00\% & 16.13\% & 3.23\% & 0.00\% \\
keywords:frequency & 92.92\% & 91.88\% & 93.54\% & 82.92\% & 74.58\% \\
keywords:paragraph\_end & 73.89\% & 18.33\% & 76.11\% & 70.00\% & 51.11\% \\
keywords:together & 77.58\% & 78.06\% & 78.06\% & 71.45\% & 78.39\% \\
length:max\_words & 75.02\% & 83.42\% & 63.16\% & 49.09\% & 73.15\% \\
length:range\_words & 75.47\% & 73.66\% & 68.54\% & 66.23\% & 32.99\% \\
\bottomrule
\end{tabular}
\caption{Detailed evaluation results for Japanese language}
\end{table*}

\begin{table*}[htbp]
\footnotesize
\centering
\setlength{\tabcolsep}{2pt}
\begin{tabular}{lccccc}
\toprule
\textbf{Metrics} & \textbf{GPT-4o} & \textbf{Claude-3.5 Sonnet} & \textbf{Gemini-1.5 Pro} & \textbf{Gemini-1.5 Flash} & \textbf{GPT-3.5 Turbo} \\
\midrule
keywords:banned & 96.13\% & 99.03\% & 87.10\% & 90.32\% & 61.94\% \\
keywords:first\_word & 54.84\% & 48.39\% & 58.06\% & 45.16\% & 48.39\% \\
keywords:frequency & 87.71\% & 87.50\% & 85.42\% & 82.29\% & 70.00\% \\
keywords:paragraph\_end & 41.11\% & 26.67\% & 29.44\% & 28.89\% & 32.78\% \\
keywords:together & 70.00\% & 77.90\% & 71.77\% & 62.10\% & 64.68\% \\
length:max\_words & 89.09\% & 88.57\% & 47.51\% & 67.14\% & 30.57\% \\
length:range\_words & 74.12\% & 78.40\% & 63.06\% & 68.26\% & 22.54\% \\
\bottomrule
\end{tabular}
\caption{Detailed evaluation results for Korean language}
\end{table*}

\begin{table*}[htbp]
\footnotesize
\centering
\setlength{\tabcolsep}{2pt}
\begin{tabular}{lccccc}
\toprule
\textbf{Metrics} & \textbf{GPT-4o} & \textbf{Claude-3.5 Sonnet} & \textbf{Gemini-1.5 Pro} & \textbf{Gemini-1.5 Flash} & \textbf{GPT-3.5 Turbo} \\
\midrule
keywords:banned & 98.06\% & 100.00\% & 93.23\% & 94.19\% & 12.90\% \\
keywords:first\_word & 87.10\% & 70.97\% & 80.65\% & 87.10\% & 19.35\% \\
keywords:frequency & 80.83\% & 81.88\% & 78.54\% & 75.63\% & 66.25\% \\
keywords:paragraph\_end & 77.22\% & 54.44\% & 50.00\% & 32.78\% & 32.22\% \\
keywords:together & 83.06\% & 71.13\% & 76.94\% & 61.45\% & 34.52\% \\
length:max\_words & 99.42\% & 96.90\% & 97.68\% & 97.12\% & 68.27\% \\
length:range\_words & 85.54\% & 39.35\% & 73.73\% & 73.76\% & 16.35\% \\
\bottomrule
\end{tabular}
\caption{Detailed evaluation results for Kyrgyz language}
\end{table*}

\begin{table*}[htbp]
\footnotesize
\centering
\setlength{\tabcolsep}{2pt}
\begin{tabular}{lccccc}
\toprule
\textbf{Metrics} & \textbf{GPT-4o} & \textbf{Claude-3.5 Sonnet} & \textbf{Gemini-1.5 Pro} & \textbf{Gemini-1.5 Flash} & \textbf{GPT-3.5 Turbo} \\
\midrule
keywords:banned & 87.74\% & 92.26\% & 79.68\% & 77.10\% & 48.39\% \\
keywords:first\_word & 64.52\% & 64.52\% & 61.29\% & 51.61\% & 12.90\% \\
keywords:frequency & 87.71\% & 87.08\% & 83.54\% & 71.25\% & 73.75\% \\
keywords:paragraph\_end & 91.67\% & 78.33\% & 69.44\% & 67.22\% & 10.56\% \\
keywords:together & 77.10\% & 85.81\% & 57.58\% & 59.03\% & 49.52\% \\
length:max\_words & 72.53\% & 92.47\% & 99.20\% & 85.00\% & 74.02\% \\
length:range\_words & 73.79\% & 70.81\% & 80.62\% & 63.87\% & 19.36\% \\
\bottomrule
\end{tabular}
\caption{Detailed evaluation results for Malagasy language}
\end{table*}

\begin{table*}[htbp]
\footnotesize
\centering
\setlength{\tabcolsep}{2pt}
\begin{tabular}{lccccc}
\toprule
\textbf{Metrics} & \textbf{GPT-4o} & \textbf{Claude-3.5 Sonnet} & \textbf{Gemini-1.5 Pro} & \textbf{Gemini-1.5 Flash} & \textbf{GPT-3.5 Turbo} \\
\midrule
keywords:banned & 96.13\% & 98.06\% & 86.13\% & 84.52\% & 79.35\% \\
keywords:first\_word & 87.10\% & 87.10\% & 80.65\% & 87.10\% & 83.87\% \\
keywords:frequency & 90.42\% & 87.29\% & 86.46\% & 88.96\% & 73.33\% \\
keywords:paragraph\_end & 86.67\% & 80.00\% & 75.00\% & 72.78\% & 51.67\% \\
keywords:together & 86.61\% & 85.81\% & 71.29\% & 78.87\% & 69.68\% \\
length:max\_words & 92.70\% & 88.82\% & 90.79\% & 86.42\% & 82.69\% \\
length:range\_words & 75.62\% & 86.84\% & 87.00\% & 67.60\% & 57.20\% \\
\bottomrule
\end{tabular}
\caption{Detailed evaluation results for Malay language}
\end{table*}

\begin{table*}[htbp]
\footnotesize
\centering
\setlength{\tabcolsep}{2pt}
\begin{tabular}{lccccc}
\toprule
\textbf{Metrics} & \textbf{GPT-4o} & \textbf{Claude-3.5 Sonnet} & \textbf{Gemini-1.5 Pro} & \textbf{Gemini-1.5 Flash} & \textbf{GPT-3.5 Turbo} \\
\midrule
keywords:banned & 93.87\% & 99.03\% & 81.61\% & 89.68\% & 65.81\% \\
keywords:first\_word & 90.32\% & 74.19\% & 70.97\% & 61.29\% & 70.97\% \\
keywords:frequency & 92.71\% & 87.92\% & 91.88\% & 86.04\% & 77.71\% \\
keywords:paragraph\_end & 86.67\% & 78.89\% & 71.67\% & 70.00\% & 60.00\% \\
keywords:together & 80.00\% & 75.97\% & 72.58\% & 63.71\% & 72.58\% \\
length:max\_words & 79.14\% & 80.62\% & 71.23\% & 78.15\% & 47.75\% \\
length:range\_words & 67.39\% & 81.62\% & 73.95\% & 65.02\% & 69.40\% \\
\bottomrule
\end{tabular}
\caption{Detailed evaluation results for Portuguese language}
\end{table*}

\begin{table*}[htbp]
\footnotesize
\centering
\setlength{\tabcolsep}{2pt}
\begin{tabular}{lccccc}
\toprule
\textbf{Metrics} & \textbf{GPT-4o} & \textbf{Claude-3.5 Sonnet} & \textbf{Gemini-1.5 Pro} & \textbf{Gemini-1.5 Flash} & \textbf{GPT-3.5 Turbo} \\
\midrule
keywords:banned & 18.39\% & 100.00\% & 91.61\% & 73.23\% & 3.23\% \\
keywords:first\_word & 29.03\% & 87.10\% & 51.61\% & 74.19\% & 0.00\% \\
keywords:frequency & 81.67\% & 84.38\% & 67.29\% & 69.17\% & 70.42\% \\
keywords:paragraph\_end & 37.78\% & 83.89\% & 45.56\% & 44.44\% & 23.89\% \\
keywords:together & 45.48\% & 77.58\% & 56.77\% & 56.29\% & 45.32\% \\
length:max\_words & 78.95\% & 100.00\% & 100.00\% & 80.18\% & 86.83\% \\
length:range\_words & 9.52\% & 13.67\% & 14.54\% & 26.08\% & 5.69\% \\
\bottomrule
\end{tabular}
\caption{Detailed evaluation results for Quechua language}
\end{table*}

\begin{table*}[htbp]
\footnotesize
\centering
\setlength{\tabcolsep}{2pt}
\begin{tabular}{lccccc}
\toprule
\textbf{Metrics} & \textbf{GPT-4o} & \textbf{Claude-3.5 Sonnet} & \textbf{Gemini-1.5 Pro} & \textbf{Gemini-1.5 Flash} & \textbf{GPT-3.5 Turbo} \\
\midrule
keywords:banned & 98.06\% & 99.03\% & 95.16\% & 94.19\% & 86.45\% \\
keywords:first\_word & 90.32\% & 83.87\% & 83.87\% & 74.19\% & 61.29\% \\
keywords:frequency & 64.58\% & 68.96\% & 68.13\% & 65.42\% & 58.13\% \\
keywords:paragraph\_end & 68.89\% & 50.56\% & 45.56\% & 35.00\% & 41.11\% \\
keywords:together & 40.32\% & 54.52\% & 34.35\% & 35.16\% & 26.29\% \\
length:max\_words & 86.82\% & 80.24\% & 84.18\% & 83.81\% & 66.10\% \\
length:range\_words & 82.13\% & 89.21\% & 85.51\% & 73.08\% & 72.42\% \\
\bottomrule
\end{tabular}
\caption{Detailed evaluation results for Romanian language}
\end{table*}

\begin{table*}[htbp]
\footnotesize
\centering
\setlength{\tabcolsep}{2pt}
\begin{tabular}{lccccc}
\toprule
\textbf{Metrics} & \textbf{GPT-4o} & \textbf{Claude-3.5 Sonnet} & \textbf{Gemini-1.5 Pro} & \textbf{Gemini-1.5 Flash} & \textbf{GPT-3.5 Turbo} \\
\midrule
keywords:banned & 97.10\% & 98.06\% & 90.97\% & 91.94\% & 92.26\% \\
keywords:first\_word & 74.19\% & 74.19\% & 77.42\% & 74.19\% & 67.74\% \\
keywords:frequency & 78.75\% & 82.29\% & 84.58\% & 82.92\% & 70.83\% \\
keywords:paragraph\_end & 81.67\% & 73.33\% & 62.22\% & 52.78\% & 46.11\% \\
keywords:together & 74.19\% & 77.26\% & 74.84\% & 69.35\% & 56.61\% \\
length:max\_words & 88.08\% & 93.52\% & 89.48\% & 92.09\% & 76.36\% \\
length:range\_words & 86.16\% & 98.49\% & 81.17\% & 75.51\% & 71.33\% \\
\bottomrule
\end{tabular}
\caption{Detailed evaluation results for Swedish language}
\end{table*}

\begin{table*}[htbp]
\footnotesize
\centering
\setlength{\tabcolsep}{2pt}
\begin{tabular}{lccccc}
\toprule
\textbf{Metrics} & \textbf{GPT-4o} & \textbf{Claude-3.5 Sonnet} & \textbf{Gemini-1.5 Pro} & \textbf{Gemini-1.5 Flash} & \textbf{GPT-3.5 Turbo} \\
\midrule
keywords:banned & 99.03\% & 97.10\% & 93.23\% & 93.23\% & 83.23\% \\
keywords:first\_word & 16.13\% & 16.13\% & 16.13\% & 12.90\% & 6.45\% \\
keywords:frequency & 78.54\% & 77.71\% & 68.96\% & 78.75\% & 68.12\% \\
keywords:paragraph\_end & 20.56\% & 18.33\% & 18.33\% & 18.33\% & 13.89\% \\
keywords:together & 66.29\% & 42.74\% & 27.10\% & 27.58\% & 35.32\% \\
length:max\_words & 99.93\% & 100.00\% & 98.50\% & 99.83\% & 94.95\% \\
length:range\_words & 31.12\% & 12.12\% & 63.95\% & 58.00\% & 35.17\% \\
\bottomrule
\end{tabular}
\caption{Detailed evaluation results for Tamil language}
\end{table*}

\begin{table*}[htbp]
\footnotesize
\centering
\setlength{\tabcolsep}{2pt}
\begin{tabular}{lccccc}
\toprule
\textbf{Metrics} & \textbf{GPT-4o} & \textbf{Claude-3.5 Sonnet} & \textbf{Gemini-1.5 Pro} & \textbf{Gemini-1.5 Flash} & \textbf{GPT-3.5 Turbo} \\
\midrule
keywords:banned & 98.06\% & 100.00\% & 98.06\% & 97.10\% & 85.48\% \\
keywords:first\_word & 12.90\% & 9.68\% & 12.90\% & 6.45\% & 12.90\% \\
keywords:frequency & 75.21\% & 86.25\% & 73.96\% & 77.71\% & 75.21\% \\
keywords:paragraph\_end & 36.67\% & 31.11\% & 33.89\% & 25.00\% & 13.89\% \\
keywords:together & 64.03\% & 59.03\% & 53.71\% & 52.74\% & 49.68\% \\
length:max\_words & 100.00\% & 97.37\% & 98.86\% & 99.16\% & 91.89\% \\
length:range\_words & 51.65\% & 44.49\% & 54.55\% & 65.52\% & 18.63\% \\
\bottomrule
\end{tabular}
\caption{Detailed evaluation results for Telugu language}
\end{table*}

\begin{table*}[htbp]
\footnotesize
\centering
\setlength{\tabcolsep}{2pt}
\begin{tabular}{lccccc}
\toprule
\textbf{Metrics} & \textbf{GPT-4o} & \textbf{Claude-3.5 Sonnet} & \textbf{Gemini-1.5 Pro} & \textbf{Gemini-1.5 Flash} & \textbf{GPT-3.5 Turbo} \\
\midrule
keywords:banned & 97.10\% & 99.03\% & 72.26\% & 67.42\% & 68.06\% \\
keywords:first\_word & 58.06\% & 48.39\% & 54.84\% & 61.29\% & 61.29\% \\
keywords:frequency & 76.67\% & 80.00\% & 77.92\% & 79.17\% & 76.25\% \\
keywords:paragraph\_end & 80.00\% & 57.78\% & 43.89\% & 43.89\% & 38.89\% \\
keywords:together & 76.94\% & 60.65\% & 76.77\% & 78.71\% & 65.32\% \\
length:max\_words & 93.14\% & 86.49\% & 90.73\% & 91.03\% & 89.69\% \\
length:range\_words & 78.86\% & 74.06\% & 77.67\% & 76.36\% & 52.38\% \\
\bottomrule
\end{tabular}
\caption{Detailed evaluation results for Turkish language}
\end{table*}

\begin{table*}[htbp]
\footnotesize
\centering
\setlength{\tabcolsep}{2pt}
\begin{tabular}{lccccc}
\toprule
\textbf{Metrics} & \textbf{GPT-4o} & \textbf{Claude-3.5 Sonnet} & \textbf{Gemini-1.5 Pro} & \textbf{Gemini-1.5 Flash} & \textbf{GPT-3.5 Turbo} \\
\midrule
keywords:banned & 99.03\% & 100.00\% & 92.26\% & 95.16\% & 43.87\% \\
keywords:first\_word & 48.39\% & 70.97\% & 41.94\% & 51.61\% & 61.29\% \\
keywords:frequency & 76.25\% & 95.00\% & 74.37\% & 69.17\% & 60.83\% \\
keywords:paragraph\_end & 81.11\% & 72.22\% & 37.22\% & 38.89\% & 25.00\% \\
keywords:together & 60.32\% & 58.39\% & 45.65\% & 33.39\% & 46.77\% \\
length:max\_words & 88.51\% & 100.00\% & 100.00\% & 94.70\% & 55.91\% \\
length:range\_words & 73.12\% & 38.67\% & 32.34\% & 43.61\% & 6.31\% \\
\bottomrule
\end{tabular}
\caption{Detailed evaluation results for Zulu language}
\end{table*}
\clearpage
\subsection{Cross-lingual Results from English and Original Results in 22 Languages}

\begin{table*}[htbp]
    \centering
    \resizebox{\textwidth}{!}{%
    \begin{tabular}{llcc}
        \toprule
        \textbf{Resource} & \textbf{Language} & \textbf{Cross-lingual Result from English} & \textbf{Original Result} \\
        \midrule
        \multirow{7}{*}{\textbf{High Resource}}
        & Swedish    & 71.11\% & 69.39\% \\
        & Portuguese & 70.03\% & 70.79\% \\
        & French     & 68.64\% & 68.10\% \\
        & Italian    & 69.39\% & 66.57\% \\
        & Chinese    & 66.59\% & 67.18\% \\
        & Japanese   & 65.80\% & 66.01\% \\
        \cmidrule{2-4}
        & \textit{Average High} & \textbf{68.59\%} & \textbf{68.01\%} \\
        \midrule
        \multirow{9}{*}{\textbf{Medium Resource}}
        & Filipino   & 69.24\% & 60.21\% \\
        & Romanian   & 67.87\% & 67.33\% \\
        & Indonesian & 69.92\% & 64.25\% \\
        & Malay      & 68.33\% & 63.94\% \\
        & Turkish    & 70.45\% & 62.77\% \\
        & Korean     & 67.21\% & 60.74\% \\
        & Bengali    & 57.29\% & 38.58\% \\
        & Hindi      & 66.01\% & 51.41\% \\
        \cmidrule{2-4}
        & \textit{Average Medium} & \textbf{67.04\%} & \textbf{58.65\%} \\
        \midrule
        \multirow{9}{*}{\textbf{Low Resource}}
        & Kyrgyz     & 59.09\% & 29.34\% \\
        & Armenian   & 51.71\% & 33.02\% \\
        & Georgian   & 52.51\% & 34.28\% \\
        & Malagasy   & 55.88\% & 29.74\% \\
        & Zulu       & 49.80\% & 30.14\% \\
        & Tamil      & 53.64\% & 33.22\% \\
        & Telugu     & 55.60\% & 33.00\% \\
        & Quechua    & 43.59\% & 29.52\% \\
        \cmidrule{2-4}
        & \textit{Average Low} & \textbf{52.73\%} & \textbf{31.53\%} \\
        \bottomrule
    \end{tabular}
    }
    \caption{Comparison of Cross-lingual Results from English and Original Results in 22 Languages on GPT-3.5 Turbo}
    \label{tab:cross_lingual_3.5}
\end{table*}

\begin{table*}[htbp]
    \centering
    \resizebox{\textwidth}{!}{%
    \begin{tabular}{llcc}
        \toprule
        \textbf{Resource} & \textbf{Language} & \textbf{Cross-lingual Result from English} & \textbf{Original Result} \\
        \midrule
        \multirow{7}{*}{\textbf{High Resource}}
        & Swedish    & 79.13\% & 77.67\% \\
        & Portuguese & 81.45\% & 78.17\% \\
        & French     & 80.39\% & 76.27\% \\
        & Italian    & 80.04\% & 76.32\% \\
        & Chinese    & 74.46\% & 74.79\% \\
        & Japanese   & 73.59\% & 77.17\% \\
        \cmidrule{2-4}
        & \textit{Average High} & \textbf{78.18\%} & \textbf{76.73\%} \\
        \midrule
        \multirow{9}{*}{\textbf{Medium Resource}}
        & Filipino   & 80.57\% & 75.75\% \\
        & Romanian   & 77.77\% & 76.10\% \\
        & Indonesian & 78.74\% & 74.09\% \\
        & Malay      & 80.45\% & 73.92\% \\
        & Turkish    & 80.07\% & 70.83\% \\
        & Korean     & 73.61\% & 70.89\% \\
        & Bengali    & 76.18\% & 68.56\% \\
        & Hindi      & 75.41\% & 67.92\% \\
        \cmidrule{2-4}
        & \textit{Average Medium} & \textbf{77.85\%} & \textbf{72.26\%} \\
        \midrule
        \multirow{9}{*}{\textbf{Low Resource}}
        & Kyrgyz     & 78.52\% & 74.89\% \\
        & Armenian   & 74.91\% & 72.60\% \\
        & Georgian   & 74.29\% & 68.55\% \\
        & Malagasy   & 79.84\% & 64.73\% \\
        & Zulu       & 76.92\% & 60.29\% \\
        & Tamil      & 72.42\% & 68.46\% \\
        & Telugu     & 73.54\% & 68.77\% \\
        & Quechua    & 66.48\% & 48.39\% \\
        \cmidrule{2-4}
        & \textit{Average Low} & \textbf{74.62\%} & \textbf{65.84\%} \\
        \bottomrule
    \end{tabular}
    }
    \caption{Comparison of Cross-lingual Results from English and Original Results in 22 Languages on Gemini-1.5 Flash}
    \label{tab:cross_lingual_flash}
\end{table*}

\end{document}